\pdfoutput=1
\documentclass[11pt]{article}

\usepackage[]{acl}

\usepackage{times}
\usepackage{latexsym}

\usepackage[T1]{fontenc}
\usepackage[utf8]{inputenc}

\usepackage{color}
\usepackage{tabularx}
\usepackage{tabu}
\usepackage{booktabs}
\usepackage[table]{xcolor}
\usepackage{array}
\usepackage{makecell}
\newcolumntype{L}{>{\raggedright\arraybackslash}p{4.6cm}}
\newcolumntype{Y}{>{\centering\arraybackslash}X}
\usepackage{tcolorbox}
\usepackage{amssymb}
\usepackage{multirow}

\usepackage{graphicx} 
\usepackage{float}
\usepackage{subfigure} 
\usepackage{amssymb}

\usepackage{hyperref}
\usepackage{tablefootnote}
\usepackage{xspace}

\usepackage{float}
\usepackage{amsmath}
\usepackage{bm}
\usepackage{algorithmicx}
\usepackage{algorithm}
\usepackage{algpseudocode}

\usepackage{arydshln}

\usepackage{amssymb}
\usepackage{pifont}
\newcommand{\cmark}{\ding{51}}%
\newcommand{\xmark}{\ding{55}}%
\usepackage{enumitem}

\usepackage{microtype}

\newcommand{\ours}{\textsc{ClinSQL}\xspace}
\newcommand{\eg}{\hbox{\emph{e.g.,}}\xspace}
\newcommand{\ie}{\hbox{\emph{i.e.,}}\xspace}
\newcommand{\huggingface}{\raisebox{-1.5pt}{\includegraphics[height=1.05em]{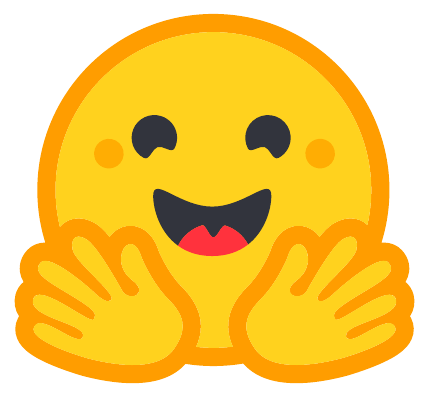}}\xspace}
\newcommand{\github}{\raisebox{-1.5pt}{\includegraphics[height=1.05em]{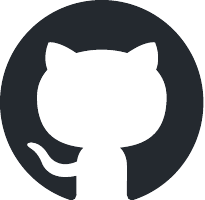}}\xspace}

\title{Patient‑Similarity Cohort Reasoning in Clinical Text-to-SQL}

\definecolor{YaleBlue}{RGB}{16, 42, 86}
\definecolor{UWPurple}{RGB}{75, 46, 131}

\newcommand{\Yale}{\hspace{.1em}^{\textcolor{YaleBlue}{\boldsymbol{Y}}}}
\newcommand{\UW}{\hspace{.1em}^{\textcolor{UWPurple}{\boldsymbol{W}}}}

\author{
Yifei Shen\thanks{~~Equal Contributions. Correspondence: Yilun Zhao (\texttt{yilun.zhao@yale.edu})}~$\UW$ \quad Yilun Zhao$^{*}$$\Yale$ \quad Justice Ou$\Yale$ \quad Tinglin Huang$\Yale$ \quad Arman Cohan$\Yale$ \vspace{4pt}\\
$\UW$ University of Washington \quad $\Yale$~Yale University
}

\begin{document}
\maketitle
\begin{abstract}
Real-world clinical text-to-SQL requires reasoning over heterogeneous EHR tables, temporal windows, and patient-similarity cohorts to produce executable queries. We introduce \ours, a benchmark of 633 expert-annotated tasks on MIMIC-IV v3.1 that demands multi-table joins, clinically meaningful filters, and executable SQL. Solving \ours entails navigating schema metadata and clinical coding systems, handling long contexts, and composing multi-step queries beyond traditional text-to-SQL. We evaluate 22 proprietary and open-source models under Chain-of-Thought self-refinement and use rubric-based SQL analysis with execution checks that prioritize critical clinical requirements. Despite recent advances, performance remains far from clinical reliability: on the test set, GPT-5-mini attains 74.7\% execution score, DeepSeek-R1 leads open-source at 69.2\% and Gemini-2.5-Pro drops from 85.5\% on Easy to 67.2\% on Hard. Progress on \ours marks tangible advances toward clinically reliable text-to-SQL for real-world EHR analytics.

\begin{small}
\begin{center}
\begin{tabular}{cll}
\huggingface & \textbf{Data} & \href{https://huggingface.co/datasets/yifeis02/ClinSQL} {\path{yifeis02/ClinSQL}}\\
\github & \textbf{Code} & \href{https://github.com/Barryshen1/ClinSQL}{\path{Barryshen1/ClinSQL}}\\
\end{tabular}
\end{center} 
\end{small}

\end{abstract}

\begin{table*}[!t]
\centering
\renewcommand{\arraystretch}{1.05}
\resizebox{\textwidth}{!}{%
\addtolength{\tabcolsep}{-0.45em}
\begin{tabular}{llllc}
\toprule
\multirow{2}{*}{\textbf{Dataset}} & \multirow{2}{*}{\textbf{Task}} & \multirow{2}{*}{\textbf{Source}} & \multirow{2}{*}{\textbf{Data Construction}} & \multirow{2}{*}{\textbf{\begin{tabular}[c]{@{}c@{}}Patient\_id\\ Optional\end{tabular}}} \\
& & & & \\
\midrule
\multicolumn{5}{c}{\textbf{\emph{General Text-to-SQL Benchmarks}}}\\
WikiSQL~\citep{zhong2017seq2sql} & Single-table Text-to-SQL & Wikipedia and SQL tables & Crowdsourcing & - \\
Spider~\citep{yu2018spider} & Cross-domain, multi-table Text-to-SQL & Diverse real DB schemas & Expert annotation & - \\
Spider~2.0~\citep{wei2024spider} & Real-world enterprise workflows & Enterprise-scale DBs & Expert + synthetic & - \\
KaggleDBQA~\citep{lee2021kaggledbqa} & Realistic DBs from Kaggle & Real-world multi-table DBs & Author-written Qs & - \\
BIRD~\citep{li2023bird} & Large-scale Text-to-SQL & 95 DBs across 37 domains & Crowdsourcing + expert review & - \\
LiveBench~\citep{white2024livebench} & Contamination-limited evaluation & Mixed sources incl.\ DB tasks & Expert-authored, verifiable & - \\
\midrule
\multicolumn{5}{c}{\textbf{\emph{Healthcare Benchmarks}}} \\
PubMedQA~\citep{jin2019pubmedqa} & Biomedical QA & PubMed abstracts & Heuristic generation + manual labels & - \\
MedQA~\citep{jin2021medqa} & Exam-style multiple-choice QA & Medical board exam questions & Exam scrape & - \\
MedMCQA~\citep{pal2022medmcqa} & Broad medical MCQ QA & Multi-subject exam questions & Exam scrape & - \\
MedExQA~\citep{kim-etal-2024-medexqa} & Medical QA w/ explanations & Mock tests \& online exams & Manual collection/cleaning & - \\
MedXpertQA~\citep{zhang2025medxpertqa} & Expert-level multimodal medical QA & Specialty board Qs; multimodal clinical info & Collection + filtering + synthesis; expert review & - \\
emrQA~\citep{pampari2018emrqa} & Template-driven clinical QA & De-identified clinical notes i2b2 & Template generation (i2b2) & \xmark \\
DrugEHRQA~\citep{wang2022drugehrqa} & Medication-centric QA & EHR notes + structured meds & Template generation + sample human check & \xmark \\
EHRXQA~\citep{bae2023ehrxqa} & Multi-modal EHR QA & Notes + chest X-ray images & Derived from MIMIC-CXR-VQA \& EHRSQL; curated & \xmark \\
EHRNoteQA~\citep{kweon2024ehrnoteqa} & Discharge-summary QA & Real EHR discharge summaries & GPT-4 generation + clinician review & \xmark \\
DischargeQA~\citep{ou2025experienceretrievalaugmentationelectronichealth} & Discharge-related clinical QA & EHR discharge summaries & Generated from discharge data & \xmark \\
RadQA~\citep{soni2022radqa} & Radiology report QA & Radiology reports & Physician-authored Qs + span annotation & \xmark \\
\midrule
\multicolumn{5}{c}{\textbf{\emph{EHR Text-to-SQL Benchmarks}}} \\
MIMICSQL~\citep{wang2020mimicsql} & NL $\rightarrow$ SQL clinical & MIMIC-III structured tables & Auto-generated Qs + crowdsourcing filter & \cmark \\
EHRSQL~\citep{lee2022ehrsql} & Practical NL $\rightarrow$ SQL & Hospital EHR schemas & Hospital-staff utterances + manual SQL annotation & \cmark \\
EHRSQL-ST~\citep{lee-etal-2024-overview} & Reliable Text-to-SQL evaluation & Same family of EHR schemas & Organizer-curated evaluation splits & \cmark \\
EHR-SeqSQL~\citep{ryu-etal-2024-ehr} & NL $\rightarrow$ SQL EHR & Institutional EHR DB & Decomposition of EHRSQL into sequential tasks & \cmark \\
\midrule\midrule
\textbf{\ours} & Text-to-SQL with advanced reasoning & EHR tables & Expert annotation + validation; fine-grained eval rubrics & \cmark \\
\bottomrule
\end{tabular}%
}
\caption{Comparison of \ours with existing Text-to-SQL and Healthcare Benchmarks. The ``Patient\_id Optional'' column indicates whether a benchmark supports supplying an optional de-identified anchor patient identifier (\eg MIMIC \texttt{subject\_id}/\texttt{hadm\_id}) alongside the question to ground patient-similarity or patient-specific queries. \cmark: supported; \xmark: not supported; ``-": not applicable.}
\label{tab:benchmarks_comparison}
\end{table*}

\section{Introduction}

Automating clinical data analysis requires bridging natural-language questions from clinicians to executable queries over complex electronic health record (EHR) databases. While large language models (LLMs) have recently excelled at text-to-SQL and database reasoning on general-domain benchmarks~\cite{yu2018spider, wei2024spider, yang-etal-2025-table}, real-world clinical analysis presents distinct challenges: specialized medical terminology, fine-grained temporal reasoning across heterogeneous tables, and cohort-level clinical reasoning that goes beyond point retrieval to compare \emph{similar} patients under clinically meaningful constraints \citep{yu2018spider,li2023bird,wei2024spider}. These requirements are not merely larger versions of the classic text-to-SQL problem; they demand workflows that integrate domain knowledge, temporal windows, coding systems, and outcome-aware analytics over longitudinal data \citep{johnson2023mimic}.
\begin{figure}[!t]
  \centering
  \includegraphics[width=\columnwidth]{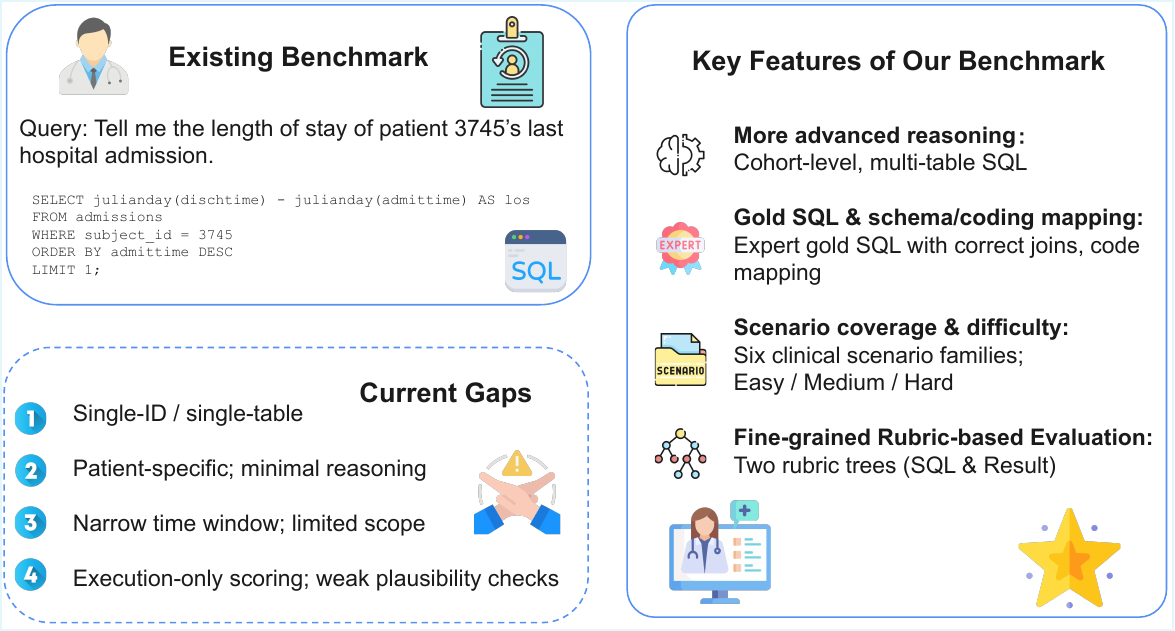}
  \caption{Overview of the \ours benchmark.}
  \label{fig:overview}
\end{figure}

Foundational text-to-SQL evaluations (\eg WikiSQL, Spider 1.0, BIRD) catalyze progress on cross-domain parsing and database generalization \citep{zhong2017seq2sql,yu2018spider,li2023bird}. Recent enterprise-style benchmarks (\ie Spider~2.0) further expose challenges from large schemas, diverse SQL dialects, and multi-step workflows \citep{wei2024spider}. However, clinical settings introduce additional, domain-specific hurdles: temporal abstractions (\eg first 24/48/72 hours), clinical ranges/units, ICD/medication coding, and cohort construction for outcome comparison. Prior clinical text-to-SQL datasets, notably MIMICSQL~\citep{wang2020mimicsql} and EHRSQL~\cite{lee2022ehrsql}, demonstrate feasibility on EHR schemas but predominantly emphasize single-patient or statistical summaries and seldom require \emph{patient-similarity} cohort reasoning central to real-world clinical decision making.

\begin{figure*}[t]
  \centering
  \includegraphics[width=0.97\textwidth]{\detokenize{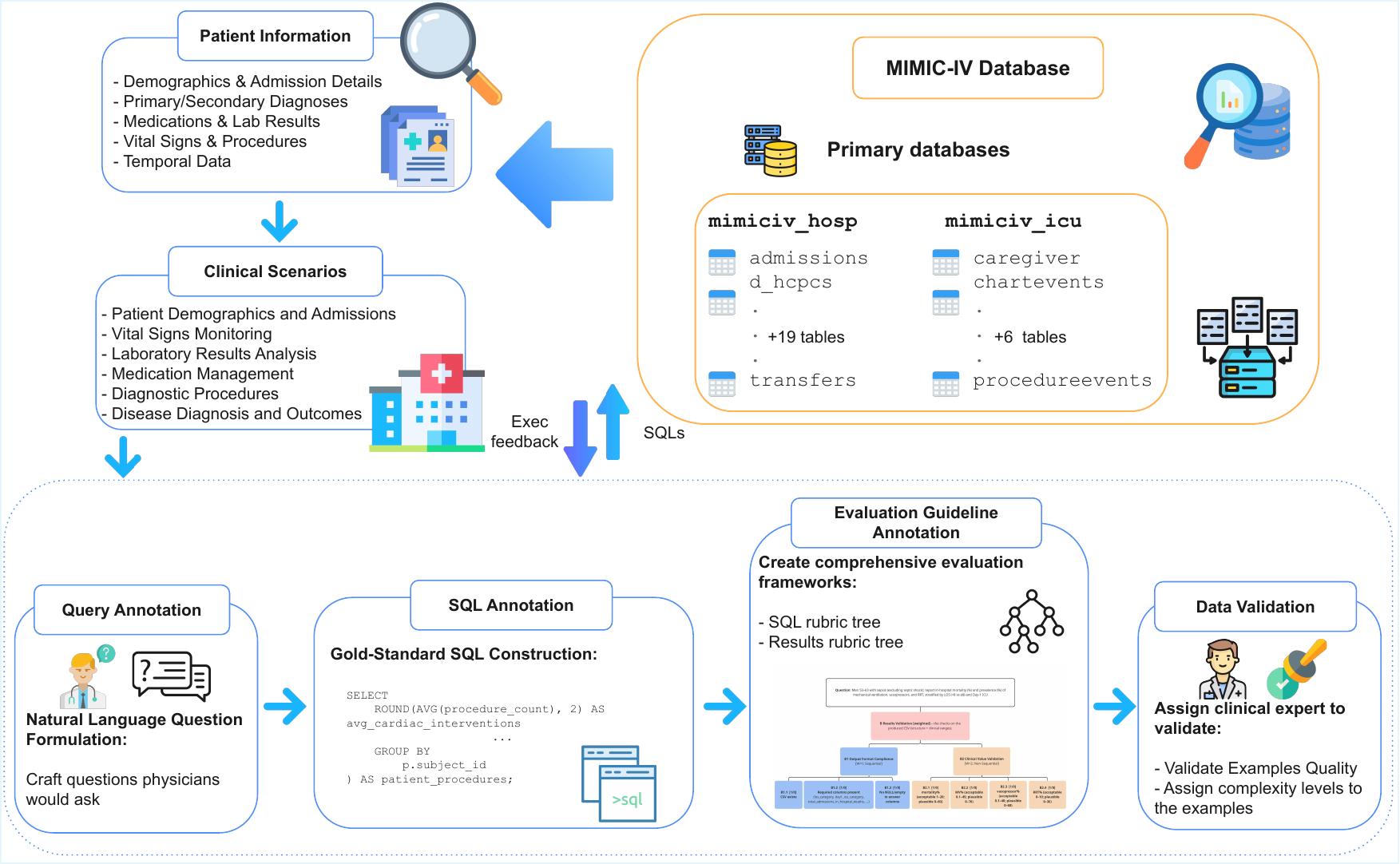}}
  \caption{Overview of \ours construction pipeline. The process begins with scenario design and patient selection, followed by question authoring. Annotators then perform database analysis and schema mapping on MIMIC-IV, write executable gold SQL, and construct tree-structured rubrics for SQL and results validation.}
  \label{fig:figure1}
\end{figure*}

To bridge this gap, we introduce \ours, a benchmark of 633 expert-annotated clinical text-to-SQL tasks on the MIMIC-IV v3.1 database \citep{johnson2023mimic}. A high-level benchmark overview appears in \autoref{fig:overview}, and \autoref{fig:figure1} details the construction pipeline: we design six scenario types to reflect real clinical settings. Each example is grounded in a concrete scenario and requires composing multi-table, temporally aware SQL with \emph{patient-similarity} cohort construction. Difficulty is stratified by SQL and clinical reasoning complexity. We adopt rubric-based evaluation with critical-first aggregation and execution checks that verify result format and clinical plausibility while allowing equivalent formulations.

We evaluate 22 proprietary and open-source models with Chain-of-Thought self-refinement and find that \ours remains challenging: Gemini-2.5-Pro drops from 85.5\% execution on Easy to 67.2\% on Hard; GPT-5-mini leads overall test execution at 74.7\%, and DeepSeek-R1 tops open-source models at 69.1\%. Even for these models, Hard split execution remains below 70\%, underscoring the difficulty of \ours.
Our error analysis reveals that most failures stem from cohort specification drift (\eg relaxed ICD/item constraints), schema or output mismatches, and mis-specified clinical aggregations, even for top-performing models. Guided by these findings, we further study a schema-hinted inference setting that foregrounds clinically validated filters and expected outputs, yielding consistent execution gains, especially on medium and hard cases.

We summarize our contributions as follows:
\begin{itemize} [leftmargin=*]
\itemsep0em 
\item We introduce a clinically grounded text-to-SQL benchmark that requires \emph{patient-similarity} cohort construction and multi-step temporal reasoning over heterogeneous EHR tables.

\item We curate six families of realistic clinical scenarios and provide a rubric-structured evaluation for reliable automated evaluation.

\item We benchmark 22 proprietary and open-source models and release a rubric-based error taxonomy that highlights the challenges that future clinical text-to-SQL systems must address.

\end{itemize}

\section{Related Work}

\paragraph{General Text-to-SQL Benchmarks.}
General-domain Text-to-SQL has evolved through successive foundational benchmarks~\citep{zhong2017seq2sql, yu2018spider, lee2021kaggledbqa, li2023bird, wei2024spider,livebench}. However, healthcare Text-to-SQL presents fundamental challenges that distinguish it from general-domain applications, requiring medical terminology, complex temporal relationships, and clinical reasoning that extends beyond standard database operations to incorporate medical decision-making logic~\cite{lee2022ehrsql,wang2020mimicsql}. Most critically, the patient similarity reasoning paradigm central to \ours represents a fundamental departure from general Text-to-SQL evaluation, as healthcare queries require identifying patient cohorts based on multi-dimensional similarity criteria rather than simple retrieval or aggregation operations.

\paragraph{Healthcare NLP/ML Benchmarks.}
Healthcare NLP benchmarks have evolved from general medical knowledge QA~\citep{jin-etal-2019-pubmedqa, hendrycks2020measuring, jin2021medqa, pal2022medmcqa, singhal2022largelanguagemodelsencode, wang2024mmlupro, kim-etal-2024-medexqa} to sophisticated clinical QA tasks with expert-level capabilities and real clinical data utilization~\citep{pampari2018emrqa, Saleh2019ClefEhealthCLIR,Suominen2020ClefEhealth, bardhan-etal-2022-drugehrqa, bae2023ehrxqa, kweon2024ehrnoteqa, kim-etal-2024-medexqa, zhang2025medxpertqa, chen-etal-2025-benchmarking, ou2025experienceretrievalaugmentationelectronichealth}. 
% Multilingual EHR evaluation efforts from CLEF eHealth emphasize cross-language clinical IR/IE and patient-facing understanding\citep{Suominen2013ShareClef,Saleh2019ClefEhealthCLIR,Suominen2020ClefEhealth}, and our structured, SQL-based cohort analytics are complementary. This line of work points to multilingual clinical Text-to-SQL as a natural extension. 
% Key clinical Text-to-SQL datasets differ in scope.
% MIMICSQL~\citep{wang2020mimicsql} pairs mostly auto-generated question-SQL pairs on MIMIC-III with crowdsourced filtering, targeting single-patient lookups or simple statistics and evaluated via exact-match or execution accuracy. EHRSQL~\citep{lee2022ehrsql} maps hospital-staff utterances to deterministic SQL over production schemas for patient-specific or small aggregate retrieval, with execution-based evaluation and no rubric for partially correct cohorts. The follow-on EHRSQL-ST~\citep{lee-etal-2024-overview} primarily strengthens data splits and contamination controls while preserving the same per-patient and aggregate tasks. EHR-SeqSQL~\citep{ryu-etal-2024-ehr} decomposes EHRSQL queries into sequential sub-queries to study interactive generation but still centers on patient-level or simple statistical requests rather than similarity-based cohorts. 
However, as shown in \autoref{tab:benchmarks_comparison}, 
% while existing clinical Text-to-SQL approaches have demonstrated the feasibility of automated clinical database querying through foundational EHR systems, 
existing clinical Text-to-SQL datasets~\citep{wang2020mimicsql, lee2022ehrsql, ryu-etal-2024-ehr, lee-etal-2024-overview, sivasubramaniam2024smtexttoquery, kim-etal-2024-ku} predominantly emphasize statistical analyses rather than addressing authentic clinical questions encountered in real-world practice, and consistently assume queries target specific patients with known identifiers, thereby representing only a limited subset of actual clinical analysis scenarios. 
Our work addresses these limitations by centering on patient-similarity cohort reasoning over MIMIC-IV v3.1, requiring models to define cohorts, apply temporal and phenotyping logic, and compute stratified cohort-level outcomes that mirror real clinical workflows.

\section{Benchmark Construction}
\begin{table*}[t]
\centering
\scriptsize
\begin{tabular}{p{0.35\textwidth}p{0.59\textwidth}}
\toprule
\textbf{Clinical Scenarios} & \textbf{Example Question Provided in Appendix~\ref{app:example}.} \\
\midrule
\textbf{Patient Demographics and Admissions} Analysis of patient demographics and administrative data (admissions, length of stay), testing foundational SQL skills and understanding of clinical administrative workflow. & 
For an 81‑year‑old female: among female Medicare patients aged 76–86 transferred from another hospital with principal AMI (ICD‑9 410*/ICD‑10 I21*), report 30‑day readmission rate; median index LOS for readmitted vs not; percent index stays $>4$ days.
\newline \textit{Complexity: Hard}
\hfill(Appendix ~\ref{app:e1})
\\
\midrule 
\textbf{Vital Signs Monitoring} Temporal analysis of vital signs (e.g., blood pressure, heart rate), designed to test time-series reasoning, understanding of clinical normal ranges, and trend identification capabilities. & 
I have a 60‑year‑old man in the ICU. In male ICU patients aged 55–65 with HFNC within 24 hours versus condition‑matched ICU controls, what are the instability score median and p25/p75/p95, tachycardia and hypotension burden, ICU LOS and mortality?
\newline \textit{Complexity: Medium}
\hfill(Appendix ~\ref{app:e2})
\\
\midrule 
\textbf{Laboratory Results Analysis} Analysis of trends in laboratory results, designed to test knowledge of medical terminology, unit conversions, and the ability to correlate lab values with clinical conditions. & 
I have a 51-year-old female with suspected ACS. Among female ACS admissions age 46–56, what are counts, percentages, and mean hospital length of stay for first hs‑TnT: Normal, Borderline, Myocardial Injury?
\newline \textit{Complexity: Medium}
\hfill(Appendix ~\ref{app:e3})
\\
\midrule 
\textbf{Medication Management} Analysis of medication regimens (prescriptions, dosing, interactions), designed to test complex temporal reasoning and pharmacological knowledge to ensure medication safety. & 
I have a 64‑year‑old female inpatient. Among females aged 59–69, what's the IQR of single inpatient amiodarone prescription durations (days)?
\newline \textit{Complexity: Easy}
\hfill(Appendix ~\ref{app:e4})
\\
\midrule 
\textbf{Diagnostic Procedures} Temporal sequencing of diagnostic procedures and interventions, designed to evaluate the understanding of clinical workflows, procedural relationships, and care coordination. & 
Evaluating an 88-year-old man: among male patients aged 83–93 with sepsis on their first ICU stay, stratify first‑72‑hour diagnostic intensity (distinct procedures) into quartiles and report mean procedure counts, mean ICU LOS in days, and mortality (\%) per quartile.
\newline \textit{Complexity: Hard}
\hfill(Appendix ~\ref{app:e5})
\\
\midrule 
\textbf{Disease Diagnosis and Outcomes} Analysis of diagnoses (ICD-9/10 codes), comorbidities, and clinical outcomes, designed to test knowledge of medical coding, integrated clinical reasoning, and the ability to assess treatment effectiveness and patient prognosis. & 
I have a 75-year-old female inpatient with pulmonary embolism. For female inpatients aged 70–80 with PE, stratify into risk-score quintiles and report per quintile: 90‑day mortality, general 70–80 female 90‑day mortality (comparison), AKI and ARDS rates, and median survivor LOS.
\newline \textit{Complexity: Hard}
\hfill(Appendix ~\ref{app:e6})
\\
\bottomrule
\end{tabular}
\caption{Definition of clinical scenario types in \ours.}
\label{tab:scenario_types}
\end{table*}
\ours is designed to comprehensively evaluate Text-to-SQL capabilities within realistic clinical scenarios. Our benchmark, built upon MIMIC-IV v3.1~\citep{johnson2023mimic}, incorporates complex analytical scenarios that require sophisticated clinical reasoning and the multi-step integration of diverse clinical data. \autoref{fig:figure1} provides an overview of our benchmark construction pipeline. \autoref{tab:scenario_types} presents the six core clinical scenarios that reflect real-world healthcare data analysis needs and clinical decision-making. Concrete scenario examples and rubric trees are provided in Appendix \ref{app:example}. In the following sections, we detail the query annotation, SQL annotation, evaluation guideline annotation, and data validation.
The expert annotator biographies are summarized in Appendix \ref{app:benchmark-construction}, and the annotation interface is shown in Appendix \ref{app:annotation-interface}.

\subsection{Query Annotation}
\paragraph{Clinical Scenario Development.} Each annotator is assigned one of six scenario types and selects a representative patient from MIMIC‑IV v3.1~\citep{johnson2023mimic}. Sampling is stratified across five dimensions: \textbf{(1) Age} uses scenario‑specific ranges spanning 25–85 years; \textbf{(2) Clinical condition} covers major categories (\eg cardiovascular, respiratory, metabolic, infectious, post‑operative); \textbf{(3) Healthcare utilization} varies admission type (\eg emergency, elective, urgent), insurance (\eg Medicare, Medicaid, commercial), and length of stay (\eg 2–15 days); \textbf{(4) Acuity} distinguishes settings (\eg ward vs ICU) with risk strata and monitoring intensity; and \textbf{(5) Temporal windows} include early windows (\eg first 24/48/72 hours), the full hospitalization, and procedure‑specific periods. To prevent data contamination and ensure benchmark integrity, all information related to the selected patients is removed from the database prior to model evaluation.

\paragraph{Natural Language Question Formulation.} Annotators craft natural language questions that physicians would realistically ask given the provided patient information and scenario type. Each question must require database querying and cannot be answered through simple observation (\eg questions requiring temporal analysis or aggregation across multiple records). Questions incorporate appropriate medical terminology while maintaining clarity and clinical authenticity.

\subsection{SQL Annotation}

Our SQL annotation process is divided into two steps: database analysis and schema mapping, and gold-standard SQL construction. Each clinical question undergoes comprehensive database analysis by annotators, followed by the development of gold-standard SQL to produce high-quality executable queries.

\paragraph{Database Analysis and Schema Mapping.} 

For each natural language clinical question, annotators first conduct a systematic database analysis to identify the required MIMIC tables and establish the necessary relationships between clinical entities. This process includes locating relevant database tables (e.g., patients, admissions, diagnoses\_icd), identifying key features including specific columns and clinical values (e.g., gender='M', icd\_code LIKE '410\%'), and mapping clinical concepts to database schema elements while considering temporal constraints and data integrity requirements. A concise schema reference for MIMIC-IV is provided in Appendix \ref{app:mimic4-schema}.

\begin{figure}[!t]
    \centering
    \includegraphics[width=\linewidth]{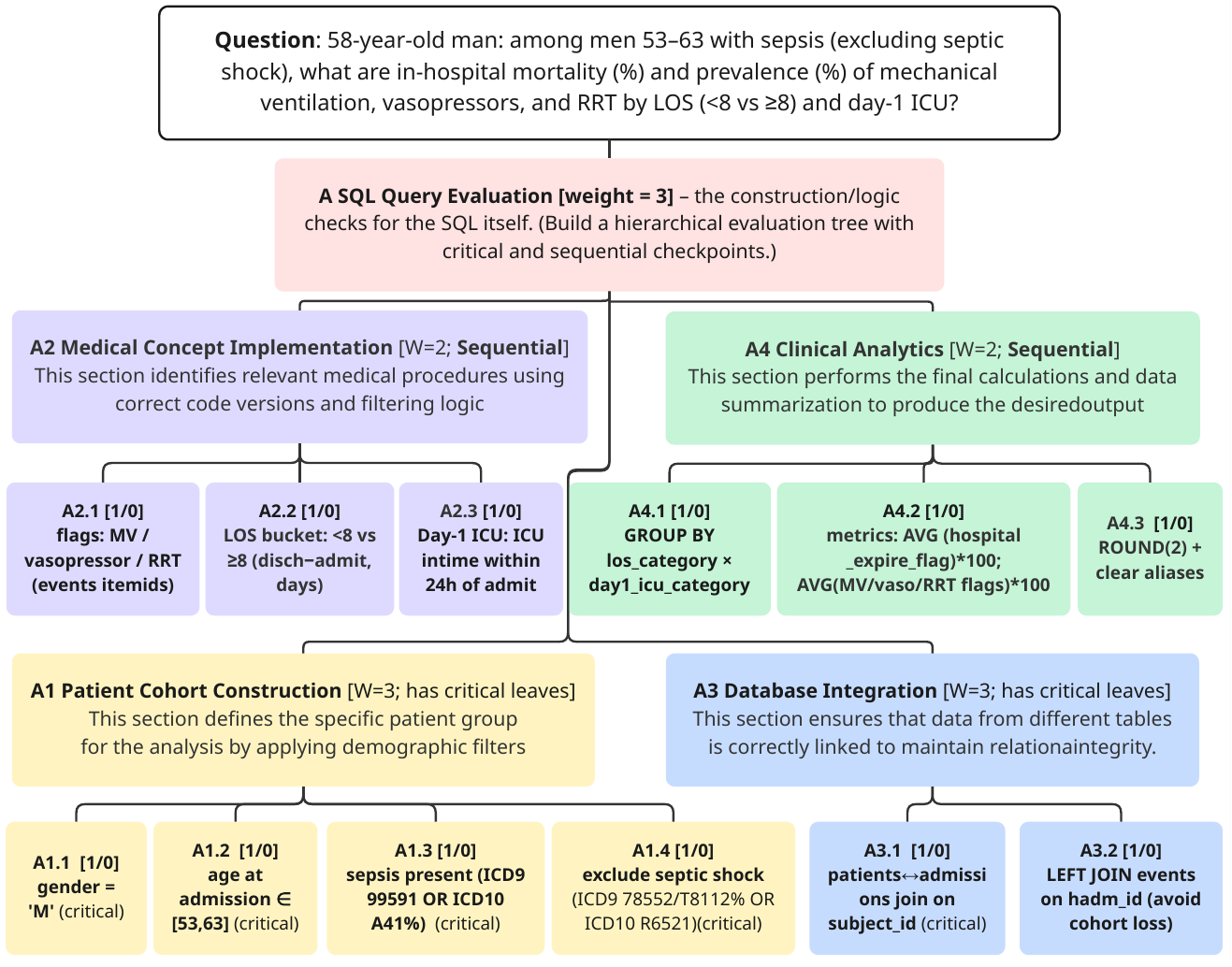}
    \caption{Example of a SQL evaluation rubric tree.}
    \label{fig:sql_rubric_tree}
\end{figure}

\begin{figure}[!t]
    \centering
    \includegraphics[width=\linewidth]{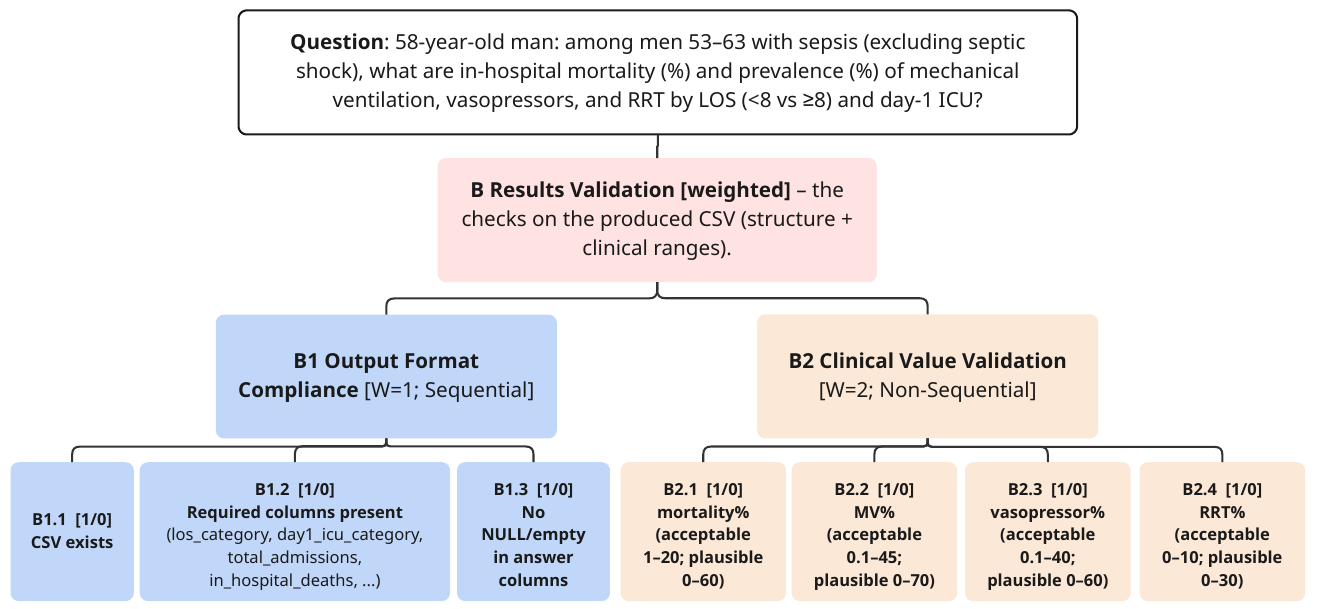}
    \caption{Example of an executed result rubric tree.}
    \label{fig:results_rubric_tree}
\end{figure}

\paragraph{Gold-Standard SQL Construction.}
Following the database analysis phase, expert annotators develop comprehensive gold-standard SQL queries that accurately translate clinical questions into executable database operations. Each SQL implementation undergoes rigorous development processes including multi-table join construction with proper foreign key relationships to ensure data consistency, temporal constraint implementation using appropriate date functions and time-based filtering (\eg DATE\_DIFF for length of stay calculations, charttime-based temporal analysis), clinical value range validation incorporating medical domain knowledge and normal physiological parameters (\eg hemoglobin levels between 7-18 g/dL, age calculations using anchor\_age and anchor\_year), edge case handling for common clinical database issues including null value management and data quality constraints, and query optimization to maintain computational efficiency while preserving clinical accuracy.

\subsection{Evaluation Guideline Annotation}
To support reliable automated evaluation, each clinical question is accompanied by a guideline comprising two rubric trees: one for SQL evaluation and one for executed results.

\begin{algorithm}[!t]
\caption{\textbf{Critical-First Score Aggregation}}
\label{alg:score-aggregation}
\small
\begin{algorithmic}
\Require rubric tree $T$, node weights $W$, critical flags $C$, sequential flags $S$
\Function{EvaluateNode}{$node$}
\If{$node$ is leaf}
\State \Return LLM judge evaluation score $\{0, 1\}$
\EndIf
\State $critical\_children \gets \{c \in children : C[c] = \text{true}\}$
\State $noncritical\_children \gets \{c \in children : C[c] = \text{false}\}$
\For{$c \in critical\_children$}
\State $score[c] \gets$ \Call{EvaluateNode}{$c$}
\If{$score[c] \neq 1$} \Return $0$ \EndIf
\EndFor
\If{$noncritical\_children = \emptyset$} \Return $1$ \EndIf
\State $sum \gets 0$, $total\_weight \gets 0$
\For{$c \in noncritical\_children$}
\State $score[c] \gets$ \Call{EvaluateNode}{$c$}
\State $sum \gets sum + W[c] \times score[c]$, $total\_weight \gets total\_weight + W[c]$
\If{$S[c] = \text{true} \land score[c] = 0$} \textbf{break} \EndIf
\EndFor
\State \Return $\frac{sum}{total\_weight}$
\EndFunction
\State $final\_score \gets$ \Call{EvaluateNode}{$root$}
\State \Return $final\_score$
\end{algorithmic}
\end{algorithm}

\paragraph{Rubric Design Principles.} Our evaluation framework employs tree-structured rubrics that hierarchically decompose complex evaluation tasks into granular, verifiable criteria~\citep{gou2025mind2web2evaluatingagentic}. Each rubric tree consists of internal nodes representing high-level evaluation aspects and leaf nodes defining specific binary verification criteria. The SQL evaluation rubric (\autoref{fig:sql_rubric_tree}) assesses query construction across four primary dimensions: Patient Cohort Construction, Medical Concept Implementation, Database Integration, and Clinical Analytics, while the results rubric (\autoref{fig:results_rubric_tree}) focuses on output validation and clinical value assessment. Following practices in automated evaluation~\citep{starace2025paperbenchevaluatingaisability}, we implement three key structural components: (1) \emph{Critical vs. Non-Critical Nodes}, where critical nodes represent essential requirements whose failure immediately causes parent failure, while non-critical nodes allow partial scoring; (2) \emph{Sequential Dependencies}, where sequential nodes indicate dependencies and earlier failures short-circuit subsequent evaluations; (3) \emph{Weighted Scoring}, where each node is assigned a weight from 1 to 3 based on importance: 1 denotes basic supportive criteria, 2 indicates standard requirements, and 3 marks critical elements that are essential to validity and require substantial domain expertise.

Our scoring system employs Critical-First Scoring adapted from recent agentic evaluation frameworks~\citep{gou2025mind2web2evaluatingagentic}, detailed in Algorithm~\ref{alg:score-aggregation}.

\begin{table}[!t]
\centering 
\setlength{\tabcolsep}{3pt}
\renewcommand{\arraystretch}{1.1}
\resizebox{\linewidth}{!}{
\begin{tabular}{lrrr}
\toprule
\textbf{Statistics} & \textbf{Easy} & \textbf{Med.} & \textbf{Hard} \\
\midrule
\textbf{Total Examples} & 190  & 254 & 189 \\
\midrule
Avg. Question Length & 22.86 & 36.04 & 45.52 \\
Avg. SQL Tokens & 158.79 & 452.53 & 615.62 \\
\textbf{Evaluation Trees} & & & \\
\quad SQL (Nodes/Depth) & 14.80 / 3.05 & 18.27 / 3.06 & 19.03 / 3.11 \\
\quad SQL Avg. Words & 15.55 & 16.97 & 17.74 \\
\quad Results (Nodes/Depth) & 10.01 / 4.00 & 19.62 / 4.02 & 23.54 / 4.04 \\
\quad Results Avg. Words & 6.52 & 8.21 & 7.98 \\
\bottomrule
\end{tabular}
}
\caption{Basic statistics of \ours.}
\label{tab:basic-stats}
\end{table}

\subsection{Data Validation}

Each annotated example undergoes comprehensive validation by an expert annotator within the medical research field. The validation framework examines four critical aspects: clinical question assessment evaluates real-world relevance, medical terminology accuracy, and linguistic quality; SQL implementation review verifies correct MIMIC-IV database~\citep{johnson2023mimic} standards and technical execution; output verification confirms structural integrity and medical plausibility; evaluation framework review ensures comprehensive component coverage and unambiguous scoring standards. Validators revise examples with minor issues or reject those with significant problems. Upon passing all validation checks, examples receive "Validated" status for dataset inclusion.
To assess problem difficulty and provide fine-grained evaluation of model capabilities, we stratify our dataset into three difficulty levels based on SQL complexity and clinical reasoning requirements: (1) \emph{Easy} (30\%): Few-table queries with basic filtering and clinical concepts; (2) \emph{Medium} (40\%): Multi-table joins with temporal filtering and moderate reasoning; (3) \emph{Hard} (30\%): Complex multi-join queries with nested subqueries and advanced logic. 
\autoref{tab:basic-stats} presents the data statistics of \ours.

\section{Evaluation Protocol}
\subsection{SQL Analysis using Evaluation Guideline}

SQL evaluation employs a rubric-driven process that decomposes each candidate query into clinically relevant checks. The guideline distinguishes critical from supportive requirements and encodes sequential dependencies such that subsequent reasoning is evaluated only if prerequisite steps are satisfied. Candidates must construct an appropriate cohort and map clinical concepts to schema and codes; subsequently, table relationships, join keys, type handling, grouping, and aggregation are verified, followed by task-specific interpretation. Leaf criteria receive binary decisions. Scores are aggregated with a critical-first rule: any failed critical node collapses its parent, whereas non-critical checks contribute via weighted averaging only after critical prerequisites are met. Consistency is ensured by employing GPT-5 as the judge, which leverages strong instruction following and long-context capacity to compare rubric text, gold SQL, and schema hints. Concise rationales are recorded for each leaf decision to support error analysis.

\begin{table*}[!t]
\centering
\setlength{\tabcolsep}{3pt}
\renewcommand\arraystretch{1.1}
\footnotesize
\resizebox{0.95\textwidth}{!}{%
\begin{tabular}{@{}l*{10}{>{\centering\arraybackslash}p{1.1cm}}@{}}
\toprule[1pt]
 & \multicolumn{6}{c}{\textbf{Test Set}} 
 & \multicolumn{2}{c}{\multirow{2}{*}[-0.6ex]{\textbf{Avg. Validation}}}
 & \multicolumn{2}{c}{\multirow{2}{*}[-0.6ex]{\textbf{Avg. Test}}} \\
\cmidrule(lr){2-7}\noalign{\vskip 1ex}
\multirow{2}{*}{\textbf{Model}} 
 & \multicolumn{2}{c}{\textbf{Easy}} 
 & \multicolumn{2}{c}{\textbf{Medium}} 
 & \multicolumn{2}{c}{\textbf{Hard}} 
 &  &  &  &  \\
\noalign{\vskip 0.5ex}
 & \textbf{SQL} & \textbf{Exec} & \textbf{SQL} & \textbf{Exec} & \textbf{SQL} & \textbf{Exec} & \textbf{SQL} & \textbf{Exec} & \textbf{SQL} & \textbf{Exec} \\
\midrule[0.7pt]
\multicolumn{11}{c}{\textbf{\textit{Proprietary Models}}} \\
\noalign{\vskip 1ex}
% Sorted by Avg. Test Exec (desc)
GPT-5-mini & 54.07 & \cellcolor{red!20}{81.31} & 37.12 & \cellcolor{red!35}{73.46} & 38.94 & \cellcolor{red!35}{69.69} & 42.30 & \cellcolor{red!20}{75.16} & 42.72 & \cellcolor{red!35}{74.67} \\
Gemini-2.5-Pro & 53.15 & \cellcolor{red!35}{85.46} & \cellcolor{red!35}{46.34} & \cellcolor{red!20}{69.89} & \cellcolor{red!35}{42.71} & \cellcolor{red!20}{67.22} & \cellcolor{red!20}{45.31} & \cellcolor{red!35}{76.14} & \cellcolor{red!20}{47.28} & \cellcolor{red!20}{73.73} \\
GPT-5 & \cellcolor{red!20}{58.56} & \cellcolor{red!5}{73.79} & 41.29 & 66.94 & \cellcolor{red!5}{39.58} & \cellcolor{red!5}{65.08} & 42.62 & 66.52 & 45.93 & \cellcolor{red!5}{68.42} \\
GPT-4.1 & \cellcolor{red!35}{59.54} & 71.20 & \cellcolor{red!5}{42.30} & \cellcolor{red!5}{68.55} & 38.25 & 63.39 & \cellcolor{red!5}{44.69} & \cellcolor{red!5}{69.92} & \cellcolor{red!5}{46.23} & 67.79 \\
Gemini-2.5-Flash & \cellcolor{red!5}{55.86} & 72.64 & \cellcolor{red!20}{45.61} & 64.05 & \cellcolor{red!20}{41.67} & 58.72 & \cellcolor{red!35}{47.06} & 66.75 & \cellcolor{red!35}{47.48} & 65.01 \\
OpenAI o4-mini & 54.62 & 67.96 & 36.59 & 58.37 & 34.13 & 51.68 & 39.41 & 59.07 & 41.23 & 59.22 \\
Grok-4-Fast-Reason. & 49.47 & 70.82 & 40.34 & 53.18 & 39.40 & 53.23 & 42.67 & 56.58 & 42.78 & 58.46 \\
GPT-5-nano & 40.83 & 55.42 & 34.11 & 54.67 & 34.17 & 45.14 & 33.18 & 51.58 & 36.13 & 52.03 \\
Mistral-Medium & 42.61 & 56.05 & 32.81 & 42.50 & 31.47 & 37.68 & 32.71 & 43.81 & 35.33 & 45.10 \\
Grok-4-Fast-Non-Reason. & 54.16 & 39.31 & 32.33 & 29.80 & 35.69 & 22.39 & 34.77 & 30.10 & 39.85 & 30.41 \\
\midrule
\multicolumn{11}{c}{\textbf{\textit{Open-source Models}}} \\
\noalign{\vskip 1ex}
% Sorted by Avg. Test Exec (desc)
DeepSeek-R1 & {45.32} & \cellcolor{red!35}{75.73} & \cellcolor{red!35}{45.91} & \cellcolor{red!35}{68.43} & \cellcolor{red!35}{43.16} & \cellcolor{red!35}{63.59} & \cellcolor{red!35}{42.63} & \cellcolor{red!35}{69.79} & \cellcolor{red!35}{44.91} & \cellcolor{red!35}{69.15} \\
DeepSeek-V3.1 & \cellcolor{red!35}{57.54} & \cellcolor{red!5}{71.63} & \cellcolor{red!5}{38.78} & \cellcolor{red!20}{58.38} & \cellcolor{red!20}{34.83} & \cellcolor{red!5}{52.99} & \cellcolor{red!20}{38.90} & \cellcolor{red!20}{61.46} & \cellcolor{red!20}{43.19} & \cellcolor{red!20}{60.71} \\
% --- Qwen model family ---
Qwen3-235B-A22B-Ins. & 38.14 & \cellcolor{red!20}{71.85} & 37.72 & 54.48 & 32.77 & 51.05 & 36.24 & 58.15 & 36.36 & \cellcolor{red!5}{58.63} \\
Qwen3-Coder-480B-A35B-Ins. & \cellcolor{red!20}{48.97} & {64.58} & {36.00} & \cellcolor{red!5}{56.04} & 33.27 & \cellcolor{red!20}{54.69} & {35.54} & \cellcolor{red!5}{60.51} & \cellcolor{red!5}{39.05} & {58.18} \\
Qwen3-Next-80B-A3B-Ins. & \cellcolor{red!5}48.34 & 67.85 & 29.93 & 45.83 & 26.81 & 35.38 & 34.48 & 43.41 & 34.48 & 49.26 \\
Qwen3-235B-A22B-Think. & 34.71 & 55.51 & \cellcolor{red!20}{43.41} & 44.72 & \cellcolor{red!5}{34.71} & 46.67 & \cellcolor{red!5}{38.08} & 51.11 & 38.20 & 48.54 \\
Llama-4-Maverick-17B-128E-Ins. & 39.82 & 59.14 & 27.23 & 47.42 & 20.98 & 36.01 & 29.36 & 51.63 & 29.11 & 47.49 \\
Llama-4-Scout-17B-16E-Ins. & 40.79 & 36.57 & 24.04 & 28.69 & 20.23 & 26.55 & 26.88 & 31.44 & 27.89 & 30.40 \\
Qwen3-Next-80B-A3B-Think. & 46.07 & 27.73 & 37.30 & 30.66 & 31.25 & 21.36 & 37.77 & 29.06 & 38.13 & 27.01 \\
% --- Additional open-source models ---
Baichuan-M2-32B & 36.09 & 23.10 & 30.44 & 13.24 & 23.17 & 10.11 & 26.60 & 11.40 & 29.97 & 15.27 \\
MedGemma-27B & 32.66 & 6.52 & 16.60 & 3.12 & 14.92 & 2.65 & 21.03 & 4.46 & 20.92 & 4.00 \\
SQLCoder-7B-2 & 6.65 & 0.00 & 6.50 & 0.00 & 2.30 & 0.00 & 3.99 & 0.00 & 5.29 & 0.00 \\
\bottomrule[1pt]
\end{tabular}%
}
\caption{SQL score and execution score (\%) on \ours{} validation and test sets using CoT prompting with self-refinement. 
% Test Set columns report Easy/Medium/Hard difficulty splits; Avg. columns report set-level averages. 
Scenario-level scores are presented in Appendix~\ref{app:scenario-results}.}
\label{tab:main_results}
\end{table*}

\subsection{SQL Execution Result Evaluation}

The execution-level evaluation examines CSV outputs produced by executing the candidate SQL. Format compliance is first enforced, including file presence, exact column names, absence of nulls, and basic type checks; clinical plausibility is then assessed using per-column value ranges grounded in the cohort and task definition. Plausible bands admit clinically equivalent answers, whereas acceptable bands tighten tolerance; these criteria support equivalence across alternative but valid formulations. Sequential gating prevents downstream clinical judgments from obscuring upstream format defects. Leaf decisions are binary and are aggregated under the same critical-first rule. The same GPT-5 judge issues decisions and concise rationales by comparing the CSV with the rubric and gold references, yielding interpretable discrepancies in schema adherence, unit handling, rounding, and cohort-conditioned statistics.

The LLM-as-Judge prompts are provided in Appendix \ref{app:judge-prompt}. We provide the reliability and reconciliation analyses of our proposed rubric-based evaluation in Appendix \ref{app:human-gpt-agreement} and \ref{app:iaa-reconciliation}, with SQL-execution divergence statistics summarized in Appendix \ref{app:sql-exec-reconciliation}.
\section{Experiment}
This section discusses the experiment setup and our experiment results and analysis.

\subsection{Experiment Setup}
We evaluate all models on \ours using rubric-based metrics specifically designed for clinical text-to-SQL tasks. Our primary evaluation metrics are the \textbf{SQL Score} and the \textbf{Execution Score}.
We consider the following categories of models:
(1) \textbf{Open-source general-purpose LLMs:}
DeepSeek-R1~\cite{deepseekai2025deepseekr1incentivizingreasoningcapability},
DeepSeek-V3.1~\cite{deepseekai2025deepseekV3technicalreport},
Qwen3-Coder series,
Qwen3-Instruct series,
Qwen3-Thinking series~\cite{qwen3techreport2025},
Llama-4 series~\cite{meta2024llama4}.
(2) \textbf{Proprietary models:}
GPT-5 series~\cite{openai2025gpt5},
Gemini-2.5 series~\cite{comanici2025gemini25pushingfrontier},
GPT-4.1~\cite{openai2025gpt41},
o4-mini~\cite{openai2025o4mini},
Grok-4-Fast series~\cite{xai2025grok4fast,xai2025grok4fastnonreasoning}, and
Mistral-Medium~\cite{mistralai2025mistralmedium}.
(3) \textbf{Text-to-SQL models:}
SQLCoder-7B-2~\cite{defog2024sqlcoder7b2}.
(4) \textbf{Medical-domain LLMs:}
MedGemma-27B~\cite{medgemma27btextit} and
Baichuan-M2-32B~\cite{baichuanm232b}.
For open-source models, we perform inference using \texttt{vLLM} pipeline~\cite{kwon2023efficientmemorymanagementlarge}, while proprietary models are accessed through official APIs.

We evaluate models under two prompting regimes: Direct Output and Chain-of-Thought (CoT). In both regimes, the model must return a single executable BigQuery query. If execution fails, we run up to two \emph{self-refinement} rounds that feed the question, the prior SQL, and the BigQuery error back to the model with minimal-edit instructions. We then extract the final fenced SQL block and execute it. We apply \emph{self-refinement} to both regimes because many models have low first-pass execution success, and a single correction round is insufficient; Appendix \ref{app:execution-success-rates} reports the attempt-wise success rates that motivate this choice. Prompt templates for both regimes and the refinement procedure are provided in Appendix \ref{app:model-prompts}. Parameter settings and model configurations appear in Appendix \ref{app:model-configs}.

\subsection{Main Findings}
\autoref{tab:main_results} presents SQL and execution scores on \ours. We highlight the following findings:

\paragraph{\ours presents substantial challenges for current foundation models.}
While GPT-5-mini achieves the best average execution score, performance on the Hard split remains modest: leading proprietary models stay under 70\% (\eg GPT-5-mini 69.7\% and Gemini-2.5-Pro 67.2\%). Gemini-2.5-Pro also drops by 18.24\% from Easy to Hard. 
% These results show that many strong proprietary models still underperform on complex settings, underscoring that \ours is essential for evaluating database-grounded clinical text-to-SQL reasoning.

\paragraph{Open-sourced models performance.}
DeepSeek-R1 attains 69.2\% average test execution with a 44.9\% SQL score, and it surpasses several proprietary baselines, including o4-mini and both Grok-4 variants. However, open-source models still lag the strongest proprietary models: the best proprietary model (GPT-5-mini) reaches 74.7\% execution, about 5.5 points higher than DeepSeek-R1 and roughly 14 to 16 points ahead of DeepSeek-V3.1 (60.7\%) and Qwen3-Coder-480B-A35B-Instruct (58.2\%). Even so, these results show the gap is narrowing as techniques mature.

\begin{figure}[t]
  \hfill
  \begin{minipage}{0.48\textwidth}
    \centering
    \includegraphics[width=\linewidth,trim=2mm 2mm 2mm 2mm,clip]{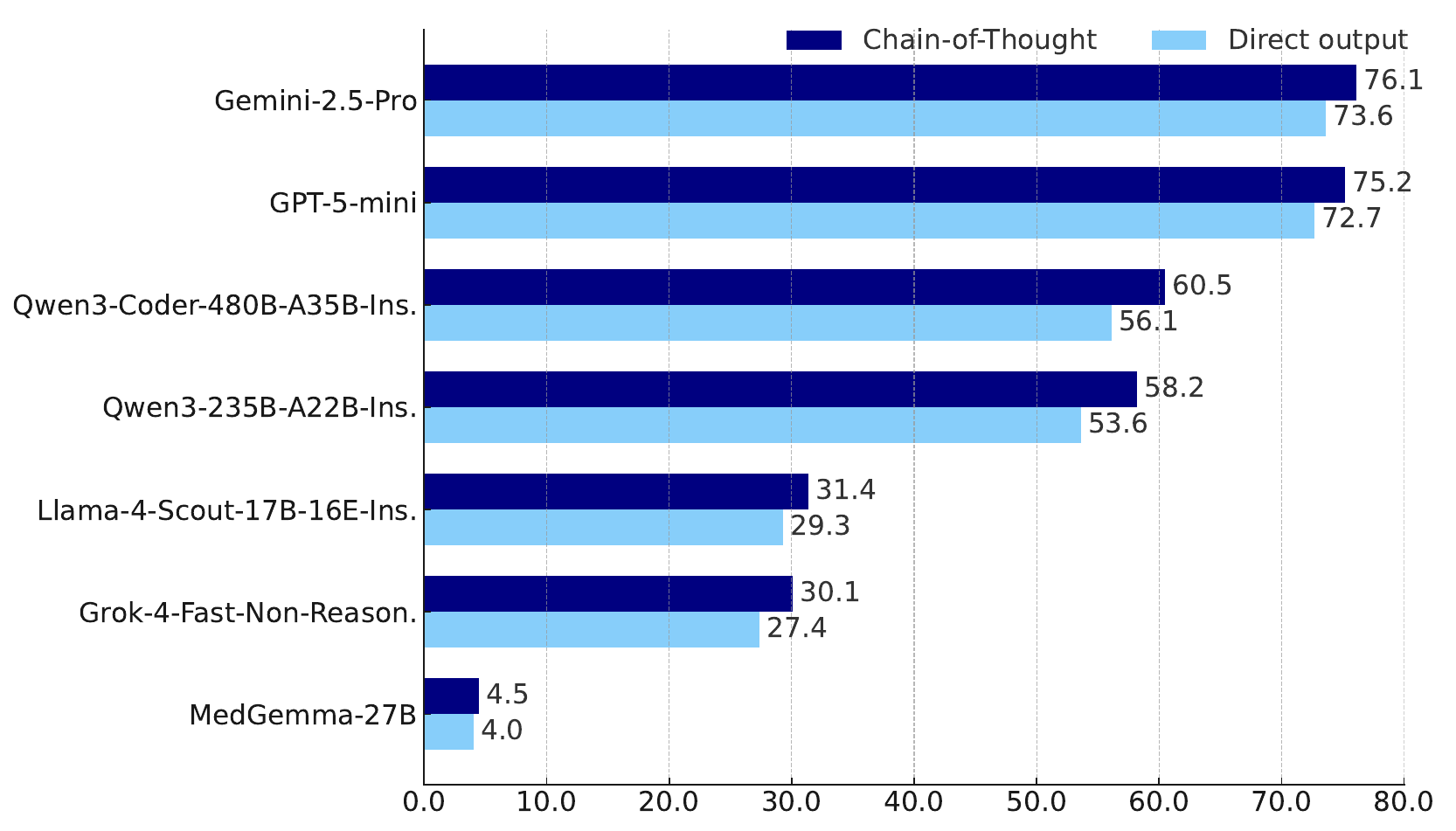}
    \caption{Execution score comparison on the validation set for representative models. Full SQL and execution comparisons for all models are provided in \autoref{app:validation-sql-comparison}.}
    \label{fig:exec_cot_compare_main}
  \end{minipage}
\end{figure}

\paragraph{CoT reasoning generally improves model performance compared to directly outputting the final SQL.}
As shown in \autoref{fig:exec_cot_compare_main}, the extent of improvement varies across models. Qwen3-235B-A22B-Instruct increases from 53.6\% to 58.2\% and Qwen3-Coder-480B-A35B-Instruct from 56.1\% to 60.5\%, while Llama-4-Scout-17B-16E-Instruct rises only from 29.3\% to 31.4\% and Grok-4-Fast-Non-Reasoning from 27.4\% to 30.1\%. Strong proprietary models see modest gains (Gemini-2.5-Pro from 73.6\% to 76.1\%, GPT-5-mini from 72.7\% to 75.2\%). Even lower baselines benefit (MedGemma-27B from 4.0\% to 4.5\%). 

\subsection{Error Analysis and Case Study}
To better characterize failure modes, we randomly select 10 cases from each of six validation scenarios generated by GPT-5-mini and analyze rubric feedback from our judge model. We observe three common error types: \textbf{Cohort specification \& coding} (54\%): in which explicit ICD or \texttt{itemid} constraints are replaced by keyword heuristics or key joins are relaxed, broadening cohorts; \textbf{Output schema \& formatting} (24\%): omitted required columns, invalid values, or naming mismatches that trigger schema checks; and \textbf{Aggregation \& clinical statistics} (14\%): mis-specified denominators or missing normalization leading to implausible rates. \textbf{Other observed errors} include occasional temporal boundary mistakes and pattern-specific issues seen in GPT-5-mini outputs, such as using quartiles instead of percentiles in \texttt{APPROX\_QUANTILES} (\eg employing quartile buckets rather than 100-quantile offsets), which yields incorrect reported statistics. Examples for each error type are provided in \autoref{app:err}.

\begin{table}[t]
\centering
\small
\setlength{\tabcolsep}{4pt}
\renewcommand{\arraystretch}{1.1}
\resizebox{0.4\textwidth}{!}{%
\begin{tabular}{llll}
\toprule
\textbf{Setting} & \multicolumn{1}{c}{\textbf{Easy}} & \multicolumn{1}{c}{\textbf{Medium}} & \multicolumn{1}{c}{\textbf{Hard}} \\
\midrule
\multicolumn{4}{l}{\textbf{SQL Score}} \\
\quad Baseline CoT & 54.78 & 36.40 & 37.90 \\
\quad Schema-hinted CoT & 58.27\textsuperscript{$\uparrow$3.49} & 41.09\textsuperscript{$\uparrow$4.69} & 42.04\textsuperscript{$\uparrow$4.14} \\
\midrule
\multicolumn{4}{l}{\textbf{Execution Score}} \\
\quad Baseline CoT & 79.96 & 75.63 & 69.83 \\
\quad Schema-hinted CoT & 82.74\textsuperscript{$\uparrow$2.78} & 85.85\textsuperscript{$\uparrow$10.21} & 77.03\textsuperscript{$\uparrow$7.20} \\
\bottomrule
\end{tabular}
}
\caption{Validation performance of GPT-5-mini under baseline and \emph{schema-hinted} Chain-of-Thought configurations. Superscripts denote absolute percentage-point gains of the schema-hinted setting over the baseline.}
\label{tab:rubric_guided_difficulty}
\end{table}

\subsection{Schema-Hinted Inference Analysis} The preceding failure analysis underscores cohort drift and schema mismatches as dominant error sources. This motivates our exploration of a Schema-Hinted inference configuration designed to mitigate these common failures. The setting augments the standard Chain-of-Thought prompting and self-refinement schedule with schema hints, foregrounding clinically validated ICD filters and expected result columns. Full setup details are provided in Appendix \ref{app:schema-hinted-setup}.
We evaluate this configuration with GPT-5-mini on the validation split, and the results show consistent gains over the baseline. As summarised in \autoref{tab:rubric_guided_difficulty}, SQL and execution accuracy improve across all difficulty tiers. The most pronounced execution gains are observed on medium and hard queries, where providing clinically validated ICD filters and expected result columns better constrains the inference process.

\section{Conclusion}
We propose \ours, a benchmark for realistic clinical text-to-SQL analytics. It captures core challenges of real EHR practice, including heterogeneous tables, temporal windows, and patient-similarity cohort construction. We assess a broad set of models using the developed rubric-based evaluation protocols and observe that, despite recent advances, performance remains well short of clinically reliable operation: execution frequently exceeds SQL correctness and errors cluster around cohort specification, schema/formatting, and aggregation/clinical statistics. \ours establishes a rigorous, domain-grounded target for clinical research and advance trustworthy EHR analytics.

\section*{Limitations}
While \ours advances clinically grounded text-to-SQL evaluation, several limitations remain. First, the current benchmark is built on MIMIC-IV v3.1—data from a single health system—and targets a single SQL environment, which may limit transferability to other EHR ecosystems, data models, and database backends. Second, \ours depends on substantial domain-expert involvement for scenario specification, gold-standard SQL authoring, and rubric-aligned leaf rationales. Although this expert curation provides high-fidelity supervision, the associated training, annotation, and review burden reduces throughput and makes it difficult to scale to substantially larger datasets without additional tooling or alternative supervision strategies.

\bibliography{anthology,custom}

@article{johnson2023mimic,
  title={MIMIC-IV, a freely accessible electronic health record dataset},
  author={Johnson, Alistair EW and Bulgarelli, Lucas and Shen, Lu and Gayles, Alvin and Shammout, Ayad and Horng, Steven and Pollard, Tom J and Hao, Sicheng and Moody, Benjamin and Gow, Brian and others},
  journal={Scientific Data},
  volume={10},
  number={1},
  pages={1--9},
  year={2023},
  publisher={Nature Publishing Group}
}

@article{jin2021medqa,
  title={What Disease does this Patient Have? A Large-scale Open Domain Question Answering Dataset from Medical Exams},
  author={Jin, Di and Pan, Eileen and Oufattole, Nassim and Weng, Wei-Hung and Fang, Hanyi and Szolovits, Peter},
  journal={Applied Sciences},
  volume={11},
  number={14},
  pages={6421},
  year={2021}
}

@article{jin2019pubmedqa,
  title={PubMedQA: A Dataset for Biomedical Research Question Answering},
  author={Jin, Qiao and Dhingra, Bhuwan and Liu, Zhengping and Cohen, William and Lu, Xinghua},
  journal={Proceedings of the 2019 Conference on Empirical Methods in Natural Language Processing and the 9th International Joint Conference on Natural Language Processing (EMNLP-IJCNLP)},
  pages={2567--2577},
  year={2019}
}

@article{pal2022medmcqa,
  title={MedMCQA: A Large-scale Multi-Subject Multi-Choice Dataset for Medical domain Question Answering},
  author={Pal, Ankit and Umapathi, Logesh Kumar and Sankarasubbu, Malaikannan},
  journal={Proceedings of the Conference on Health, Inference, and Learning},
  pages={248--260},
  year={2022}
}

@inproceedings{hendrycks2020measuring,
  title={Measuring Massive Multitask Language Understanding},
  author={Hendrycks, Dan and Burns, Collin and Basart, Steven and Zou, Andy and Mazeika, Mantas and Song, Dawn and Steinhardt, Jacob},
  booktitle={International Conference on Learning Representations},
  year={2021}
}

@article{kweon2024ehrnoteqa,
  title={EHRNoteQA: An LLM Benchmark for Real-World Clinical Practice Using Discharge Summaries},
  author={Kweon, Sunjun and Lee, Jiyoun and Yi, Joon Sung and Choi, Edward},
  journal={arXiv preprint arXiv:2402.16040},
  year={2024}
}

@inproceedings{wang2022drugehrqa,
  title={DrugEHRQA: A Question Answering Dataset on Structured and Unstructured Electronic Health Records},
  author={Wang, Logesh Kumar Umapathi and Pal, Ankit and Sankarasubbu, Malaikannan},
  booktitle={Proceedings of the Language Resources and Evaluation Conference},
  pages={2789--2797},
  year={2022}
}

@article{zhang2025medxpertqa,
  title={MedXpertQA: Benchmarking Expert-Level Medical Reasoning and Understanding},
  author={Zhang, Yifan and Chen, Jingqing and Yuan, Qiao and Ding, Zhihao and Luo, Huaishao and Pan, Kaixiong and Zhang, Mengzhuo and Yu, Haiyang and Chen, Qingyun and Wang, Xiangru and others},
  journal={arXiv preprint arXiv:2501.18362},
  year={2025}
}

@article{bae2023ehrxqa,
  title={EHRXQA: A Multi-Modal Question Answering Dataset for Electronic Health Records with Chest X-ray Images},
  author={Bae, Seongsu and Kweon, Sunjun and Jang, Taewoong and Choi, Edward},
  journal={arXiv preprint arXiv:2310.18652},
  year={2023}
}

@inproceedings{soni2022radqa,
  title={RadQA: A Question Answering Dataset to Improve Comprehension of Radiology Reports},
  author={Soni, Siddharth and Roberts, Kirk and Wang, Daqing and Subburathinam, Arvind and Long, Sivaramakrishnan},
  booktitle={Proceedings of the Language Resources and Evaluation Conference},
  pages={6567--6577},
  year={2022}
}

@article{pampari2018emrqa,
  title={emrQA: A Large Corpus for Question Answering on Electronic Medical Records},
  author={Pampari, Anusri and Raghavan, Preethi and Liang, Jennifer and Peng, Jian},
  journal={Proceedings of the 2018 Conference on Empirical Methods in Natural Language Processing},
  pages={2357--2368},
  year={2018}
}

@inproceedings{Suominen2020ClefEhealth,
  author    = {Suominen, Hanna and Kelly, Liadh and Goeuriot, Lorraine and Krallinger, Martin},
  title     = {CLEF eHealth Evaluation Lab 2020},
  booktitle = {Advances in Information Retrieval: 42nd European Conference on Information Retrieval (ECIR 2020), Part II},
  series    = {Lecture Notes in Computer Science},
  volume    = {12036},
  pages     = {587--594},
  publisher = {Springer},
  year      = {2020},
  doi       = {10.1007/978-3-030-45442-5_76},
  url       = {https://pmc.ncbi.nlm.nih.gov/articles/PMC7148004/}
}

@inproceedings{Saleh2019ClefEhealthCLIR,
  author    = {Shadi Saleh and Pavel Pecina},
  title     = {An Extended {CLEF} eHealth Test Collection for Cross-Lingual
               Information Retrieval in the Medical Domain},
  booktitle = {Advances in Information Retrieval: 41st European Conference on Information
               Retrieval (ECIR 2019)},
  series    = {Lecture Notes in Computer Science},
  volume    = {11438},
  pages     = {188--195},
  publisher = {Springer},
  year      = {2019},
  doi       = {10.1007/978-3-030-15719-7_24}
}

@inproceedings{lee2022ehrsql,
  title={EHRSQL: A Practical Text-to-SQL Benchmark for Electronic Health Records},
  author={Lee, Gyubok and Hwang, Hyeonji and Bae, Seongsu and Kwon, Yeonsu and Shin, Woncheol and Yang, Seongjun and Seo, Minjoon and Kim, Jong-Yeup and Choi, Edward},
  booktitle={Thirty-sixth Conference on Neural Information Processing Systems Datasets and Benchmarks Track},
  year={2022}
}

@inproceedings{wang2020mimicsql,
  title={MIMICSQL: Text-to-SQL Generation for Question Answering on Electronic Medical Records},
  author={Wang, Xiaoxuan and Kapanipathi, Pavan and Musa, Ryan and Yu, Mo and Talamadupula, Kartik and Abdelaziz, Ibrahim and Chang, Maria and Fokoue, Achille and Makni, Bassem and Mattei, Nicholas and others},
  booktitle={Proceedings of The Web Conference 2020},
  pages={2506--2516},
  year={2020}
}

@inproceedings{yu2018spider,
  title={Spider: A Large-Scale Human-Labeled Dataset for Complex and Cross-Domain Semantic Parsing and Text-to-SQL Task},
  author={Yu, Tao and Zhang, Rui and Yang, Kai and Yasunaga, Michihiko and Wang, Dongxu and Li, Zifan and Ma, James and Li, Irene and Yao, Qingning and Roman, Shanelle and others},
  booktitle={Proceedings of the 2018 Conference on Empirical Methods in Natural Language Processing},
  pages={3911--3921},
  year={2018}
}

@article{wei2024spider,
  title={Spider 2.0: Evaluating Language Models on Real-World Enterprise Text-to-SQL Workflows},
  author={Wei, Fangyu and Chen, Jixuan and Ye, Yuxiao and Cao, Ruisheng and Shin, Dongchan and Su, Hongjin and Suo, Zhaoqing and Gao, Hongcheng and Hu, Wenjing and Yin, Pengcheng and others},
  journal={arXiv preprint arXiv:2411.07763},
  year={2024}
}

@article{li2023bird,
  title={Can LLM Already Serve as a Database Interface? A BIg Bench for LaRge-scale Database Grounded Text-to-SQL Evaluation},
  author={Li, Jinyang and Hui, Binyuan and Qu, Ge and Yang, Jiaxi and Li, Binhua and Li, Bowen and Wang, Bailin and Qin, Bowen and Geng, Ruiying and Huo, Nan and others},
  journal={Advances in Neural Information Processing Systems},
  volume={36},
  year={2023}
}

@article{white2024livebench,
  title={LiveBench: A Challenging, Contamination-Limited LLM Benchmark},
  author={White, Colin and Dooley, Samuel and Roberts, Manley and Pal, Arka and Feuer, Ben and Jain, Siddhartha and Shwartz-Ziv, Ravid and Jain, Neel and Saifullah, Khalid and Naidu, Siddartha and others},
  journal={arXiv preprint arXiv:2406.19314},
  year={2024}
}

@inproceedings{zhong2017seq2sql,
  title={Seq2SQL: Generating Structured Queries from Natural Language using Reinforcement Learning},
  author={Zhong, Victor and Xiong, Caiming and Socher, Richard},
  booktitle={arXiv preprint arXiv:1709.00103},
  year={2017}
}

@inproceedings{lee2021kaggledbqa,
  title={KaggleDBQA: Realistic Evaluation of Text-to-SQL Parsers},
  author={Lee, Chia-Hsuan and Polozov, Oleksandr and Richardson, Matthew},
  booktitle={Proceedings of the 59th Annual Meeting of the Association for Computational Linguistics and the 11th International Joint Conference on Natural Language Processing (Volume 1: Long Papers)},
  pages={2261--2273},
  year={2021}
}

@inproceedings{
wang2024mmlupro,
title={{MMLU}-Pro: A More Robust and Challenging Multi-Task Language Understanding Benchmark},
author={Yubo Wang and Xueguang Ma and Ge Zhang and Yuansheng Ni and Abhranil Chandra and Shiguang Guo and Weiming Ren and Aaran Arulraj and Xuan He and Ziyan Jiang and Tianle Li and Max Ku and Kai Wang and Alex Zhuang and Rongqi Fan and Xiang Yue and Wenhu Chen},
booktitle={The Thirty-eight Conference on Neural Information Processing Systems Datasets and Benchmarks Track},
year={2024},
url={https://openreview.net/forum?id=y10DM6R2r3}
}

@misc{ou2025experienceretrievalaugmentationelectronichealth,
      title={Experience Retrieval-Augmentation with Electronic Health Records Enables Accurate Discharge QA}, 
      author={Justice Ou and Tinglin Huang and Yilun Zhao and Ziyang Yu and Peiqing Lu and Rex Ying},
      year={2025},
      eprint={2503.17933},
      archivePrefix={arXiv},
      primaryClass={cs.CL},
      url={https://arxiv.org/abs/2503.17933}, 
}

@misc{singhal2022largelanguagemodelsencode,
      title={Large Language Models Encode Clinical Knowledge}, 
      author={Karan Singhal and Shekoofeh Azizi and Tao Tu and S. Sara Mahdavi and Jason Wei and Hyung Won Chung and Nathan Scales and Ajay Tanwani and Heather Cole-Lewis and Stephen Pfohl and Perry Payne and Martin Seneviratne and Paul Gamble and Chris Kelly and Nathaneal Scharli and Aakanksha Chowdhery and Philip Mansfield and Blaise Aguera y Arcas and Dale Webster and Greg S. Corrado and Yossi Matias and Katherine Chou and Juraj Gottweis and Nenad Tomasev and Yun Liu and Alvin Rajkomar and Joelle Barral and Christopher Semturs and Alan Karthikesalingam and Vivek Natarajan},
      year={2022},
      eprint={2212.13138},
      archivePrefix={arXiv},
      primaryClass={cs.CL},
      url={https://arxiv.org/abs/2212.13138}, 
}

@inproceedings{chen-etal-2025-benchmarking,
    title = "Benchmarking Large Language Models on Answering and Explaining Challenging Medical Questions",
    author = "Chen, Hanjie  and
      Fang, Zhouxiang  and
      Singla, Yash  and
      Dredze, Mark",
    editor = "Chiruzzo, Luis  and
      Ritter, Alan  and
      Wang, Lu",
    booktitle = "Proceedings of the 2025 Conference of the Nations of the Americas Chapter of the Association for Computational Linguistics: Human Language Technologies (Volume 1: Long Papers)",
    month = apr,
    year = "2025",
    address = "Albuquerque, New Mexico",
    publisher = "Association for Computational Linguistics",
    url = "https://aclanthology.org/2025.naacl-long.182/",
    doi = "10.18653/v1/2025.naacl-long.182",
    pages = "3563--3599",
    ISBN = "979-8-89176-189-6"
}

@inproceedings{
sivasubramaniam2024smtexttoquery,
title={{SM}3-Text-to-Query: Synthetic Multi-Model Medical Text-to-Query Benchmark},
author={Sithursan Sivasubramaniam and Cedric Osei-Akoto and Yi Zhang and Kurt Stockinger and Jonathan Fuerst},
booktitle={The Thirty-eight Conference on Neural Information Processing Systems Datasets and Benchmarks Track},
year={2024},
url={https://openreview.net/forum?id=Pm0UzCehgB}
}

@inproceedings{livebench,
  title={LiveBench: A Challenging, Contamination-Free {LLM} Benchmark},
  author={Colin White and Samuel Dooley and Manley Roberts and Arka Pal and Benjamin Feuer and Siddhartha Jain and Ravid Shwartz-Ziv and Neel Jain and Khalid Saifullah and Sreemanti Dey and Shubh-Agrawal and Sandeep Singh Sandha and Siddartha Venkat Naidu and Chinmay Hegde and Yann LeCun and Tom Goldstein and Willie Neiswanger and Micah Goldblum},
  booktitle={The Thirteenth International Conference on Learning Representations},
  year={2025},
}

@misc{gou2025mind2web2evaluatingagentic,
      title={Mind2Web 2: Evaluating Agentic Search with Agent-as-a-Judge}, 
      author={Boyu Gou and Zanming Huang and Yuting Ning and Yu Gu and Michael Lin and Weijian Qi and Andrei Kopanev and Botao Yu and Bernal Jiménez Gutiérrez and Yiheng Shu and Chan Hee Song and Jiaman Wu and Shijie Chen and Hanane Nour Moussa and Tianshu Zhang and Jian Xie and Yifei Li and Tianci Xue and Zeyi Liao and Kai Zhang and Boyuan Zheng and Zhaowei Cai and Viktor Rozgic and Morteza Ziyadi and Huan Sun and Yu Su},
      year={2025},
      eprint={2506.21506},
      archivePrefix={arXiv},
      primaryClass={cs.AI},
      url={https://arxiv.org/abs/2506.21506}, 
}

@misc{starace2025paperbenchevaluatingaisability,
      title={PaperBench: Evaluating AI's Ability to Replicate AI Research}, 
      author={Giulio Starace and Oliver Jaffe and Dane Sherburn and James Aung and Jun Shern Chan and Leon Maksin and Rachel Dias and Evan Mays and Benjamin Kinsella and Wyatt Thompson and Johannes Heidecke and Amelia Glaese and Tejal Patwardhan},
      year={2025},
      eprint={2504.01848},
      archivePrefix={arXiv},
      primaryClass={cs.AI},
      url={https://arxiv.org/abs/2504.01848}, 
}

@misc{comanici2025gemini25pushingfrontier,
      title={Gemini 2.5: Pushing the Frontier with Advanced Reasoning, Multimodality, Long Context, and Next Generation Agentic Capabilities}, 
      author = {Comanici, Gheorghe and many others and the Gemini Team},
      year={2025},
      eprint={2507.06261},
      archivePrefix={arXiv},
      primaryClass={cs.CL},
      url={https://arxiv.org/abs/2507.06261}, 
}

@misc{deepseekai2025deepseekr1incentivizingreasoningcapability,
      title={DeepSeek-R1: Incentivizing Reasoning Capability in LLMs via Reinforcement Learning}, 
      author={DeepSeek-AI and Daya Guo and Dejian Yang and Haowei Zhang and Junxiao Song and Ruoyu Zhang and Runxin Xu and Qihao Zhu and Shirong Ma and Peiyi Wang and Xiao Bi and Xiaokang Zhang and Xingkai Yu and Yu Wu and Z. F. Wu and Zhibin Gou and Zhihong Shao and Zhuoshu Li and Ziyi Gao and Aixin Liu and Bing Xue and Bingxuan Wang and Bochao Wu and Bei Feng and Chengda Lu and Chenggang Zhao and Chengqi Deng and Chenyu Zhang and Chong Ruan and Damai Dai and Deli Chen and Dongjie Ji and Erhang Li and Fangyun Lin and Fucong Dai and Fuli Luo and Guangbo Hao and Guanting Chen and Guowei Li and H. Zhang and Han Bao and Hanwei Xu and Haocheng Wang and Honghui Ding and Huajian Xin and Huazuo Gao and Hui Qu and Hui Li and Jianzhong Guo and Jiashi Li and Jiawei Wang and Jingchang Chen and Jingyang Yuan and Junjie Qiu and Junlong Li and J. L. Cai and Jiaqi Ni and Jian Liang and Jin Chen and Kai Dong and Kai Hu and Kaige Gao and Kang Guan and Kexin Huang and Kuai Yu and Lean Wang and Lecong Zhang and Liang Zhao and Litong Wang and Liyue Zhang and Lei Xu and Leyi Xia and Mingchuan Zhang and Minghua Zhang and Minghui Tang and Meng Li and Miaojun Wang and Mingming Li and Ning Tian and Panpan Huang and Peng Zhang and Qiancheng Wang and Qinyu Chen and Qiushi Du and Ruiqi Ge and Ruisong Zhang and Ruizhe Pan and Runji Wang and R. J. Chen and R. L. Jin and Ruyi Chen and Shanghao Lu and Shangyan Zhou and Shanhuang Chen and Shengfeng Ye and Shiyu Wang and Shuiping Yu and Shunfeng Zhou and Shuting Pan and S. S. Li and Shuang Zhou and Shaoqing Wu and Shengfeng Ye and Tao Yun and Tian Pei and Tianyu Sun and T. Wang and Wangding Zeng and Wanjia Zhao and Wen Liu and Wenfeng Liang and Wenjun Gao and Wenqin Yu and Wentao Zhang and W. L. Xiao and Wei An and Xiaodong Liu and Xiaohan Wang and Xiaokang Chen and Xiaotao Nie and Xin Cheng and Xin Liu and Xin Xie and Xingchao Liu and Xinyu Yang and Xinyuan Li and Xuecheng Su and Xuheng Lin and X. Q. Li and Xiangyue Jin and Xiaojin Shen and Xiaosha Chen and Xiaowen Sun and Xiaoxiang Wang and Xinnan Song and Xinyi Zhou and Xianzu Wang and Xinxia Shan and Y. K. Li and Y. Q. Wang and Y. X. Wei and Yang Zhang and Yanhong Xu and Yao Li and Yao Zhao and Yaofeng Sun and Yaohui Wang and Yi Yu and Yichao Zhang and Yifan Shi and Yiliang Xiong and Ying He and Yishi Piao and Yisong Wang and Yixuan Tan and Yiyang Ma and Yiyuan Liu and Yongqiang Guo and Yuan Ou and Yuduan Wang and Yue Gong and Yuheng Zou and Yujia He and Yunfan Xiong and Yuxiang Luo and Yuxiang You and Yuxuan Liu and Yuyang Zhou and Y. X. Zhu and Yanhong Xu and Yanping Huang and Yaohui Li and Yi Zheng and Yuchen Zhu and Yunxian Ma and Ying Tang and Yukun Zha and Yuting Yan and Z. Z. Ren and Zehui Ren and Zhangli Sha and Zhe Fu and Zhean Xu and Zhenda Xie and Zhengyan Zhang and Zhewen Hao and Zhicheng Ma and Zhigang Yan and Zhiyu Wu and Zihui Gu and Zijia Zhu and Zijun Liu and Zilin Li and Ziwei Xie and Ziyang Song and Zizheng Pan and Zhen Huang and Zhipeng Xu and Zhongyu Zhang and Zhen Zhang},
      year={2025},
      eprint={2501.12948},
      archivePrefix={arXiv},
      primaryClass={cs.CL},
      url={https://arxiv.org/abs/2501.12948}, 
}

@misc{deepseekai2025deepseekv3technicalreport,
      title={DeepSeek-V3 Technical Report}, 
      author={DeepSeek-AI and Aixin Liu and Bei Feng and Bing Xue and Bingxuan Wang and Bochao Wu and Chengda Lu and Chenggang Zhao and Chengqi Deng and Chenyu Zhang and Chong Ruan and Damai Dai and Daya Guo and Dejian Yang and Deli Chen and Dongjie Ji and Erhang Li and Fangyun Lin and Fucong Dai and Fuli Luo and Guangbo Hao and Guanting Chen and Guowei Li and H. Zhang and Han Bao and Hanwei Xu and Haocheng Wang and Haowei Zhang and Honghui Ding and Huajian Xin and Huazuo Gao and Hui Li and Hui Qu and J. L. Cai and Jian Liang and Jianzhong Guo and Jiaqi Ni and Jiashi Li and Jiawei Wang and Jin Chen and Jingchang Chen and Jingyang Yuan and Junjie Qiu and Junlong Li and Junxiao Song and Kai Dong and Kai Hu and Kaige Gao and Kang Guan and Kexin Huang and Kuai Yu and Lean Wang and Lecong Zhang and Lei Xu and Leyi Xia and Liang Zhao and Litong Wang and Liyue Zhang and Meng Li and Miaojun Wang and Mingchuan Zhang and Minghua Zhang and Minghui Tang and Mingming Li and Ning Tian and Panpan Huang and Peiyi Wang and Peng Zhang and Qiancheng Wang and Qihao Zhu and Qinyu Chen and Qiushi Du and R. J. Chen and R. L. Jin and Ruiqi Ge and Ruisong Zhang and Ruizhe Pan and Runji Wang and Runxin Xu and Ruoyu Zhang and Ruyi Chen and S. S. Li and Shanghao Lu and Shangyan Zhou and Shanhuang Chen and Shaoqing Wu and Shengfeng Ye and Shengfeng Ye and Shirong Ma and Shiyu Wang and Shuang Zhou and Shuiping Yu and Shunfeng Zhou and Shuting Pan and T. Wang and Tao Yun and Tian Pei and Tianyu Sun and W. L. Xiao and Wangding Zeng and Wanjia Zhao and Wei An and Wen Liu and Wenfeng Liang and Wenjun Gao and Wenqin Yu and Wentao Zhang and X. Q. Li and Xiangyue Jin and Xianzu Wang and Xiao Bi and Xiaodong Liu and Xiaohan Wang and Xiaojin Shen and Xiaokang Chen and Xiaokang Zhang and Xiaosha Chen and Xiaotao Nie and Xiaowen Sun and Xiaoxiang Wang and Xin Cheng and Xin Liu and Xin Xie and Xingchao Liu and Xingkai Yu and Xinnan Song and Xinxia Shan and Xinyi Zhou and Xinyu Yang and Xinyuan Li and Xuecheng Su and Xuheng Lin and Y. K. Li and Y. Q. Wang and Y. X. Wei and Y. X. Zhu and Yang Zhang and Yanhong Xu and Yanhong Xu and Yanping Huang and Yao Li and Yao Zhao and Yaofeng Sun and Yaohui Li and Yaohui Wang and Yi Yu and Yi Zheng and Yichao Zhang and Yifan Shi and Yiliang Xiong and Ying He and Ying Tang and Yishi Piao and Yisong Wang and Yixuan Tan and Yiyang Ma and Yiyuan Liu and Yongqiang Guo and Yu Wu and Yuan Ou and Yuchen Zhu and Yuduan Wang and Yue Gong and Yuheng Zou and Yujia He and Yukun Zha and Yunfan Xiong and Yunxian Ma and Yuting Yan and Yuxiang Luo and Yuxiang You and Yuxuan Liu and Yuyang Zhou and Z. F. Wu and Z. Z. Ren and Zehui Ren and Zhangli Sha and Zhe Fu and Zhean Xu and Zhen Huang and Zhen Zhang and Zhenda Xie and Zhengyan Zhang and Zhewen Hao and Zhibin Gou and Zhicheng Ma and Zhigang Yan and Zhihong Shao and Zhipeng Xu and Zhiyu Wu and Zhongyu Zhang and Zhuoshu Li and Zihui Gu and Zijia Zhu and Zijun Liu and Zilin Li and Ziwei Xie and Ziyang Song and Ziyi Gao and Zizheng Pan},
      year={2025},
      eprint={2412.19437},
      archivePrefix={arXiv},
      primaryClass={cs.CL},
      url={https://arxiv.org/abs/2412.19437}, 
}

@misc{meta2024llama4,
  title = {Introducing LLaMA 4: Advancing multimodal intelligence},
  author = {{Meta AI}},
  year = {2024},
  url = {https://ai.meta.com/blog/llama-4-multimodal-intelligence/}
}

@misc{openai2025o4mini,
  title = {OpenAI o4‑mini: reasoning model release},
  author = {OpenAI},
  year = {2025},
  note = {Blog post (16\,Apr\,2025) announcing the o3 and o4‑mini models; o4‑mini is a small model optimized for fast, cost‑efficient reasoning},
  url = {https://openai.com/research/o3-and-o4-mini}
}

@misc{openai2025gpt41,
  title = {GPT‑4.1 and GPT‑4.1 mini release notes},
  author = {OpenAI},
  year = {2025},
  note = {Release notes (14\,May\,2025) announcing GPT‑4.1 and GPT‑4.1 mini; GPT‑4.1 excels at coding and precise instruction following, while GPT‑4.1 mini is a fast, efficient model with a 1\,M‑token context window},
  url = {https://help.openai.com/en/articles/9180355}
}

@misc{openai2025gpt5,
  author  = {OpenAI},
  title   = {Introducing {GPT-5} for developers},
  year    = {2025},
  month   = aug,
  url     = {https://openai.com/index/introducing-gpt-5-for-developers/},
  urldate = {2025-09-26},
  note    = {Announces API model sizes including \texttt{gpt-5-mini}}
}

@misc{kwon2023efficientmemorymanagementlarge,
      title={Efficient Memory Management for Large Language Model Serving with PagedAttention}, 
      author={Woosuk Kwon and Zhuohan Li and Siyuan Zhuang and Ying Sheng and Lianmin Zheng and Cody Hao Yu and Joseph E. Gonzalez and Hao Zhang and Ion Stoica},
      year={2023},
      eprint={2309.06180},
      archivePrefix={arXiv},
      primaryClass={cs.LG},
      url={https://arxiv.org/abs/2309.06180}, 
}

@misc{qwen3techreport2025,
  title        = {Qwen3 Technical Report},
  author       = {{Qwen Team}},
  year         = {2025},
  eprint       = {2505.09388},
  archivePrefix= {arXiv},
  primaryClass = {cs.CL},
  url          = {https://arxiv.org/abs/2505.09388}
}

@misc{mistralai2025mistralmedium,
  title        = {Mistral Medium: Model Overview and Documentation},
  author       = {{Mistral AI}},
  year         = {2025},
  howpublished = {Product Documentation},
  url          = {https://docs.mistral.ai/}
}

@misc{xai2025grok4fast,
  title        = {Grok 4 Fast (Reasoning) — Model Overview},
  author       = {{xAI}},
  year         = {2025},
  howpublished = {xAI Documentation/News},
  url          = {https://x.ai/news/grok-4-fast}
}

@misc{xai2025grok4fastnonreasoning,
  title        = {Grok 4 Fast (Non-Reasoning) — Model Overview},
  author       = {{xAI}},
  year         = {2025},
  howpublished = {xAI Documentation},
  url          = {https://docs.x.ai/docs/models/grok-4-fast-non-reasoning}
}

@misc{defog2024sqlcoder7b2,
  title        = {{SQLCoder-7B-2}},
  author       = {{Defog.ai}},
  year         = {2024},
  howpublished = {\url{https://huggingface.co/defog/sqlcoder-7b-2}},
  note         = {Model card},
  urldate      = {2025-10-05}
}

@misc{medgemma27btextit,
  title        = {{MedGemma 27B Text-only (medgemma-27b-text-it)}},
  author       = {{Google DeepMind}},
  year         = {2025},
  howpublished = {\url{https://huggingface.co/google/medgemma-27b-text-it}},
  note         = {Model card},
  urldate      = {2025-10-05}
}

@misc{baichuanm232b,
  title        = {{Baichuan-M2-32B}},
  author       = {{Baichuan AI}},
  year         = {2025},
  howpublished = {\url{https://huggingface.co/baichuan-inc/Baichuan-M2-32B}},
  note         = {Model card},
  urldate      = {2025-10-05}
}

@article{ke2025early,
  title={Early warning of cryptocurrency reversal risks via multi-source data},
  author={Ke, Zong and Cao, Yuqing and Chen, Zhenrui and Yin, Yuchen and He, Shouchao and Cheng, Yu},
  journal={Finance Research Letters},
  pages={107890},
  year={2025},
  publisher={Elsevier}
}

@article{ke2025stable,
  title={A stable technical feature with GRU-CNN-GA fusion},
  author={Ke, Zong and Shen, Jiaqing and Zhao, Xuanyi and Fu, Xinghao and Wang, Yang and Li, Zichao and Liu, Lingjie and Mu, Huailing},
  journal={Applied Soft Computing},
  pages={114302},
  year={2025},
  publisher={Elsevier}
}

@article{ouyang2024learn,
  title={Learn from global correlations: Enhancing evolutionary algorithm via spectral gnn},
  author={Ouyang, Kaichen and Ke, Zong and Fu, Shengwei and Liu, Lingjie and Zhao, Puning and Hu, Dayu},
  journal={arXiv preprint arXiv:2412.17629},
  year={2024}
}

@inproceedings{yang-etal-2025-table,
    title = "Table-R1: Inference-Time Scaling for Table Reasoning Tasks",
    author = "Yang, Zheyuan  and
      Chen, Lyuhao  and
      Cohan, Arman  and
      Zhao, Yilun",
    editor = "Christodoulopoulos, Christos  and
      Chakraborty, Tanmoy  and
      Rose, Carolyn  and
      Peng, Violet",
    booktitle = "Proceedings of the 2025 Conference on Empirical Methods in Natural Language Processing",
    month = nov,
    year = "2025",
    address = "Suzhou, China",
    publisher = "Association for Computational Linguistics",
    url = "https://aclanthology.org/2025.emnlp-main.1040/",
    doi = "10.18653/v1/2025.emnlp-main.1040",
    pages = "20605--20624",
    ISBN = "979-8-89176-332-6"
}

\appendix

\onecolumn
\section{\ours Benchmark Construction}
\label{app:benchmark-construction}
\begin{table*}[h]
\centering
\small
\renewcommand{\arraystretch}{1.05}
\begin{tabular}{llllc}
\toprule
\textbf{ID} & \textbf{Year} & \textbf{Major} & \textbf{Assigned Scenario} & \textbf{Author?} \\
\midrule
1  & 3rd-year PhD & Health Informatics & Patient Demographics \& Admissions & \xmark \\
2  & \textemdash{} & \textemdash{} & \textemdash{} & \cmark \\
3 & 5th-year PhD & Biochemistry & Laboratory Results Analysis & \xmark \\
4 & 3rd-year PhD & Medicine & Medication Management & \xmark \\
5 & 4th-year PhD & Biomedical Engineering & Diagnostic Procedures & \xmark \\
6 & \textemdash{} & \textemdash{} & \textemdash{} & \cmark \\
\bottomrule
\end{tabular}
\caption{Overview of the 6 expert annotators who contributed to the \ours\ Benchmark Construction. Two annotators (IDs 2 and 6) are paper authors; to preserve confidentiality, their rows omit year, major, and scenario details.}
\label{tab:expert_annotators}
\end{table*}

\clearpage
\section{Example Clinical Questions}
\label{app:example}

\subsection{Patient Demographics Example}
\label{app:e1}
% Add detailed example content here
\paragraph{Query} 
For an 81‑year‑old female: among female Medicare patients aged 76–86 transferred from another hospital with principal AMI (ICD‑9 410*/ICD‑10 I21*), report 30‑day readmission rate; median index LOS for readmitted vs not; percent index stays $>4$ days.

\begin{figure}[ht]
    \centering
    \includegraphics[width=\linewidth]{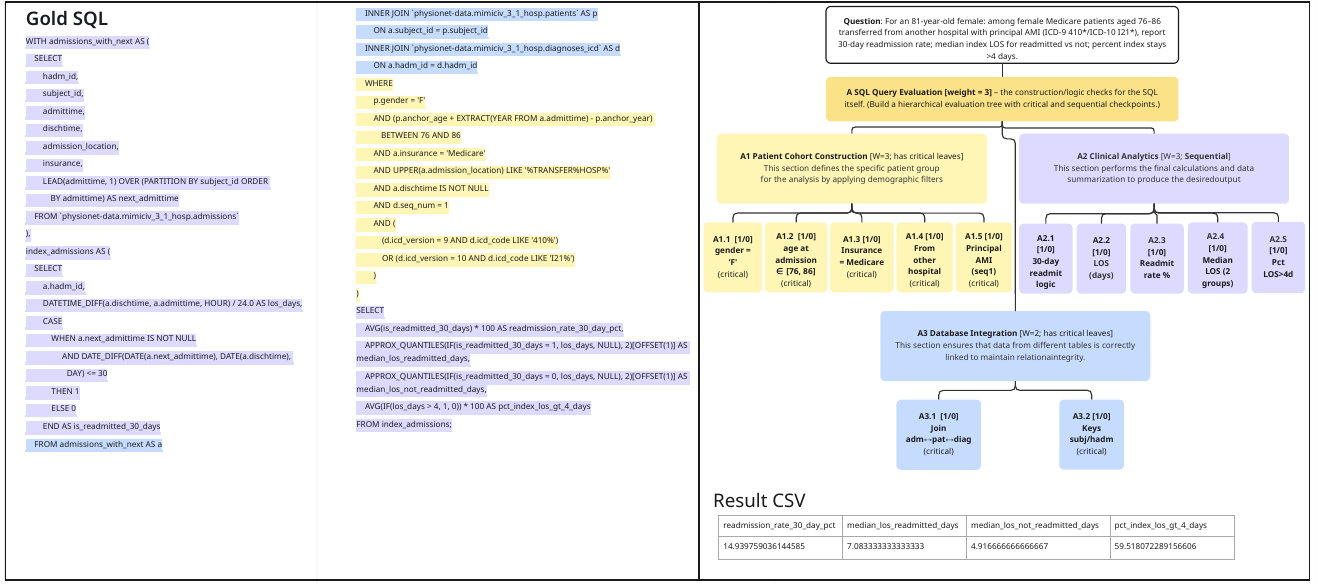}
    \caption{SQL evaluation rubric tree and Result for a \ours sample: Patient Demographics.}
    \label{fig:Eg_PD}
\end{figure}

\paragraph{SQL} 
As Figure~\ref{fig:Eg_PD}

\subsection{Vital Signs Monitoring Example}
\label{app:e2}
\begin{figure}[ht]
    \centering
    \includegraphics[width=0.95\linewidth]{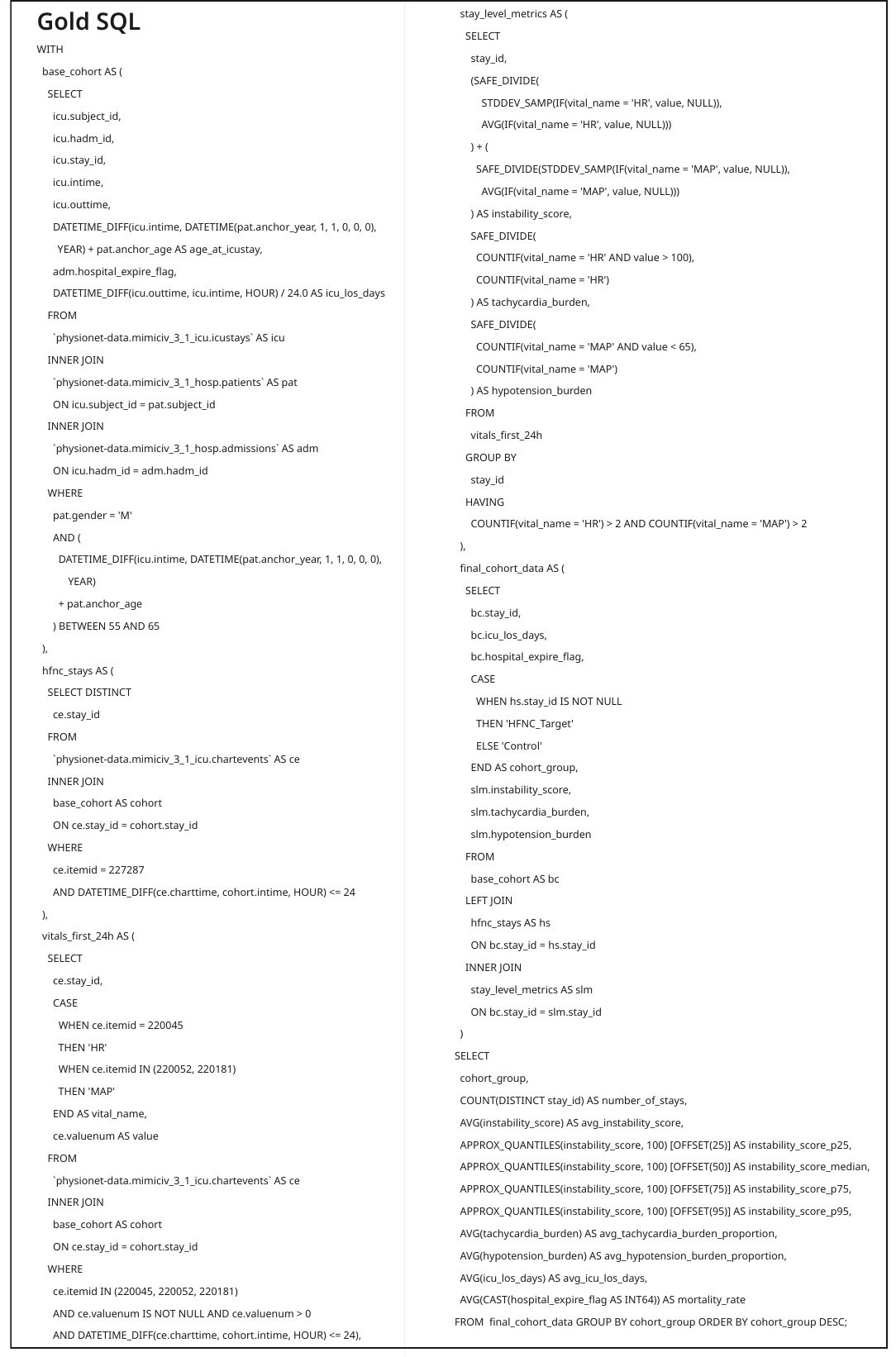}
    \caption{SQL sample: Vital Signs Monitoring.}
    \label{fig:Eg_Vital}
\end{figure}

% Add detailed example content here
\paragraph{Query} 
I have a 60‑year‑old man in the ICU. In male ICU patients aged 55–65 with HFNC within 24 hours versus condition‑matched ICU controls, what are the instability score median and p25/p75/p95, tachycardia and hypotension burden, ICU LOS and mortality?
\paragraph{SQL} 
As Figure~\ref{fig:Eg_Vital}

\subsection{Laboratory Results Analysis Example}
\label{app:e3}
\begin{figure}[ht]
    \centering
    \includegraphics[width=\linewidth]{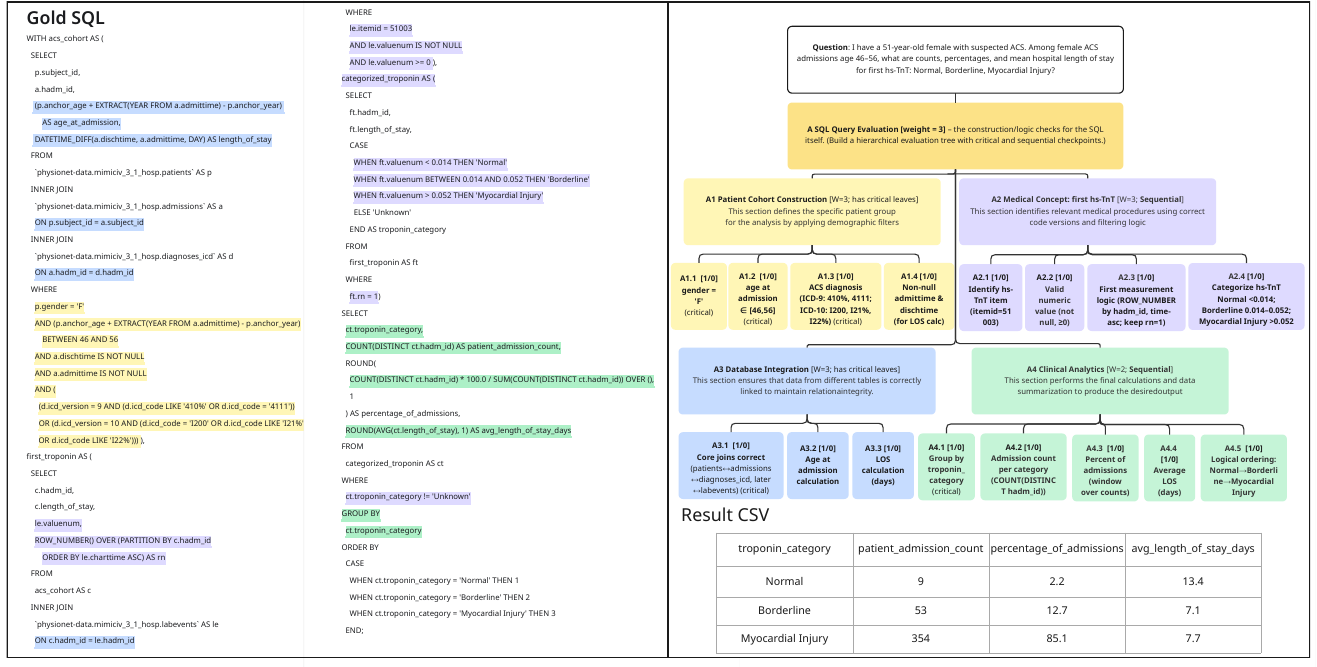}
    \caption{SQL evaluation rubric tree and Result for a \ours sample: Laboratory Results Analysis.}
    \label{fig:Lab_rubric_tree}
\end{figure}

\paragraph{Query} 
I have a 51-year-old female with suspected ACS. Among female ACS admissions age 46–56, what are counts, percentages, and mean hospital length of stay for first hs‑TnT: Normal, Borderline, Myocardial Injury?
\paragraph{SQL} 
As Figure~\ref{fig:Lab_rubric_tree}

\subsection{Medication Management Example}
\label{app:e4}
% Add detailed example content here
\paragraph{Query} 
I have a 64‑year‑old female inpatient. Among females aged 59–69, what's the IQR of single inpatient amiodarone prescription durations (days)?
\begin{figure}[ht]
    \centering
    \includegraphics[width=\linewidth]{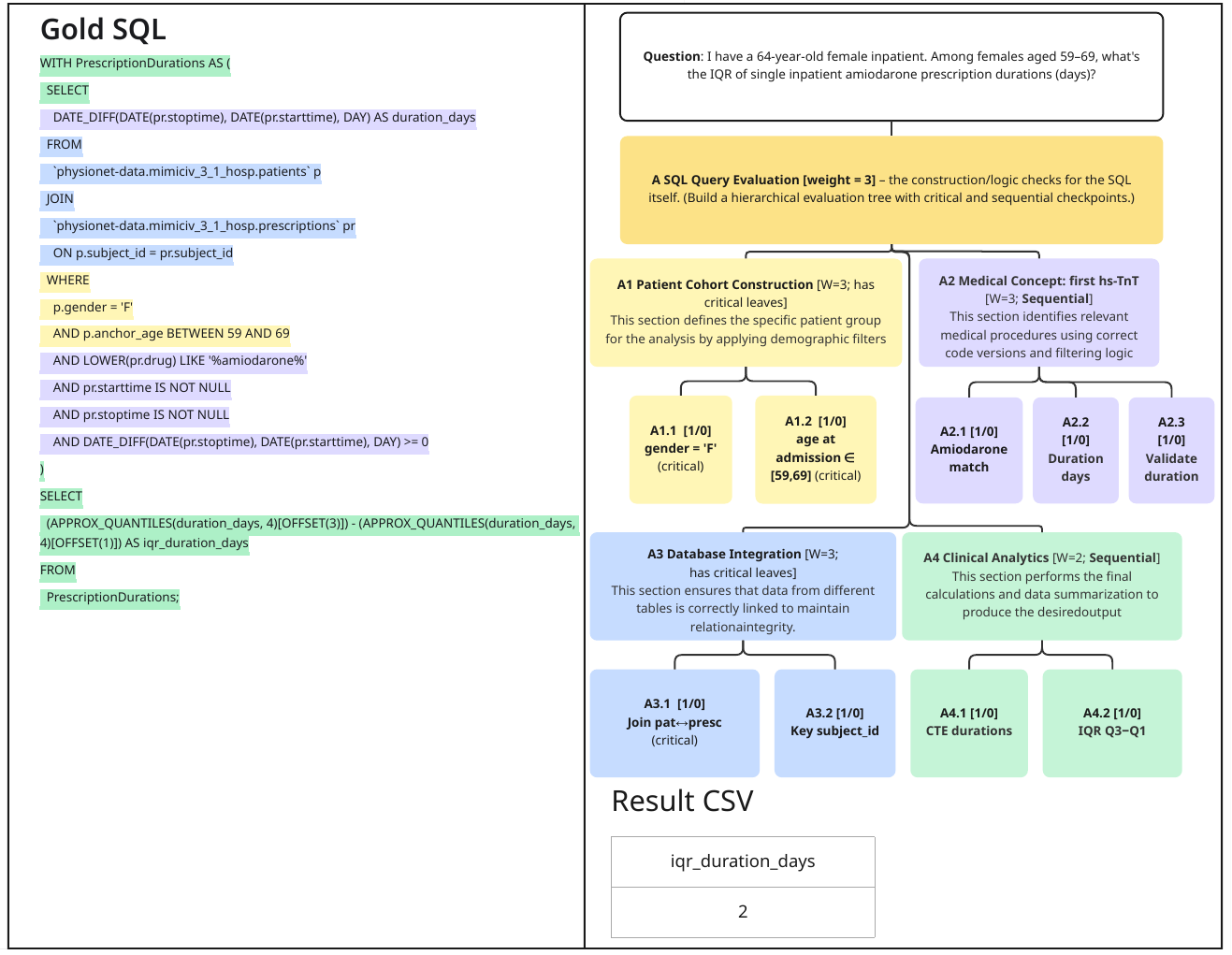}
    \caption{SQL evaluation rubric tree and Result for a \ours sample: Medication Management.}
    \label{fig:Eg_MM}
\end{figure}
\paragraph{SQL} 
As Figure~\ref{fig:Eg_MM}

\subsection{Diagnostic Procedures Example}
\label{app:e5}
% Add detailed example content here
\paragraph{Query} 
Evaluating an 88-year-old man: among male patients aged 83–93 with sepsis on their first ICU stay, stratify first‑72‑hour diagnostic intensity (distinct procedures) into quartiles and report mean procedure counts, mean ICU LOS in days, and mortality (\%) per quartile.
\begin{figure}[ht]
    \centering
    \includegraphics[width=\linewidth]{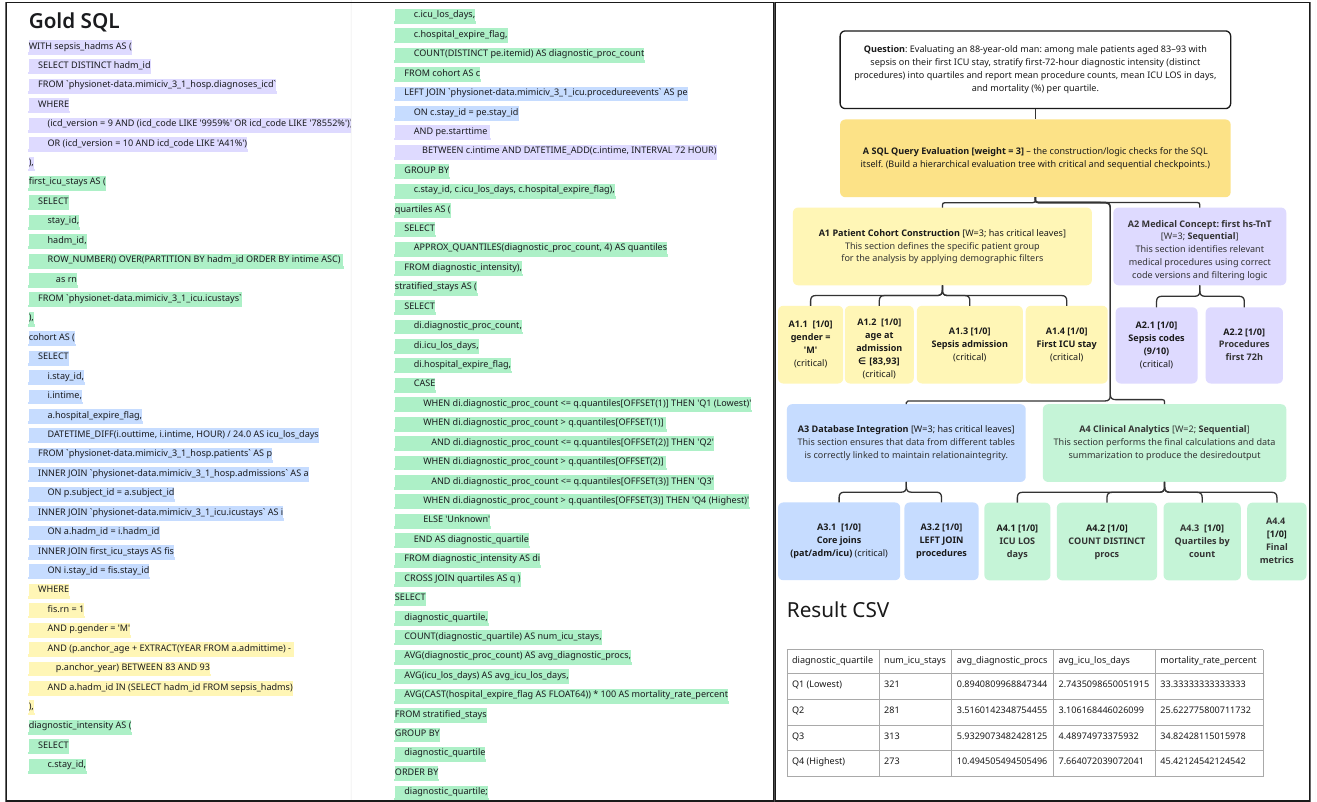}
    \caption{SQL evaluation rubric tree and Result for a \ours sample: Diagnostic Procedures.}
    \label{fig:Eg_DP}
\end{figure}
\paragraph{SQL} 
As Figure~\ref{fig:Eg_DP}

\subsection{Disease Diagnosis and Outcomes Example}
\label{app:e6}
% Add detailed example content here
\paragraph{Query} 
I have a 75-year-old female inpatient with pulmonary embolism. For female inpatients aged 70–80 with PE, stratify into risk-score quintiles and report per quintile: 90‑day mortality, general 70–80 female 90‑day mortality (comparison), AKI and ARDS rates, and median survivor LOS.

\begin{figure}[ht]
    \centering
    \includegraphics[width=\linewidth]{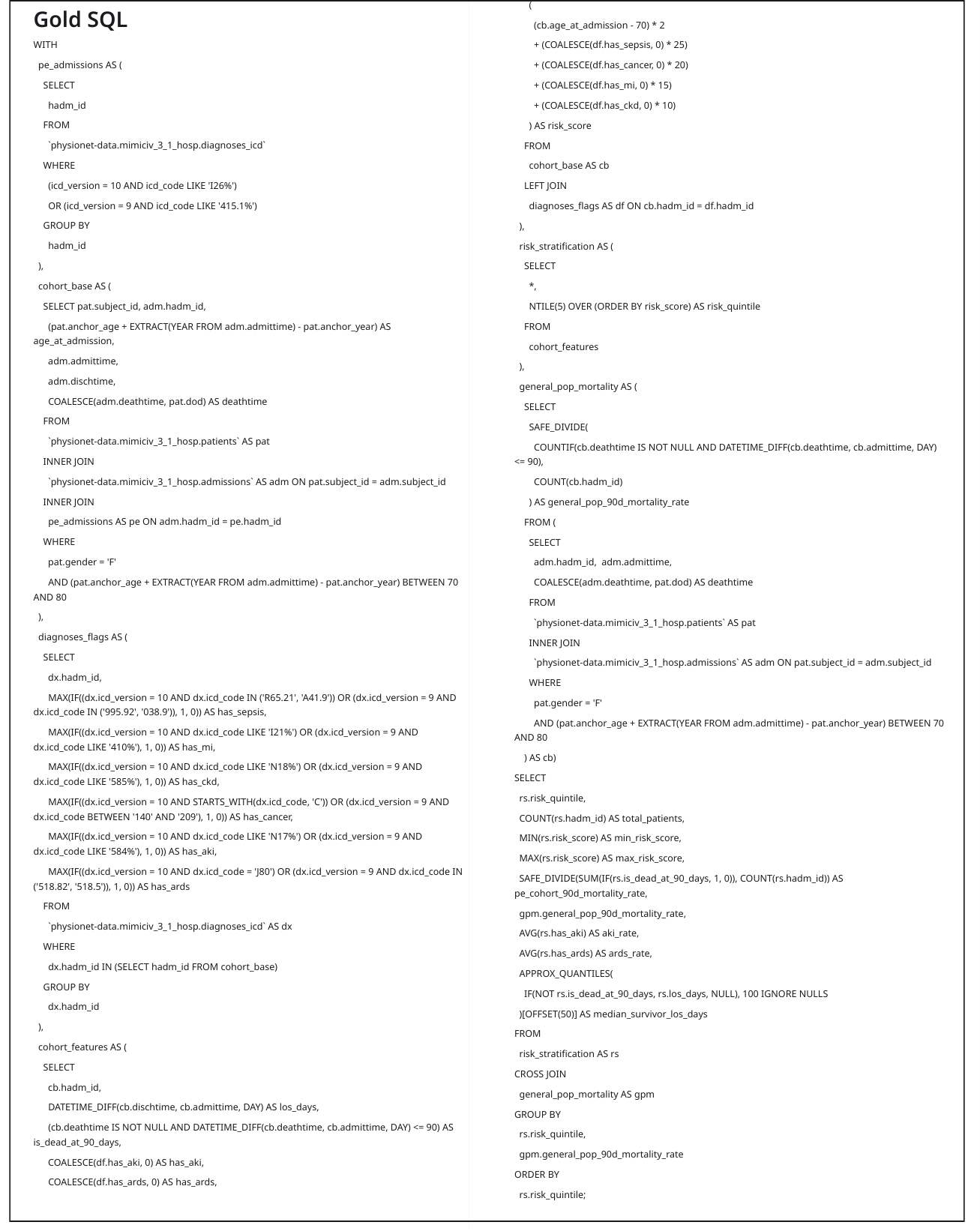}
    \caption{SQL sample: Disease Diagnosis and Outcomes.}
    \label{fig:Eg_DDO}
\end{figure}
\paragraph{SQL} 
As Figure~\ref{fig:Eg_DDO}

\clearpage
\section{Error Analysis}\label{app:err}
\begin{figure}[ht]
    \centering
    \includegraphics[width=\linewidth]{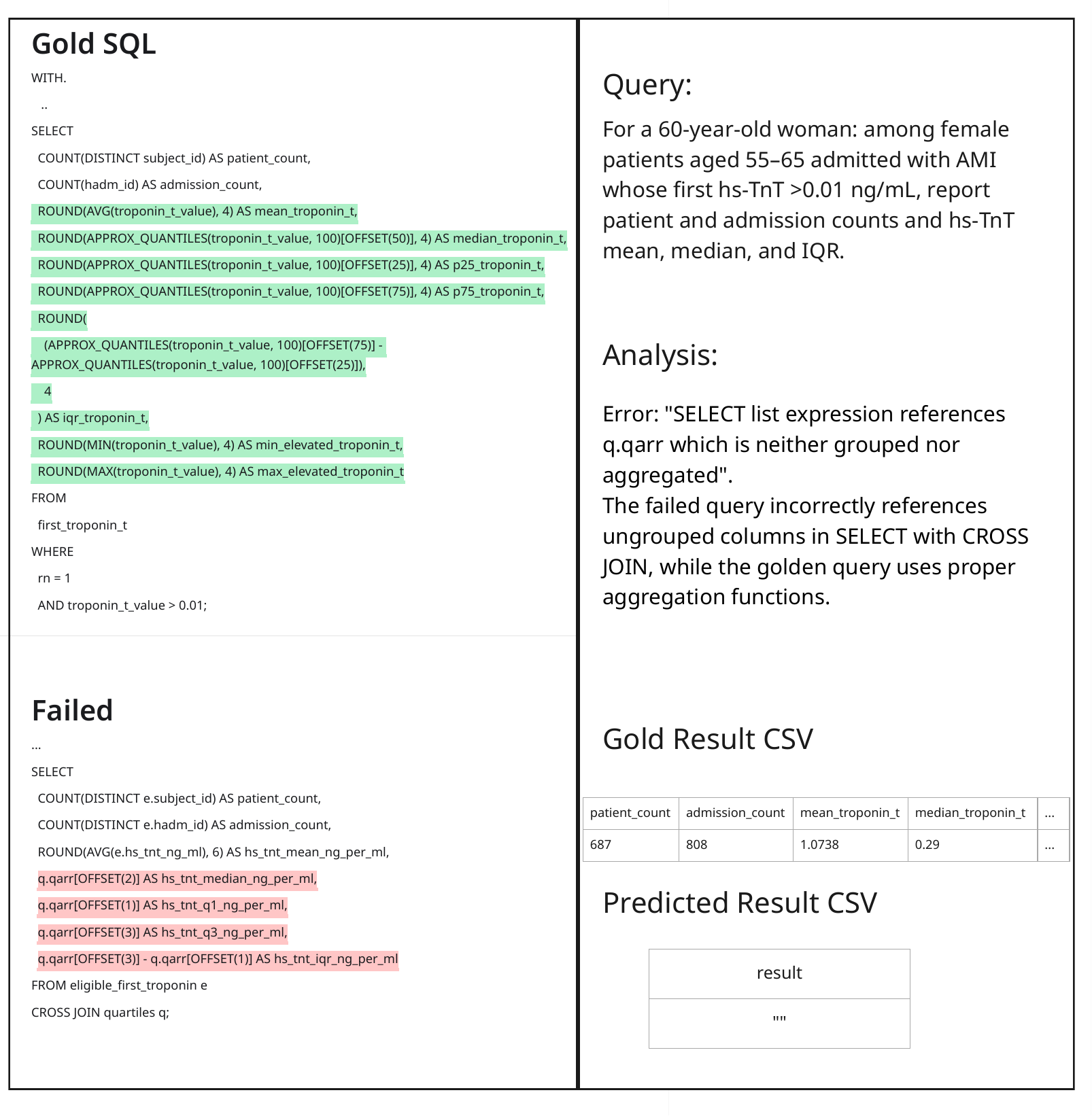}
    \caption{A sample of Error Analysis: Output Schema and Formatting}
    \label{fig:error_analysis_output}
\end{figure}

\begin{figure}[ht]
    \centering
    \includegraphics[width=\linewidth]{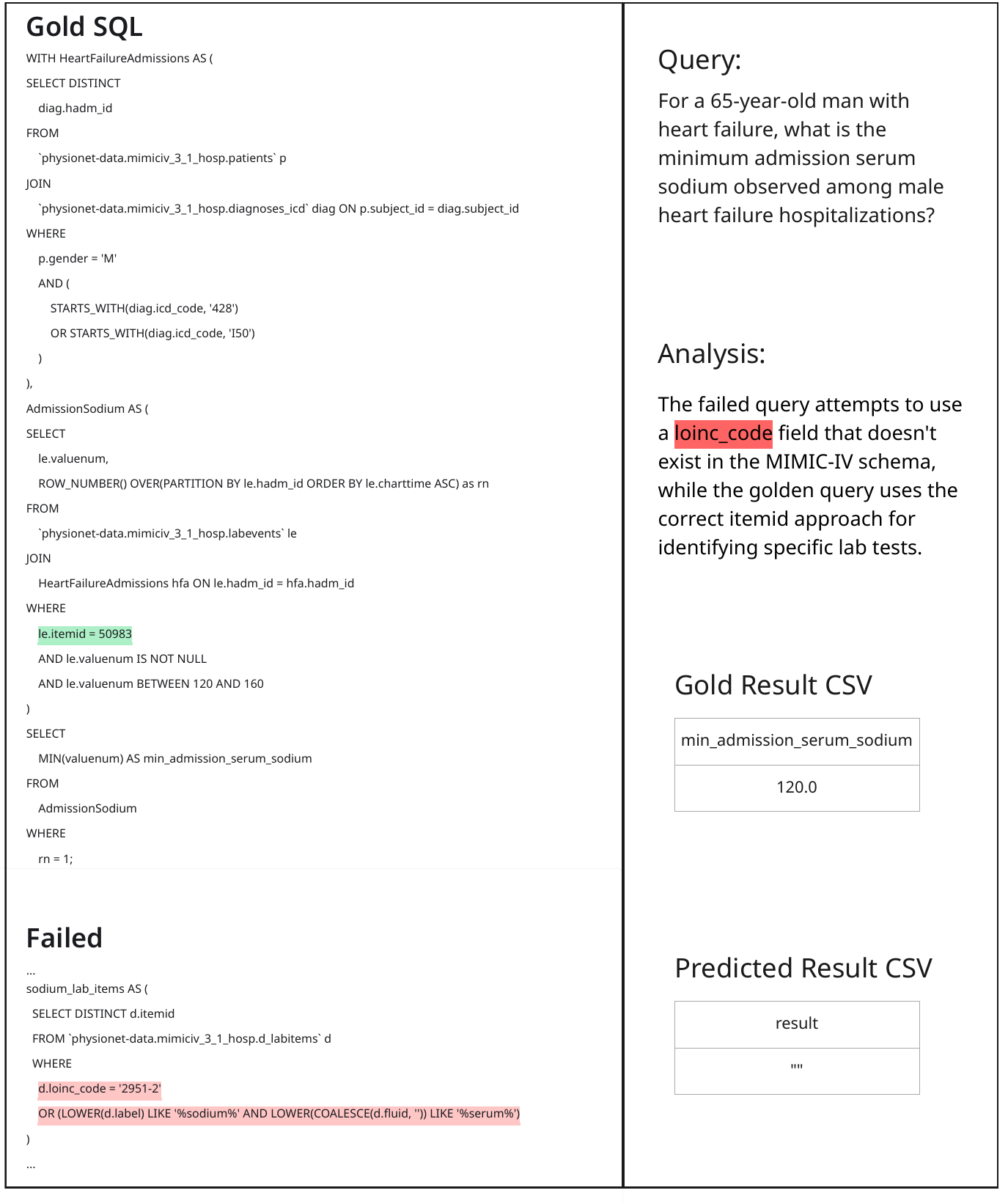}
    \caption{A sample of Error Analysis: Cohort Specification and Coding}
    \label{fig:error_analysis_cohort}
\end{figure}

\begin{figure}[ht]
    \centering
    \includegraphics[width=\linewidth]{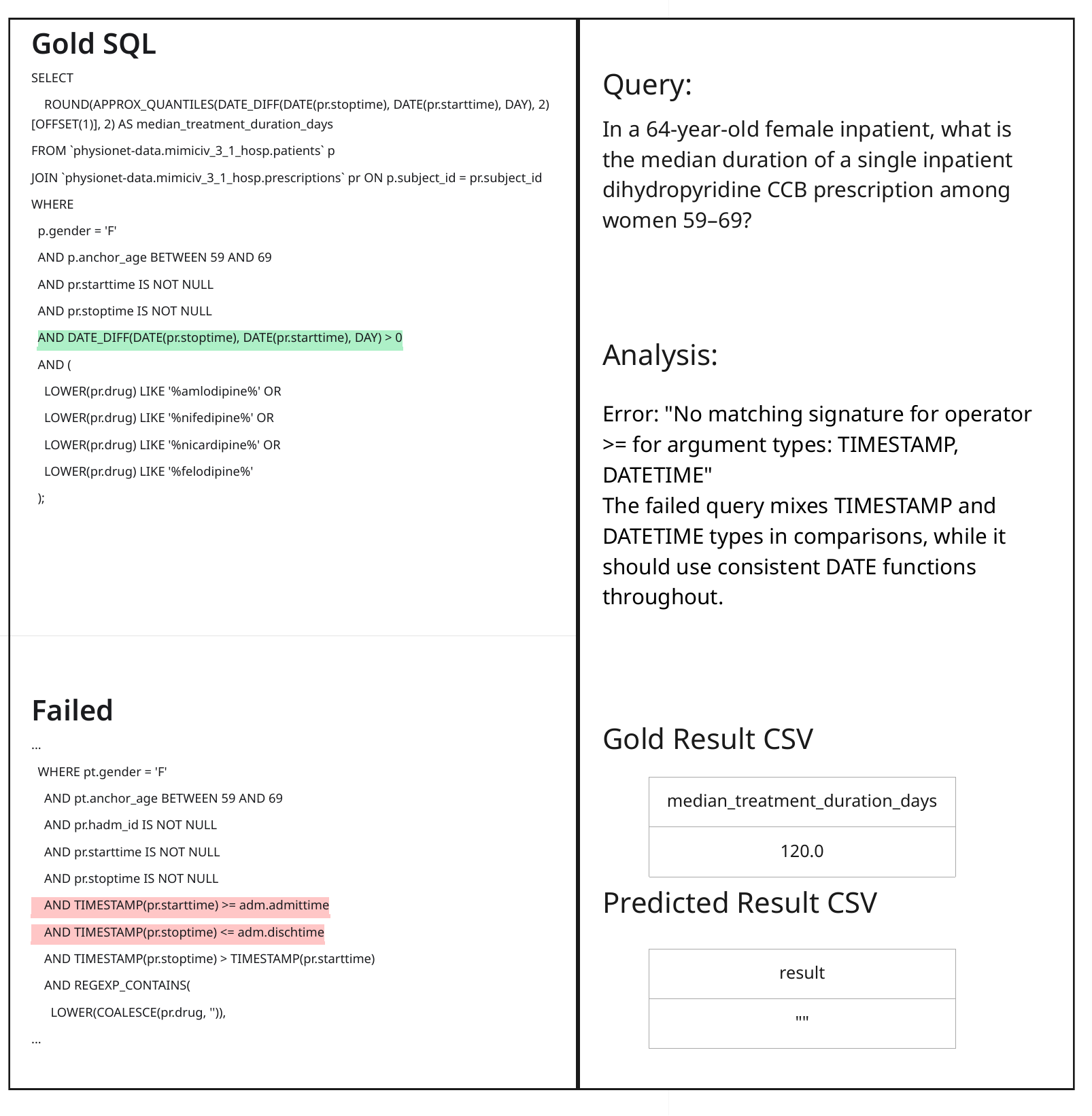}
    \caption{A sample of Error Analysis: Aggregation and Clinical Statistics}
    \label{fig:error_analysis_aggre}
\end{figure}

\clearpage
\section{SQL Generation Prompts}
\label{app:model-prompts}
We include the exact prompt templates used for model prompting and refinement in both Chain-of-Thought (CoT) and Direct Output regimes.

% CoT prompts
\begin{figure}[!h]
\begin{tcolorbox}[colback=black!3!white, colframe=black!70!white, title=SQL Generation Prompt (CoT), fontupper=\footnotesize, fonttitle=\footnotesize]

You are a clinical data analyst expert specializing in the MIMIC-IV database. Your goal is to produce a correct BigQuery SQL query for the question below.\\

Constraints:\\
- Target platform: Google BigQuery.\\
- Use the correct datasets:\\
\texttt{physionet-data.mimiciv\_3\_1\_hosp}, \texttt{physionet-data.mimiciv\_3\_1\_icu}.\\

MIMIC-IV Schema Reference (HOSP + ICU):\\
\textcolor{blue}{\{schema text\}} \\

Clinical question:\\
\textcolor{blue}{``\{Question\}''} \\

Your output should be organized in the following two parts:\\
Reasoning:\\
- Think step by step about relevant tables, joins, filters, groupings, and edge cases.\\
- Briefly justify important choices.\\
SQL (wrap the final query in a fenced code block using \texttt{```sql} and \texttt{```}):\\

Think step by step and then generate the complete SQL query.

\end{tcolorbox}
\caption{Chain-of-Thought SQL generation prompt used in our experiments.}
\label{fig:sql-prompt}
\end{figure}
\begin{figure}[!h]
\begin{tcolorbox}[colback=black!3!white, colframe=black!70!white, title=Refinement Prompt (CoT), fontupper=\footnotesize, fonttitle=\footnotesize]

You are a clinical data analyst expert for the MIMIC-IV dataset. The following SQL failed to run on Google BigQuery. Refine it to resolve the error and better answer the question.\\

Constraints:\\
- Use valid BigQuery SQL.\\
- Use the correct datasets:\\
\texttt{physionet-data.mimiciv\_3\_1\_hosp}, \texttt{physionet-data.mimiciv\_3\_1\_icu}.\\
- Modify only what is necessary; prefer minimal, correct fixes.\\

MIMIC-IV Schema Reference (HOSP + ICU):\\
\textcolor{blue}{\{schema text\}} \\

Clinical question:\\
\textcolor{blue}{\{Question\}} \\

Previous SQL attempt (for reference):\\
\textcolor{blue}{\{Previous SQL (provided as a fenced code block)\}} \\

BigQuery error message:\\
\textcolor{blue}{\{Error message\}} \\

Your output should be organized in the following two parts:\\
Reasoning:\\
- Step by step, explain the cause of the error and the fix.\\
- Justify key changes briefly.\\
SQL (wrap the final corrected query in a fenced code block using \texttt{```sql} and \texttt{```}):\\

Think step by step and then generate the complete corrected SQL query.

\end{tcolorbox}
\caption{Chain-of-Thought SQL refinement prompt used in our experiments.}
\label{fig:refinement-prompt}
\end{figure}

% Direct Output prompts
\begin{figure}[!ht]
\begin{tcolorbox}[colback=black!3!white, colframe=black!70!white, title=SQL Generation Prompt (Direct Output), fontupper=\footnotesize, fonttitle=\footnotesize]

You are a clinical data analyst expert specializing in the MIMIC-IV database. Your goal is to produce a correct BigQuery SQL query for the question below.\\

Constraints:\\
- Target platform: Google BigQuery.\\
- Use the correct datasets:\\
\texttt{physionet-data.mimiciv\_3\_1\_hosp}, \texttt{physionet-data.mimiciv\_3\_1\_icu}.\\

MIMIC-IV Schema Reference (HOSP + ICU):\\
\textcolor{blue}{\{schema text\}} \\

Clinical question:\\
\textcolor{blue}{``\{Question\}''} \\

Output format:\\
- Return only a single fenced SQL code block containing the final query (use \texttt{```sql} and \texttt{```}).\\
- Do not include explanations, or any text outside the fenced SQL block.\\

\end{tcolorbox}
\caption{Direct Output SQL generation prompt used in our experiments.}
\label{fig:sql-prompt-direct}
\end{figure}

\begin{figure}[!ht]
\begin{tcolorbox}[colback=black!3!white, colframe=black!70!white, title=Refinement Prompt (Direct Output), fontupper=\footnotesize, fonttitle=\footnotesize]

You are a clinical data analyst expert for the MIMIC-IV dataset. The following SQL failed to run on Google BigQuery. Refine it to resolve the error and better answer the question.\\

Constraints:\\
- Use valid BigQuery SQL.\\
- Use the correct datasets:\\
\texttt{physionet-data.mimiciv\_3\_1\_hosp}, \texttt{physionet-data.mimiciv\_3\_1\_icu}.\\
- Modify only what is necessary; prefer minimal, correct fixes.\\

MIMIC-IV Schema Reference (HOSP + ICU):\\
\textcolor{blue}{\{schema text\}} \\

Clinical question:\\
\textcolor{blue}{\{Question\}} \\

Previous SQL attempt (for reference):\\
\textcolor{blue}{\{Previous SQL (provided as a fenced code block)\}} \\

BigQuery error message:\\
\textcolor{blue}{\{Error message\}} \\

Output format:\\
- Return only a single fenced SQL code block containing the corrected query (use \texttt{```sql} and \texttt{```}).\\
- Do not include explanations, or any text outside the fenced SQL block.\\

Apply the minimal fix internally and output only the final corrected SQL.

\end{tcolorbox}
\caption{Direct Output SQL refinement prompt used in our experiments.}
\label{fig:refinement-prompt-direct}
\end{figure}

\clearpage
\section{Configuration of Evaluated Models}
\label{app:model-configs}
\begin{table*}[h]
\centering
\footnotesize
\resizebox{\textwidth}{!}{%
\begin{tabular}{llllc}
\toprule
\textbf{Organization} & \textbf{Model} & \textbf{Release} & \textbf{Version} & \textbf{\# Inference Pipeline} \\
\midrule
\multicolumn{5}{c}{\emph{\textbf{Proprietary Models}}} \\
\midrule
\multirow{5}{*}{OpenAI}
& GPT-5-mini & 2025-08 & \texttt{gpt-5-mini-2025-08-07} & \multirow{5}{*}{API} \\
& GPT-5-nano & 2025-08 & \texttt{gpt-5-nano-2025-08-07} & \\
& GPT-5 & 2025-08 & \texttt{gpt-5-chat-2025-08-07} & \\
& GPT-4.1 & 2025-04 & \texttt{gpt-4.1-2025-04-14} & \\
& o4-mini & 2025-04 & \texttt{o4-mini-2025-04-16} & \\
\noalign{\vskip 0.4ex}\hdashline\noalign{\vskip 0.4ex}
\multirow{2}{*}{Google}
& Gemini-2.5-Pro & 2025-06 & \texttt{gemini-2.5-pro} & \multirow{2}{*}{API} \\
& Gemini-2.5-Flash & 2025-06 & \texttt{gemini-2.5-flash} & \\
\noalign{\vskip 0.4ex}\hdashline\noalign{\vskip 0.4ex}
\multirow{2}{*}{xAI}
& Grok-4-Fast-Reason. & 2025-09 & \texttt{grok-4-fast-reasoning} & \multirow{2}{*}{API} \\
& Grok-4-Fast-Non-Reason. & 2025-09 & \texttt{grok-4-fast-non-reasoning} & \\
\noalign{\vskip 0.4ex}\hdashline\noalign{\vskip 0.4ex}
\multirow{1}{*}{Mistral AI}
& Mistral-Medium & 2025-05 & \texttt{mistral-medium-2505} & API \\
\midrule
\multicolumn{5}{c}{\emph{\textbf{Open-Source Models}}} \\
\midrule
\multirow{2}{*}{DeepSeek}
& DeepSeek-R1 & 2025-01 & \texttt{deepseek-ai/DeepSeek-R1-0528} & \multirow{2}{*}{vLLM} \\
& DeepSeek-V3.1 & 2025-08 & \texttt{deepseek-ai/DeepSeek-V3.1} & \\
\noalign{\vskip 0.4ex}\hdashline\noalign{\vskip 0.4ex}
\multirow{5}{*}{Qwen Team}
& Qwen3-Coder-480B-A35B-Ins. & 2025-07 & \texttt{Qwen/Qwen3-Coder-480B-A35B-Instruct} & \multirow{5}{*}{vLLM} \\
& Qwen3-235B-A22B-Ins. & 2025-07 & \texttt{Qwen/Qwen3-235B-A22B-Instruct-2507} & \\
& Qwen3-235B-A22B-Think. & 2025-07 & \texttt{Qwen/Qwen3-235B-A22B-Thinking-2507-FP8} & \\
& Qwen3-Next-80B-A3B-Ins. & 2025-09 & \texttt{Qwen/Qwen3-Next-80B-A3B-Instruct} & \\
& Qwen3-Next-80B-A3B-Think. & 2025-09 & \texttt{Qwen/Qwen3-Next-80B-A3B-Thinking} & \\
\noalign{\vskip 0.4ex}\hdashline\noalign{\vskip 0.4ex}
\multirow{2}{*}{Meta AI}
& Llama-4-Maverick-17B-128E-Ins. & 2025-04 & \texttt{meta-llama/Llama-4-Maverick-17B-128E-Instruct} & \multirow{2}{*}{vLLM} \\
& Llama-4-Scout-17B-16E-Ins. & 2025-04 & \texttt{meta-llama/Llama-4-Scout-17B-16E-Instruct} & \\
\noalign{\vskip 0.4ex}\hdashline\noalign{\vskip 0.4ex}
\multirow{1}{*}{Defog.ai}
& SQLCoder-7B-2 & 2024-02 & \texttt{defog/sqlcoder-7b-2} & vLLM \\
\noalign{\vskip 0.4ex}\hdashline\noalign{\vskip 0.4ex}
\multirow{1}{*}{Google}
& MedGemma-27B & 2025-06 & \texttt{google/medgemma-27b-text-it} & HF \\
\noalign{\vskip 0.4ex}\hdashline\noalign{\vskip 0.4ex}
\multirow{1}{*}{Baichuan}
& Baichuan-M2-32B & 2025-08 & \texttt{baichuan-inc/Baichuan-M2-32B} & vLLM \\
\bottomrule
\end{tabular}
}
\caption{Configuration of models evaluated in \ours. We report official release month and canonical API/HF identifiers when available.}
\label{tab:model_configuration}
\end{table*}

\section{Schema-Hinted Inference Setup}
\label{app:schema-hinted-setup}
This section describes the schema-hinted inference setup.

\paragraph{Scope.} We run GPT-5-mini on the \ours validation set, covering all six clinical domains and the easy, medium, and hard difficulty tiers. The baseline remains the standard CoT pipeline with up to two execution-driven refinements.

\paragraph{Prompt augmentation.} For each query, we construct a hint block from gold artifacts. We parse the reference SQL and extract ICD codes. We also read the header row of the reference result table to obtain the expected output column names. These hints are appended to the standard CoT prompt, instructing the model to include ICD filters and align SELECT aliases to the expected columns. Full schema-hinted CoT prompt templates are shown in \autoref{fig:sql-prompt-schema-hinted} and \autoref{fig:refinement-prompt-schema-hinted}.
\begin{figure}[!h]
\begin{tcolorbox}[colback=black!3!white, colframe=black!70!white, title=SQL Generation Prompt (Schema-Hinted CoT), fontupper=\footnotesize, fonttitle=\footnotesize]

You are a clinical data analyst expert specializing in the MIMIC-IV database. Your goal is to produce a correct BigQuery SQL query for the question below.\\

Constraints:\\
- Target platform: Google BigQuery.\\
- Use the correct datasets:\\
\texttt{physionet-data.mimiciv\_3\_1\_hosp}, \texttt{physionet-data.mimiciv\_3\_1\_icu}.\\

MIMIC-IV Schema Reference (HOSP + ICU):\\
\textcolor{blue}{\{schema text\}} \\

Schema-hinted context:\\
Relevant ICD code filters observed in validated SQL examples:\\
\textcolor{blue}{- \{ICD code patterns\}} \\
Incorporate the necessary ICD filters or joins when identifying the clinical cohort.\\
Expected column names for the final CSV output:\\
\textcolor{blue}{- \{Column names\}} \\
Align your SELECT aliases with these column names and preserve ordering when applicable.\\

Clinical question:\\
\textcolor{blue}{``\{Question\}''} \\

Your output should be organized in the following two parts:\\
Reasoning:\\
- Think step by step about relevant tables, joins, filters, groupings, and edge cases.\\
- Briefly justify important choices.\\
SQL (wrap the final query in a fenced code block using \texttt{```sql} and \texttt{```}):\\

Think step by step and then generate the complete SQL query.

\end{tcolorbox}
\caption{Schema-hinted Chain-of-Thought SQL generation prompt used in our experiments.}
\label{fig:sql-prompt-schema-hinted}
\end{figure}

\begin{figure}[!h]
\begin{tcolorbox}[colback=black!3!white, colframe=black!70!white, title=Refinement Prompt (Schema-Hinted CoT), fontupper=\footnotesize, fonttitle=\footnotesize]

You are a clinical data analyst expert for the MIMIC-IV dataset. The following SQL failed to execute. Refine it to resolve the issues and better answer the question.\\

Constraints:\\
- Use valid BigQuery SQL.\\
- Use the correct datasets:\\
\texttt{physionet-data.mimiciv\_3\_1\_hosp}, \texttt{physionet-data.mimiciv\_3\_1\_icu}.\\
- Modify only what is necessary; preserve previously correct logic.\\

MIMIC-IV Schema Reference (HOSP + ICU):\\
\textcolor{blue}{\{schema text\}} \\

Schema-hinted context:\\
Relevant ICD code filters observed in validated SQL examples:\\
\textcolor{blue}{- \{ICD code patterns\}} \\
Incorporate the necessary ICD filters or joins when identifying the clinical cohort.\\
Expected column names for the final CSV output:\\
\textcolor{blue}{- \{Column names\}} \\
Align your SELECT aliases with these column names and preserve ordering when applicable.\\

Clinical question:\\
\textcolor{blue}{\{Question\}} \\

Previous SQL attempt (for reference):\\
\textcolor{blue}{\{Previous SQL (provided as a fenced code block)\}} \\

Execution feedback:\\
\textcolor{blue}{\{Execution feedback\}} \\

Your output should be organized in the following two parts:\\
Reasoning:\\
- Step by step, explain the cause of the error and the fix.\\
- Justify key changes briefly.\\
SQL (wrap the final corrected query in a fenced code block using \texttt{```sql} and \texttt{```}):\\

Think step by step and then generate the complete corrected SQL query.

\end{tcolorbox}
\caption{Schema-hinted Chain-of-Thought SQL refinement prompt used in our experiments.}
\label{fig:refinement-prompt-schema-hinted}
\end{figure}

\clearpage
\section{Validation Score Comparisons}
\label{app:validation-sql-comparison}

\begin{figure}[ht]
    \centering
    \includegraphics[width=0.7\linewidth]{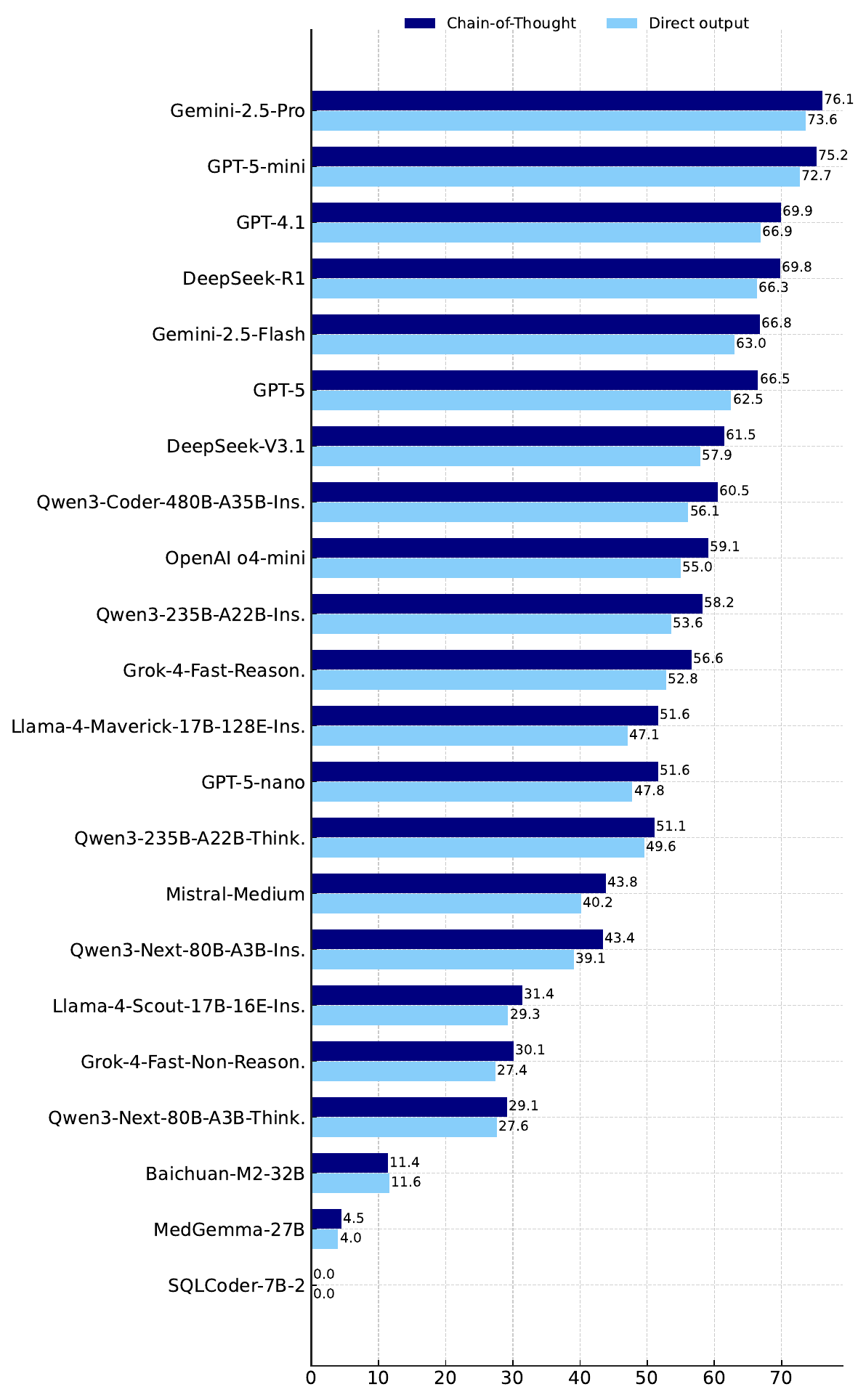}
    \caption{Full validation execution score comparison between Chain-of-Thought reasoning and Direct Output for all models.}
    \label{fig:validation-exec-cot-vs-direct}
\end{figure}

\begin{figure}[ht]
    \centering
    \includegraphics[width=0.90\linewidth]{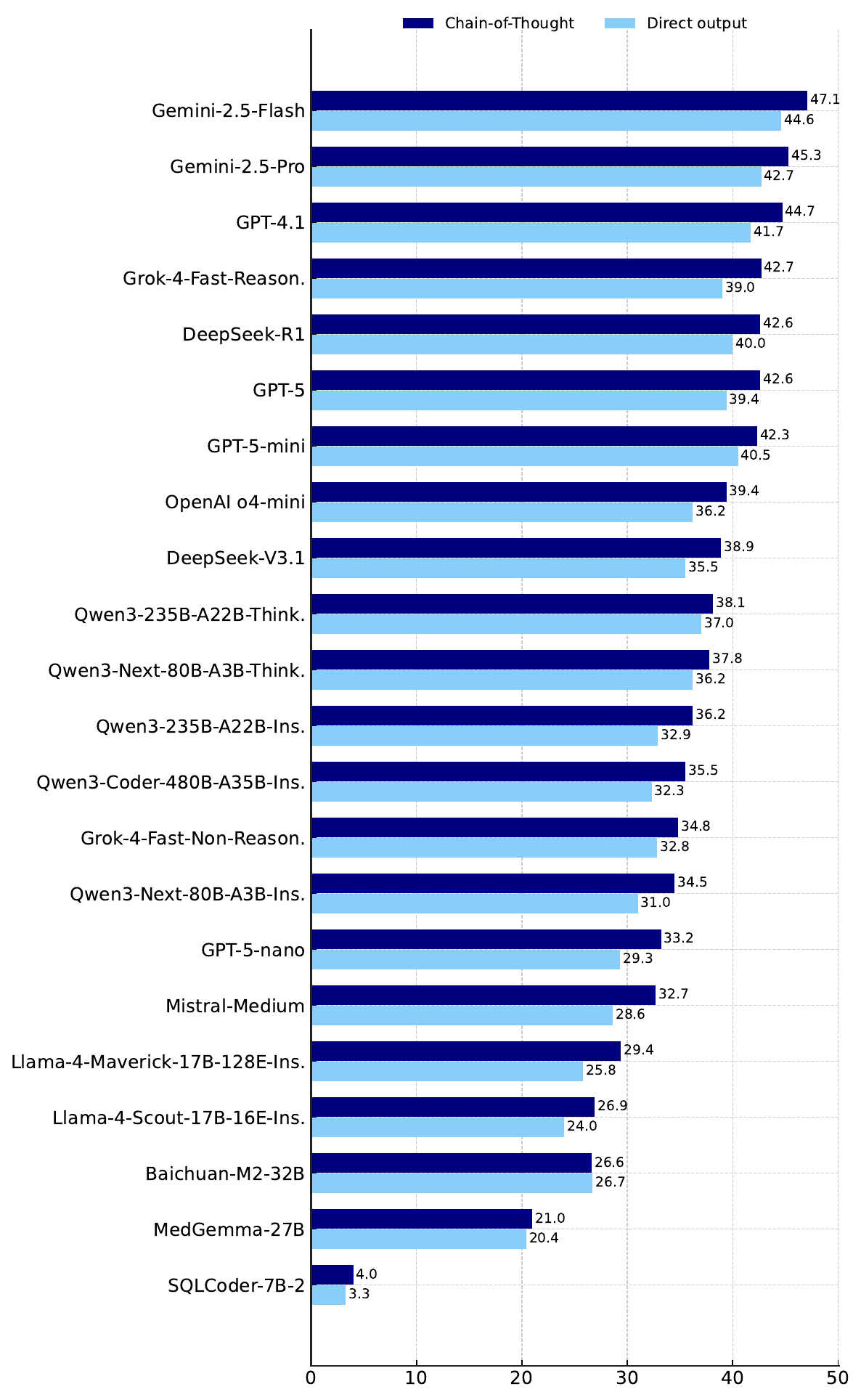}
    \caption{Full validation SQL score comparison between Chain-of-Thought reasoning and Direct Output for all models.}
    \label{fig:validation-sql-cot-vs-direct}
\end{figure}

\clearpage
\section{Scenario-Level Results}
\label{app:scenario-results}
Tables~\ref{tab:scenario1} and \ref{tab:scenario2} extend the main results by reporting scenario-specific SQL and execution score on the \ours validation and test splits.

\paragraph{Scenario abbreviations}\label{app:scenario-abbrev}
Demog.=Patient Demographics and Admissions; Vitals=Vital Signs Monitoring; Labs=Laboratory Results Analysis; Meds=Medication Management; Dx Proc.=Diagnostic Procedures; Dx \& Outc.=Disease Diagnosis and Outcomes.

% TABLE PART 1: Demog., Vitals, Labs.
\begin{table*}[!h]
\centering
\setlength{\tabcolsep}{3pt}
\renewcommand\arraystretch{1.1}
\footnotesize
\resizebox{\textwidth}{!}{%
\begin{tabular}{@{}l*{10}{>{\centering\arraybackslash}p{1.1cm}}@{}}
\toprule[1pt]
 & \multicolumn{6}{c}{\textbf{Test Set}} & \multicolumn{2}{c}{\textbf{Avg. Validation}} & \multicolumn{2}{c}{\textbf{Avg. Test}} \\
\cmidrule(lr){2-7}\noalign{\vskip 1ex}
 & \multicolumn{2}{c}{\textbf{Demog.}} & \multicolumn{2}{c}{\textbf{Vitals}} & \multicolumn{2}{c}{\textbf{Labs}} &  &  &  &  \\
\noalign{\vskip 0.5ex}
\textbf{Model} & \textbf{SQL} & \textbf{Exec} & \textbf{SQL} & \textbf{Exec} & \textbf{SQL} & \textbf{Exec} & \textbf{SQL} & \textbf{Exec} & \textbf{SQL} & \textbf{Exec} \\
\midrule[0.7pt]
\multicolumn{11}{c}{\textbf{\textit{Proprietary Models}}} \\
\noalign{\vskip 1ex}
GPT-5-mini & 50.76 & \cellcolor{red!35}{80.70} & 35.06 & \cellcolor{red!20}{72.65} & 40.70 & \cellcolor{red!35}{65.09} & 42.30 & \cellcolor{red!20}{75.16} & 42.72 & \cellcolor{red!35}{74.67} \\
GPT-5-nano & 44.44 & 49.51 & 31.39 & 43.16 & 35.08 & 49.02 & 33.18 & 51.58 & 36.13 & 52.03 \\
Gemini-2.5-Pro & 53.41 & \cellcolor{red!20}{79.22} & \cellcolor{red!20}{38.06} & \cellcolor{red!35}{73.97} & \cellcolor{red!35}{46.98} & \cellcolor{red!20}{63.66} & \cellcolor{red!20}{45.31} & \cellcolor{red!35}{76.14} & \cellcolor{red!20}{47.28} & \cellcolor{red!20}{73.73} \\
GPT-5 & \cellcolor{red!5}{54.01} & 76.63 & 35.54 & 50.53 & 39.48 & \cellcolor{red!5}{57.89} & 42.62 & 66.52 & 45.93 & \cellcolor{red!5}{68.42} \\
GPT-4.1 & \cellcolor{red!20}{56.15} & \cellcolor{red!5}{77.94} & 34.70 & 58.07 & 43.02 & 48.39 & \cellcolor{red!5}{44.69} & \cellcolor{red!5}{69.92} & \cellcolor{red!5}{46.23} & 67.79 \\
Gemini-2.5-Flash & \cellcolor{red!35}{58.82} & 70.53 & \cellcolor{red!35}{40.35} & \cellcolor{red!5}{63.55} & \cellcolor{red!20}{46.53} & 54.40 & \cellcolor{red!35}{47.06} & 66.75 & \cellcolor{red!35}{47.48} & 65.01 \\
OpenAI o4-mini & 52.15 & 67.35 & \cellcolor{red!5}{35.71} & 46.57 & 34.99 & 48.45 & 39.41 & 59.07 & 41.23 & 59.22 \\
Grok-4-Fast-Reason. & 48.81 & 65.90 & 34.83 & 51.01 & \cellcolor{red!5}{43.74} & 45.50 & 42.67 & 56.58 & 42.78 & 58.46 \\
Grok-4-Fast-Non-Reason. & 49.15 & 37.25 & 34.34 & 31.58 & 30.61 & 21.97 & 34.77 & 30.10 & 39.85 & 30.41 \\
Mistral-Medium & 46.07 & 51.97 & 29.54 & 41.22 & 32.96 & 32.22 & 32.71 & 43.81 & 35.33 & 45.10 \\
\midrule
\multicolumn{11}{c}{\textbf{\textit{Open-source Models}}} \\
\noalign{\vskip 1ex}
DeepSeek-R1 & 49.15 & \cellcolor{red!35}{74.86} & \cellcolor{red!20}{36.24} & \cellcolor{red!35}{62.57} & \cellcolor{red!35}{53.47} & \cellcolor{red!35}{66.91} & \cellcolor{red!35}{42.63} & \cellcolor{red!35}{69.79} & \cellcolor{red!35}{44.91} & \cellcolor{red!35}{69.15} \\
DeepSeek-V3.1 & \cellcolor{red!20}{54.21} & 65.66 & \cellcolor{red!5}{34.36} & \cellcolor{red!20}{54.83} & \cellcolor{red!5}{40.77} & \cellcolor{red!5}{53.66} & \cellcolor{red!20}{38.90} & \cellcolor{red!20}{61.46} & \cellcolor{red!20}{43.19} & \cellcolor{red!20}{60.71} \\
Qwen3-Coder-480B-A35B-Ins. & \cellcolor{red!35}{54.33} & \cellcolor{red!20}{71.97} & 32.53 & 38.48 & 32.92 & 49.78 & 35.54 & \cellcolor{red!5}{60.51} & \cellcolor{red!5}{39.05} & 58.18 \\
Qwen3-235B-A22B-Ins. & 45.67 & \cellcolor{red!5}{69.72} & 31.90 & 41.63 & 38.59 & \cellcolor{red!20}{55.94} & 36.24 & 58.15 & 36.36 & \cellcolor{red!5}{58.63} \\
Qwen3-Next-80B-A3B-Ins. & \cellcolor{red!5}{52.77} & 59.68 & 30.63 & 41.24 & 28.23 & 42.33 & 34.48 & 43.41 & 34.48 & 49.26 \\
Qwen3-235B-A22B-Think. & 45.25 & 59.19 & 33.11 & \cellcolor{red!5}{45.09} & \cellcolor{red!20}{42.74} & 28.94 & \cellcolor{red!5}{38.08} & 51.11 & 38.20 & 48.54 \\
Qwen3-Next-80B-A3B-Think. & 44.56 & 31.36 & \cellcolor{red!35}{38.95} & 15.68 & 35.33 & 19.75 & 37.77 & 29.06 & 38.13 & 27.01 \\
Llama-4-Maverick-17B-128E-Ins. & 43.56 & 55.67 & 25.14 & 42.47 & 22.40 & 37.66 & 29.36 & 51.63 & 29.11 & 47.49 \\
Llama-4-Scout-17B-16E-Ins. & 40.69 & 37.54 & 24.16 & 30.52 & 21.03 & 24.73 & 26.88 & 31.44 & 27.89 & 30.40 \\
Baichuan-M2-32B & 37.17 & 22.68 & 23.60 & 5.29 & 26.57 & 4.22 & 26.60 & 11.40 & 29.97 & 15.27 \\
MedGemma-27B & 32.74 & 9.27 & 18.07 & 0.22 & 16.51 & 2.31 & 21.03 & 4.46 & 20.92 & 4.00 \\
SQLCoder-7B-2 & 9.85 & 0.00 & 2.96 & 0.00 & 5.31 & 0.00 & 3.99 & 0.00 & 5.29 & 0.00 \\
\bottomrule[1pt]
\end{tabular}%
}
\caption{Scenario-level SQL and execution score (\%) on \ours{} validation and test sets. This table lists Demog., Vitals, and Labs scenarios.}
\label{tab:scenario1}
\end{table*}

% TABLE PART 2: Meds, Dx Proc., Dx & Outc.
\begin{table*}[!h]
\centering
\setlength{\tabcolsep}{3pt}
\renewcommand\arraystretch{1.1}
\footnotesize
\resizebox{\textwidth}{!}{%
\begin{tabular}{@{}l*{10}{>{\centering\arraybackslash}p{1.1cm}}@{}}
\toprule[1pt]
 & \multicolumn{6}{c}{\textbf{Test Set}} & \multicolumn{2}{c}{\textbf{Avg. Validation}} & \multicolumn{2}{c}{\textbf{Avg. Test}} \\
\cmidrule(lr){2-7}\noalign{\vskip 1ex}
 & \multicolumn{2}{c}{\textbf{Meds}} & \multicolumn{2}{c}{\textbf{Dx Proc.}} & \multicolumn{2}{c}{\textbf{Dx \& Outc.}} &  &  &  &  \\
\noalign{\vskip 0.5ex}
\textbf{Model} & \textbf{SQL} & \textbf{Exec} & \textbf{SQL} & \textbf{Exec} & \textbf{SQL} & \textbf{Exec} & \textbf{SQL} & \textbf{Exec} & \textbf{SQL} & \textbf{Exec} \\
\midrule[0.7pt]
\multicolumn{11}{c}{\textbf{\textit{Proprietary Models}}} \\
\noalign{\vskip 1ex}
GPT-5-mini & 51.23 & \cellcolor{red!20}{75.06} & 33.72 & \cellcolor{red!35}{75.64} & 44.90 & \cellcolor{red!20}{79.55} & 42.30 & \cellcolor{red!20}{75.16} & 42.72 & \cellcolor{red!35}{74.67} \\
GPT-5-nano & 40.31 & 58.35 & 27.25 & 58.29 & 38.28 & 54.28 & 33.18 & 51.58 & 36.13 & 52.03 \\
Gemini-2.5-Pro & \cellcolor{red!20}{52.39} & \cellcolor{red!35}{77.38} & \cellcolor{red!5}{41.90} & 72.36 & \cellcolor{red!20}{50.97} & 76.43 & \cellcolor{red!20}{45.31} & \cellcolor{red!35}{76.14} & \cellcolor{red!20}{47.28} & \cellcolor{red!20}{73.73} \\
GPT-5 & \cellcolor{red!5}{51.99} & \cellcolor{red!5}{71.07} & \cellcolor{red!35}{45.15} & \cellcolor{red!5}{75.38} & \cellcolor{red!5}{50.05} & \cellcolor{red!35}{80.22} & 42.62 & 66.52 & 45.93 & \cellcolor{red!5}{68.42} \\
GPT-4.1 & \cellcolor{red!35}{53.12} & 71.03 & 37.81 & \cellcolor{red!20}{75.44} & \cellcolor{red!35}{52.74} & \cellcolor{red!5}{77.52} & \cellcolor{red!5}{44.69} & \cellcolor{red!5}{69.92} & \cellcolor{red!5}{46.23} & 67.79 \\
Gemini-2.5-Flash & 48.62 & 67.96 & \cellcolor{red!20}{41.96} & 67.10 & 48.83 & 67.32 & \cellcolor{red!35}{47.06} & 66.75 & \cellcolor{red!35}{47.48} & 65.01 \\
OpenAI o4-mini & 47.45 & 70.01 & 35.00 & 66.04 & 42.53 & 58.19 & 39.41 & 59.07 & 41.23 & 59.22 \\
Grok-4-Fast-Reason. & 40.97 & 67.01 & 38.87 & 63.81 & 49.44 & 58.75 & 42.67 & 56.58 & 42.78 & 58.46 \\
Grok-4-Fast-Non-Reason. & 46.33 & 33.14 & 35.14 & 32.71 & 44.13 & 26.61 & 34.77 & 30.10 & 39.85 & 30.41 \\
Mistral-Medium & 37.97 & 58.60 & 26.62 & 45.59 & 38.95 & 41.97 & 32.71 & 43.81 & 35.33 & 45.10 \\
\midrule
\multicolumn{11}{c}{\textbf{\textit{Open-source Models}}} \\
\noalign{\vskip 1ex}
DeepSeek-R1 & \cellcolor{red!20}{43.39} & \cellcolor{red!35}{72.87} & \cellcolor{red!20}{37.52} & \cellcolor{red!20}{70.16} & \cellcolor{red!35}{49.09} & \cellcolor{red!35}{67.97} & \cellcolor{red!35}{42.63} & \cellcolor{red!35}{69.79} & \cellcolor{red!35}{44.91} & \cellcolor{red!35}{69.15} \\
DeepSeek-V3.1 & \cellcolor{red!35}{44.19} & 62.13 & \cellcolor{red!35}{41.75} & \cellcolor{red!35}{71.46} & \cellcolor{red!20}{44.32} & \cellcolor{red!5}{57.61} & \cellcolor{red!20}{38.90} & \cellcolor{red!20}{61.46} & \cellcolor{red!20}{43.19} & \cellcolor{red!20}{60.71} \\
Qwen3-Coder-480B-A35B-Ins. & 39.24 & \cellcolor{red!20}{66.38} & \cellcolor{red!5}{34.91} & 63.86 & \cellcolor{red!5}{41.01} & \cellcolor{red!20}{59.94} & 35.54 & \cellcolor{red!5}{60.51} & \cellcolor{red!5}{39.05} & 58.18 \\
Qwen3-235B-A22B-Ins. & 37.89 & 60.35 & 31.68 & \cellcolor{red!5}{69.55} & 32.46 & 55.79 & 36.24 & 58.15 & 36.36 & \cellcolor{red!5}{58.63} \\
Qwen3-Next-80B-A3B-Ins. & 31.93 & 52.76 & 31.28 & 57.37 & 32.84 & 43.37 & 34.48 & 43.41 & 34.48 & 49.26 \\
Qwen3-235B-A22B-Think. & \cellcolor{red!5}{42.01} & \cellcolor{red!5}{62.32} & 29.41 & 47.17 & 36.43 & 50.04 & \cellcolor{red!5}{38.08} & 51.11 & 38.20 & 48.54 \\
Qwen3-Next-80B-A3B-Think. & 40.21 & 45.10 & 31.08 & 31.32 & 38.67 & 19.95 & 37.77 & 29.06 & 38.13 & 27.01 \\
Llama-4-Maverick-17B-128E-Ins. & 28.86 & 47.08 & 24.13 & 51.65 & 31.15 & 51.33 & 29.36 & 51.63 & 29.11 & 47.49 \\
Llama-4-Scout-17B-16E-Ins. & 27.65 & 28.91 & 24.68 & 28.72 & 29.75 & 32.37 & 26.88 & 31.44 & 27.89 & 30.40 \\
Baichuan-M2-32B & 36.31 & 25.84 & 26.37 & 18.65 & 30.24 & 16.28 & 26.60 & 11.40 & 29.97 & 15.27 \\
MedGemma-27B & 16.87 & 2.48 & 22.52 & 9.05 & 19.70 & 1.35 & 21.03 & 4.46 & 20.92 & 4.00 \\
SQLCoder-7B-2 & 5.24 & 0.00 & 4.04 & 0.00 & 4.54 & 0.00 & 3.99 & 0.00 & 5.29 & 0.00 \\
\bottomrule[1pt]
\end{tabular}%
}
\caption{Scenario-level SQL and execution score (\%) on \ours{} validation and test sets. This table lists Meds, Dx Proc., and Dx \& Outc. scenarios.}
\label{tab:scenario2}
\end{table*}

\clearpage
\section{Annotation Interface}
\label{app:annotation-interface}
We provide the graphical interface that annotators use while labeling \ours samples, alongside the JSON file that is exported after an annotation is submitted. The pairing highlights how rubric items are surfaced during labeling and then captured in the structured log for future evaluation.

\begin{figure}[ht]
    \centering
    \includegraphics[width=\linewidth]{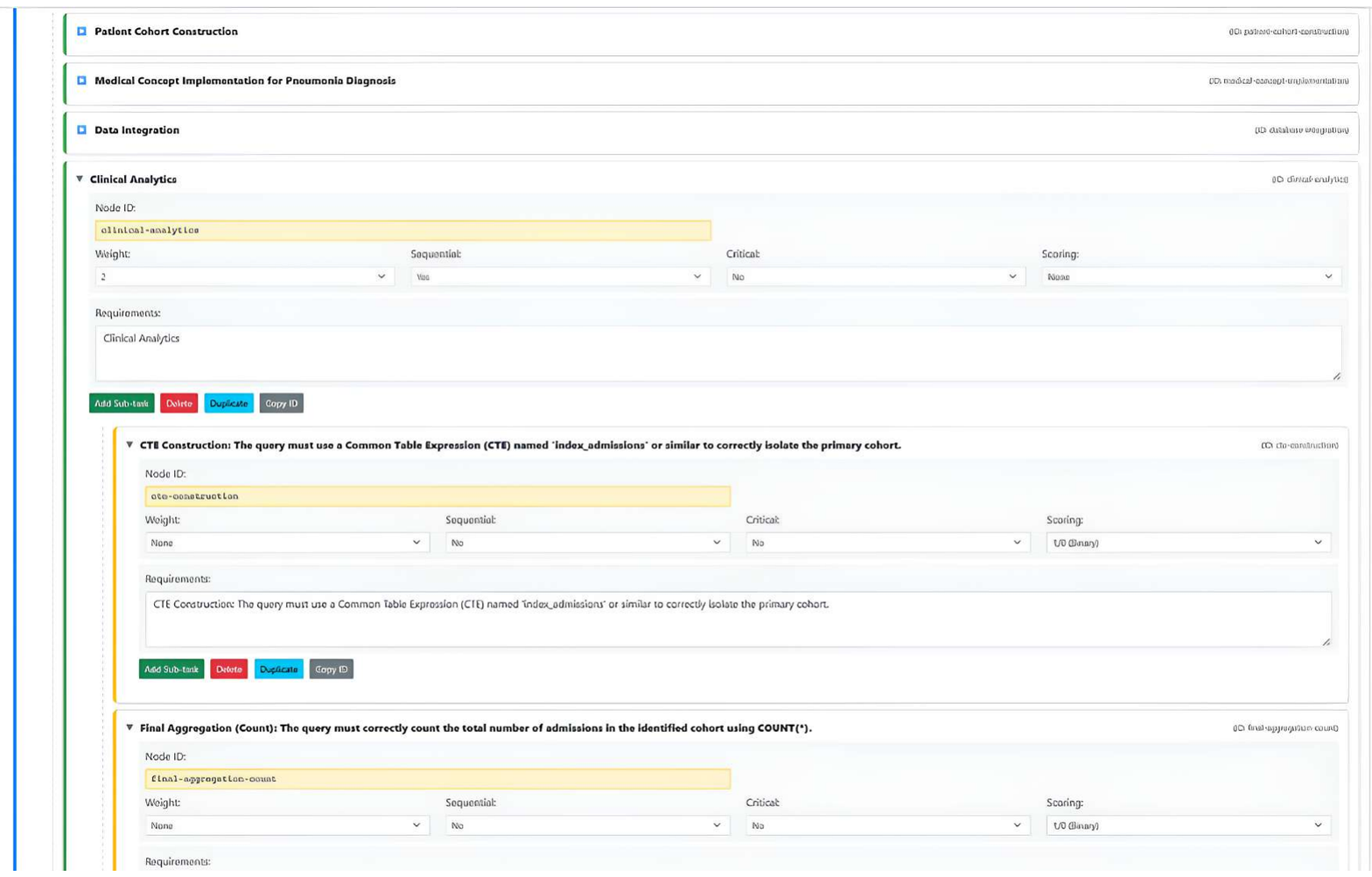}
    \caption{Graphical annotation interface used by annotators when labeling a single \ours sample.}
    \label{fig:annotation-gui}
\end{figure}

\begin{figure}[ht]
    \centering
    \includegraphics[width=\linewidth]{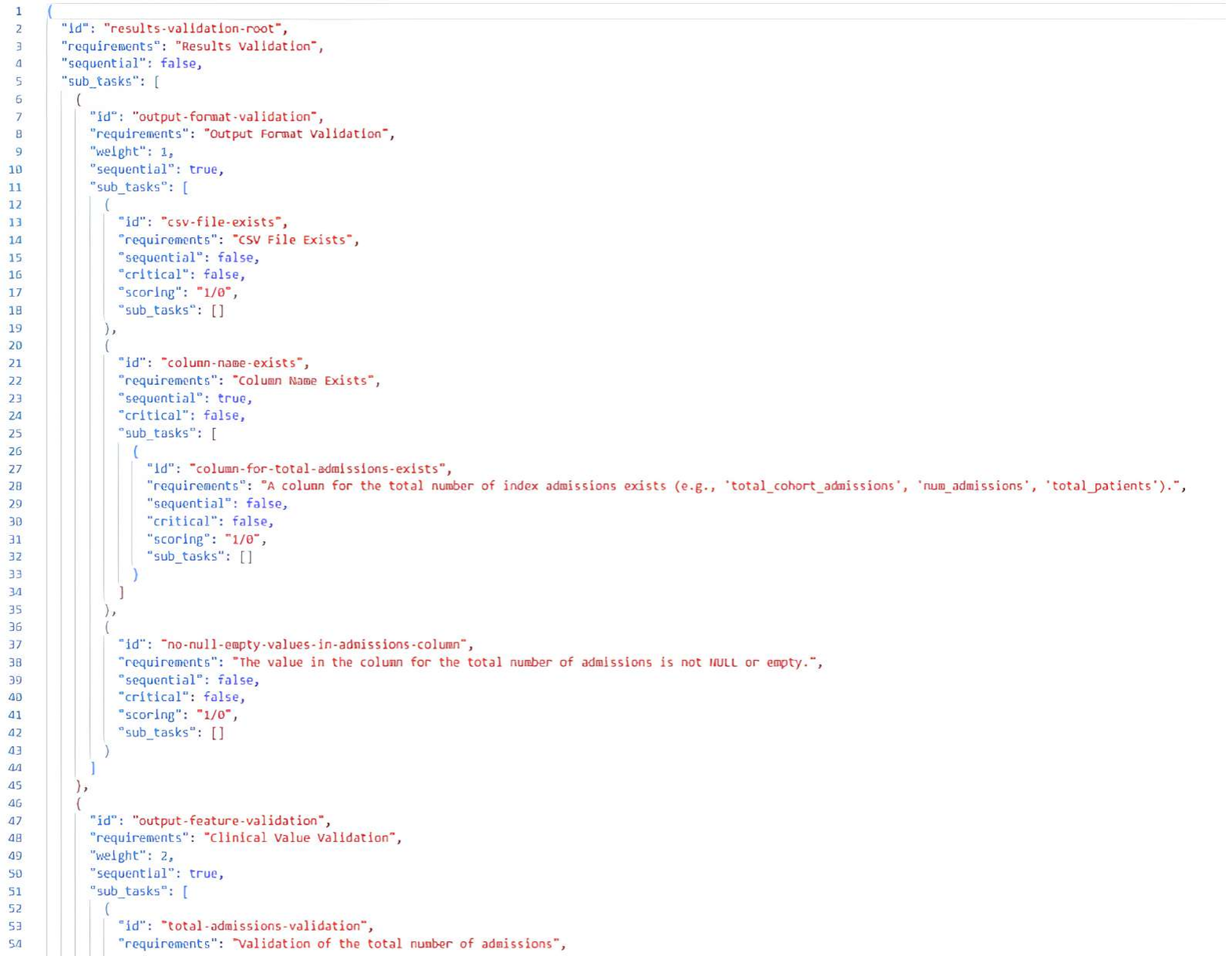}
    \caption{JSON export generated from the completed annotation, preserving rubric selections and metadata.}
    \label{fig:annotation-json}
\end{figure}

\clearpage
\section{Execution Success Rates}
\label{app:execution-success-rates}
We benchmark execution reliability across all \ours scenarios using the test split. Table~\ref{tab:execution-success-rate} reports the queries that executed without errors for every model and scenario, while Table~\ref{tab:attempt-success-rate-a} and Table~\ref{tab:attempt-success-rate-b} break down execution success by the initial query (A1) and up to two refinement attempts (A2 and A3).

\begin{table*}[!h]
\centering
\renewcommand\arraystretch{1.1} % row spacing like the example
\footnotesize
\begin{tabular}{>{\raggedright\arraybackslash}p{5.6cm} *{6}{>{\centering\arraybackslash}p{1.15cm}}}
\toprule[1pt]
\textbf{Model} & \multicolumn{6}{c}{\textbf{Scenarios (\%)}} \\
\cline{2-7}\noalign{\vskip 0.6ex}
 & \textbf{Demog.} & \textbf{Vitals} & \textbf{Labs} & \textbf{Meds} & \textbf{Dx Proc.} & \textbf{Dx \& Outc.} \\
\midrule[0.7pt]
GPT-5-mini & 100.0 & 98.7 & 92.3 & 93.2 & 97.1 & 90.5 \\
GPT-5-nano & 91.5 & 89.5 & 97.4 & 85.1 & 94.3 & 75.7 \\
Gemini-2.5-Pro & 97.2 & 97.4 & 91.0 & 97.3 & 95.7 & 90.5 \\
GPT-5 & 100.0 & 96.1 & 94.9 & 100.0 & 97.1 & 100.0 \\
GPT-4.1 & 98.6 & 94.7 & 96.2 & 93.2 & 98.6 & 93.2 \\
Gemini-2.5-Flash & 93.0 & 85.5 & 82.1 & 93.2 & 87.1 & 81.1 \\
OpenAI o4-mini & 100.0 & 94.7 & 93.6 & 97.3 & 97.1 & 86.5 \\
Grok-4-Fast-Reason. & 97.2 & 93.4 & 76.9 & 90.5 & 84.3 & 73.0 \\
Grok-4-Fast-Non-Reason. & 59.2 & 55.3 & 21.8 & 33.8 & 34.3 & 18.9 \\
Mistral-Medium & 83.1 & 89.5 & 70.5 & 90.5 & 82.9 & 74.3 \\
DeepSeek-R1 & 98.6 & 94.7 & 98.7 & 95.9 & 94.3 & 89.2 \\
DeepSeek-V3.1 & 95.8 & 93.4 & 88.5 & 87.8 & 98.6 & 82.4 \\
Qwen3-Coder-480B-A35B-Ins. & 98.6 & 92.1 & 93.6 & 95.9 & 100.0 & 93.2 \\
Qwen3-235B-A22B-Ins. & 98.6 & 85.5 & 93.6 & 86.5 & 91.4 & 64.9 \\
Qwen3-Next-80B-A3B-Ins. & 80.3 & 71.1 & 65.4 & 74.3 & 70.0 & 47.3 \\
Qwen3-235B-A22B-Think. & 100.0 & 92.1 & 94.9 & 95.9 & 85.7 & 79.7 \\
Qwen3-Next-80B-A3B-Think. & 56.3 & 57.9 & 50.0 & 77.0 & 70.0 & 47.3 \\
Llama-4-Maverick-17B-128E-Ins. & 88.7 & 77.6 & 75.6 & 77.0 & 80.0 & 79.7 \\
Llama-4-Scout-17B-16E-Ins. & 49.3 & 55.3 & 59.0 & 47.3 & 51.4 & 44.6 \\
Baichuan-M2-32B & 62.0 & 52.6 & 48.7 & 55.4 & 51.4 & 47.3 \\
MedGemma-27B & 32.4 & 17.1 & 29.5 & 20.3 & 28.6 & 12.2 \\
SQLCoder-7B-2 & 0.0 & 1.3 & 1.3 & 0.0 & 1.4 & 2.7 \\
\bottomrule[1pt]
\end{tabular}
\caption{Execution success rate (\%) per model and scenario on the test split. Abbreviations defined in \autoref{app:scenario-abbrev}.}
\label{tab:execution-success-rate}
\end{table*}

\begin{table*}[!h]
\centering
\renewcommand\arraystretch{1.1}
\scriptsize
\setlength{\tabcolsep}{2pt}
\begin{tabular}{l *{9}{>{\centering\arraybackslash}p{1.15cm}}}
\toprule[1pt]
\textbf{Model} & \multicolumn{9}{c}{\textbf{Scenarios (\%)}} \\
\cline{2-10}\noalign{\vskip 0.6ex}
 & \multicolumn{3}{c}{\textbf{Demog.}} & \multicolumn{3}{c}{\textbf{Vitals}} & \multicolumn{3}{c}{\textbf{Labs}} \\
\noalign{\vskip 0.3ex}
 & \textbf{A1} & \textbf{A2} & \textbf{A3}
 & \textbf{A1} & \textbf{A2} & \textbf{A3}
 & \textbf{A1} & \textbf{A2} & \textbf{A3} \\
\midrule[0.7pt]
GPT-5-mini & 84.5 & 14.1 & 1.4 & 63.2 & 27.6 & 7.9 & 41.0 & 41.0 & 10.3 \\
GPT-5-nano & 60.6 & 23.9 & 7.0 & 56.6 & 22.4 & 10.5 & 53.8 & 32.1 & 11.5 \\
Gemini-2.5-Pro & 88.7 & 5.6 & 2.8 & 81.6 & 13.2 & 2.6 & 71.8 & 16.7 & 2.6 \\
GPT-5 & 85.9 & 8.5 & 5.6 & 73.7 & 15.8 & 6.6 & 73.1 & 12.8 & 9.0 \\
GPT-4.1 & 78.9 & 18.3 & 1.4 & 72.4 & 19.7 & 2.6 & 74.4 & 15.4 & 6.4 \\
Gemini-2.5-Flash & 62.0 & 19.7 & 11.3 & 47.4 & 26.3 & 11.8 & 33.3 & 30.8 & 17.9 \\
OpenAI o4-mini & 73.2 & 25.4 & 1.4 & 57.9 & 26.3 & 10.5 & 59.0 & 26.9 & 7.7 \\
Grok-4-Fast-Reason. & 43.7 & 36.6 & 16.9 & 38.2 & 38.2 & 17.1 & 9.0 & 33.3 & 34.6 \\
Grok-4-Fast-Non-Reason. & 19.7 & 23.9 & 15.5 & 30.3 & 11.8 & 13.2 & 3.8 & 11.5 & 6.4 \\
Mistral-Medium & 62.0 & 12.7 & 8.5 & 51.3 & 25.0 & 13.2 & 43.6 & 17.9 & 9.0 \\
DeepSeek-R1 & 78.9 & 11.3 & 8.5 & 50.0 & 36.8 & 7.9 & 69.2 & 24.4 & 5.1 \\
DeepSeek-V3.1 & 67.6 & 22.5 & 5.6 & 71.1 & 14.5 & 7.9 & 70.5 & 12.8 & 5.1 \\
Qwen3-Coder-480B-A35B-Ins. & 77.5 & 16.9 & 4.2 & 69.7 & 19.7 & 2.6 & 71.8 & 16.7 & 5.1 \\
Qwen3-235B-A22B-Ins. & 69.0 & 23.9 & 5.6 & 47.4 & 31.6 & 6.6 & 52.6 & 29.5 & 11.5 \\
Qwen3-Next-80B-A3B-Ins. & 32.4 & 26.8 & 21.1 & 28.9 & 30.3 & 11.8 & 15.4 & 29.5 & 20.5 \\
Qwen3-235B-A22B-Think. & 42.3 & 52.1 & 5.6 & 47.4 & 32.9 & 11.8 & 42.3 & 47.4 & 5.1 \\
Qwen3-Next-80B-A3B-Think. & 42.3 & 11.3 & 2.8 & 35.5 & 14.5 & 7.9 & 30.8 & 16.7 & 2.6 \\
Llama-4-Maverick-17B-128E-Ins. & 57.7 & 28.2 & 2.8 & 47.4 & 18.4 & 11.8 & 48.7 & 23.1 & 3.8 \\
Llama-4-Scout-17B-16E-Ins. & 21.1 & 19.7 & 8.5 & 22.4 & 22.4 & 10.5 & 16.7 & 24.4 & 17.9 \\
Baichuan-M2-32B & 25.4 & 21.1 & 15.5 & 14.5 & 27.6 & 10.5 & 19.2 & 19.2 & 10.3 \\
MedGemma-27B & 15.5 & 5.6 & 11.3 & 10.5 & 6.6 & 0.0 & 19.2 & 9.0 & 1.3 \\
SQLCoder-7B-2 & 0.0 & 0.0 & 0.0 & 0.0 & 0.0 & 1.3 & 0.0 & 0.0 & 1.3 \\
\bottomrule[1pt]
\end{tabular}
\caption{Per-attempt execution success rate (\%). This table lists Demog., Vitals, Labs. Attempts A1, A2 and A3 correspond to the initial query and up to two refinements.}
\label{tab:attempt-success-rate-a}
\end{table*}

\begin{table*}[!h]
\centering
\renewcommand\arraystretch{1.1}
\scriptsize
\setlength{\tabcolsep}{2pt}
\begin{tabular}{l *{9}{>{\centering\arraybackslash}p{1.15cm}}}
\toprule[1pt]
\textbf{Model} & \multicolumn{9}{c}{\textbf{Scenarios (\%)}} \\
\cline{2-10}\noalign{\vskip 0.6ex}
 & \multicolumn{3}{c}{\textbf{Meds}} & \multicolumn{3}{c}{\textbf{Dx Proc.}} & \multicolumn{3}{c}{\textbf{Dx \& Outc.}} \\
\noalign{\vskip 0.3ex}
 & \textbf{A1} & \textbf{A2} & \textbf{A3}
 & \textbf{A1} & \textbf{A2} & \textbf{A3}
 & \textbf{A1} & \textbf{A2} & \textbf{A3} \\
\midrule[0.7pt]
GPT-5-mini & 41.9 & 37.8 & 13.5 & 51.4 & 37.1 & 8.6 & 44.6 & 32.4 & 13.5 \\
GPT-5-nano & 35.1 & 40.5 & 9.5 & 47.1 & 40.0 & 7.1 & 29.7 & 32.4 & 13.5 \\
Gemini-2.5-Pro & 79.7 & 14.9 & 2.7 & 75.7 & 18.6 & 1.4 & 74.3 & 9.5 & 6.8 \\
GPT-5 & 71.6 & 25.7 & 2.7 & 84.3 & 10.0 & 2.9 & 78.4 & 13.5 & 8.1 \\
GPT-4.1 & 73.0 & 20.3 & 0.0 & 75.7 & 21.4 & 1.4 & 78.4 & 13.5 & 1.4 \\
Gemini-2.5-Flash & 51.4 & 31.1 & 10.8 & 45.7 & 30.0 & 11.4 & 35.1 & 29.7 & 16.2 \\
OpenAI o4-mini & 59.5 & 33.8 & 4.1 & 57.1 & 37.1 & 2.9 & 56.8 & 25.7 & 4.1 \\
Grok-4-Fast-Reason. & 24.3 & 54.1 & 12.2 & 32.9 & 37.1 & 14.3 & 16.2 & 28.4 & 28.4 \\
Grok-4-Fast-Non-Reason. & 14.9 & 9.5 & 9.5 & 20.0 & 4.3 & 10.0 & 4.1 & 5.4 & 9.5 \\
Mistral-Medium & 48.6 & 31.1 & 10.8 & 48.6 & 28.6 & 5.7 & 47.3 & 17.6 & 9.5 \\
DeepSeek-R1 & 59.5 & 31.1 & 5.4 & 70.0 & 18.6 & 5.7 & 51.4 & 27.0 & 10.8 \\
DeepSeek-V3.1 & 56.8 & 27.0 & 4.1 & 75.7 & 17.1 & 5.7 & 55.4 & 8.1 & 18.9 \\
Qwen3-Coder-480B-A35B-Ins. & 67.6 & 17.6 & 10.8 & 61.4 & 32.9 & 5.7 & 60.8 & 23.0 & 9.5 \\
Qwen3-235B-A22B-Ins. & 35.1 & 33.8 & 17.6 & 60.0 & 28.6 & 2.9 & 32.4 & 24.3 & 8.1 \\
Qwen3-Next-80B-A3B-Ins. & 27.0 & 21.6 & 25.7 & 21.4 & 30.0 & 18.6 & 21.6 & 13.5 & 12.2 \\
Qwen3-235B-A22B-Think. & 27.0 & 59.5 & 9.5 & 20.0 & 52.9 & 12.9 & 28.4 & 43.2 & 8.1 \\
Qwen3-Next-80B-A3B-Think. & 31.1 & 28.4 & 17.6 & 41.4 & 21.4 & 7.1 & 35.1 & 9.5 & 2.7 \\
Llama-4-Maverick-17B-128E-Ins. & 45.9 & 20.3 & 10.8 & 61.4 & 17.1 & 1.4 & 51.4 & 20.3 & 8.1 \\
Llama-4-Scout-17B-16E-Ins. & 8.1 & 18.9 & 20.3 & 11.4 & 18.6 & 21.4 & 8.1 & 23.0 & 13.5 \\
Baichuan-M2-32B & 21.6 & 25.7 & 8.1 & 21.4 & 24.3 & 5.7 & 18.9 & 16.2 & 12.2 \\
MedGemma-27B & 12.2 & 4.1 & 4.1 & 21.4 & 5.7 & 1.4 & 2.7 & 2.7 & 6.8 \\
SQLCoder-7B-2 & 0.0 & 0.0 & 0.0 & 0.0 & 0.0 & 1.4 & 0.0 & 0.0 & 2.7 \\
\bottomrule[1pt]
\end{tabular}
\caption{Per-attempt execution success rate (\%). This table lists Meds, Dx Proc., Dx \& Outc.}
\label{tab:attempt-success-rate-b}
\end{table*}

\clearpage
\section{Human--GPT Agreement Study}
\label{app:human-gpt-agreement}
Because our rubric-based evaluation relies on GPT-5 as the judge, we run a human--GPT agreement study to validate the reliability of its rubric decisions. We randomly sample 100 \ours validation/test examples and re-score them with two medically trained annotators from the same pool that constructs the dataset and rubrics. For each example, both annotators independently evaluate rubric leaf nodes for the SQL and results trees using the same annotation guidelines, without access to GPT-5 scores or each other's labels.

We compare GPT-5's leaf-level decisions and aggregated pass/fail outcomes against each annotator, and also compare the two annotators with each other. Table~\ref{tab:human-gpt-agreement} reports agreement rates at the leaf level and at the final pass/fail level for both SQL and results. Disagreements are largely concentrated in borderline, partial-credit cases, and do not materially change relative model rankings, supporting the use of GPT-5 as a reliable and scalable rubric judge for our tree-structured evaluation.

\begin{table*}[!h]
\centering
\renewcommand\arraystretch{1.1}
\footnotesize
\begin{tabular}{>{\raggedright\arraybackslash}p{4.2cm} *{4}{>{\centering\arraybackslash}p{2.2cm}}}
\toprule[1pt]
\textbf{Pair} & \textbf{Leaf-level agreement (SQL, \%)} & \textbf{Leaf-level agreement (results, \%)} & \textbf{Pass/fail agreement (SQL, \%)} & \textbf{Pass/fail agreement (results, \%)} \\
\midrule[0.7pt]
GPT-5 vs. Annotator 1 & 83.4 & 87.1 & 90.2 & 92.3 \\
GPT-5 vs. Annotator 2 & 82.1 & 85.9 & 88.7 & 91.0 \\
Annotator 1 vs. Annotator 2 & 86.8 & 89.5 & 92.4 & 94.1 \\
\bottomrule[1pt]
\end{tabular}
\caption{Human--GPT agreement on 100 randomly sampled \ours examples. Leaf-level agreement compares rubric leaf decisions, while pass/fail agreement compares aggregated outcomes for SQL and results.}
\label{tab:human-gpt-agreement}
\end{table*}

\clearpage
\section{Inter-Annotator Agreement and Reconciliation Protocols}
\label{app:iaa-reconciliation}
The trustworthiness of an evaluation benchmark is essential for expert-domain assessment~\cite{ke2025early, ke2025stable, ouyang2024learn}. To this end, 
\ours uses validator-based quality control to stabilize scenario design, gold SQL, and rubric leaves. Each scenario is created by a primary annotator and independently reviewed by a validator. Validators re-run the SQL in BigQuery, inspect result tables, and check rubric trees for coverage and correctness. For each item they record one of three outcomes: accept as-is, accept with minor edits, or major revision or reject. Across the benchmark, validators accept 87\% of items as-is, request 9\% minor edits (e.g., tightening a time window, adjusting an inclusion criterion, or clarifying a rubric leaf), and request 4\% major revision or rejection.

To quantify inter-annotator agreement (IAA), we additionally sample 50 validation/test scenarios and ask a second annotator, distinct from both the original annotator and the validator, to re-annotate them independently at two levels. For gold SQL, the second annotator writes a fresh query based only on the natural-language description and schema. We canonicalize both SQL queries and execute them on the redacted BigQuery database. In 46/50 cases (92\%), the two queries produce identical result tables (up to row/column ordering). In 3/50 cases (6\%), the results differ only by small numerical variations and are treated as near-miss agreements. In the remaining 1/50 case (2\%), the second annotator interprets the clinical question differently, yielding a clinically distinct cohort. Exact SQL-level agreement is 92\%; counting near-miss cases as acceptable alternatives yields 98\%.

For rubric leaves, an independent annotator reconstructs the SQL and results rubrics using shared templates and guidelines. We treat each rubric as a set of atomic checks (leaf presence and criticality) and compare the two annotators' trees. Raw agreement on leaf presence and criticality is 91\% with Cohen's $\kappa=0.82$, and the critical-first aggregation yields a pass/fail agreement of 47/50 scenarios (94\%). Table~\ref{tab:iaa-reconciliation} summarizes these reconciliation and IAA statistics.

\begin{table*}[!h]
\centering
\renewcommand\arraystretch{1.1}
\footnotesize
\begin{tabular}{>{\raggedright\arraybackslash}p{8.2cm} >{\centering\arraybackslash}p{3.2cm}}
\toprule[1pt]
\textbf{Statistic} & \textbf{Value} \\
\midrule[0.7pt]
\multicolumn{2}{l}{\textbf{Validator-based quality control (all \ours items)}} \\
Accept as-is & 87\% \\
Accept with minor edits & 9\% \\
Major revision / reject & 4\% \\
\midrule
\multicolumn{2}{l}{\textbf{Double-annotation study (n=50 validation/test scenarios)}} \\
SQL exact match (identical results) & 46/50 (92\%) \\
SQL near-miss (minor numeric differences) & 3/50 (6\%) \\
SQL distinct (clinically different) & 1/50 (2\%) \\
Rubric leaf agreement (presence + criticality) & 91\% (Cohen's $\kappa=0.82$) \\
Rubric pass/fail agreement & 47/50 (94\%) \\
\bottomrule[1pt]
\end{tabular}
\caption{Inter-annotator agreement (IAA) and reconciliation statistics for \ours. Validator outcomes are reported on the full benchmark; double-annotation statistics are computed on a 50-scenario validation/test sample.}
\label{tab:iaa-reconciliation}
\end{table*}

\clearpage
\section{Reconciling Execution Passes with SQL Analysis Failures}
\label{app:sql-exec-reconciliation}
SQL analysis and execution scoring capture complementary notions of correctness: SQL analysis checks whether the query encodes the intended cohort logic, whereas execution scoring evaluates whether the observed results are clinically plausible and consistent with the gold answer. At the model level, these signals are strongly aligned. Across all models in the main tables, the Pearson correlation between SQL Score and Execution Score is $r=0.8597$, and the Spearman rank correlation is $\rho=0.8554$, indicating near-identical model rankings.

To diagnose divergence cases, we run a targeted outlier study. We identify model--dataset points with a large absolute gap between SQL Score and Execution Score (at least 20 percentage points) and randomly sample 40 such outliers for manual inspection of both SQL and result tables. Most outliers (about 95\%) exhibit high execution scores but low SQL scores, reflecting partial or imprecise cohort logic (e.g., missing secondary exclusion criteria, incomplete temporal logic, or incorrect aggregation granularity). In these cases, execution remains high because the resulting aggregates stay close to reference values despite logical deviations. The remaining outliers (about 5\%) show high SQL scores but lower execution scores, typically when correct logic yields clinically implausible values in narrow subgroups. These findings indicate that discrepancies arise from meaningful error types rather than rubric misalignment.

Table~\ref{tab:sql-exec-reconciliation} summarizes the correlation statistics and outlier breakdown.

\begin{table*}[!h]
\centering
\renewcommand\arraystretch{1.1}
\footnotesize
\begin{tabular}{>{\raggedright\arraybackslash}p{8.6cm} >{\centering\arraybackslash}p{3.0cm}}
\toprule[1pt]
\textbf{Statistic} & \textbf{Value} \\
\midrule[0.7pt]
Pearson correlation (SQL Score vs. Execution Score) & $r=0.8597$ \\
Spearman rank correlation (SQL Score vs. Execution Score) & $\rho=0.8554$ \\
Outlier threshold ($|$SQL -- Execution$| \ge 20$ points) & 40 cases sampled \\
High Execution / low SQL among outliers & $\approx 95\%$ \\
High SQL / low Execution among outliers & $\approx 5\%$ \\
\bottomrule[1pt]
\end{tabular}
\caption{Reconciliation of SQL analysis and execution scoring. Correlations are computed across all models in the main tables; outlier statistics are based on 40 sampled large-gap cases.}
\label{tab:sql-exec-reconciliation}
\end{table*}

\clearpage
\section{Annotation Guideline}
\label{app:annotation-guideline}

\subsection{Part I: Annotation Guidelines}

\subsubsection{Overview}

Each example in our benchmark consists of:

\begin{itemize}
  \item The type of realistic clinical scenario, representing one of the six scenario types defined in the paper
  \item A natural language clinical question that requires a database query to be solved
  \item A gold standard SQL query that accurately translates the clinical question into executable database operations
  \item A results table in CSV format, generated from the execution of the gold-standard SQL query
  \item An evaluation guideline for evaluating the accuracy of the SQL queries and executed results generated by other models
\end{itemize}

You will be assigned the following three sequential roles:

\begin{enumerate}
  \item \textbf{Query Annotator:} Develop the clinical scenario, provide the patient context, and formulate the natural language question.
  \item \textbf{SQL Annotator:} Analyze database requirements, construct the gold-standard SQL implementation, execute the SQL queries, analyze the results table, and annotate the specific database tables that were used and key features in the results table (columns, values).
  \item \textbf{Evaluation Guideline Annotator:} For each example, create a tailored guideline for evaluating the SQL query and validating its results.
\end{enumerate}

While the annotation interface will guide you, it is essential to follow the instructions carefully.

\subsection{Step 1: Query Annotation}

As the \textbf{Query Annotator}, your task is to create realistic clinical scenarios and formulate natural language questions that require database queries to solve. You will develop questions that reflect authentic clinical decision-making processes and information needs.

\subsubsection{Clinical Scenario Development}

\begin{enumerate}
  \item \textbf{Select and Understand Your Assigned Patient Information and Scenario Type}
  \begin{itemize}
    \item You will be assigned one of six clinical scenario types and a specific patient from the MIMIC-IV database, selected based on that scenario type for filtering: Patient Demographics \& Admissions, Vital Signs Monitoring, Laboratory Results Analysis, Medication Management, Diagnostic Procedures, or Disease Diagnosis \& Outcomes.
    \item Before you begin, familiarize yourself with the MIMIC tables and understand the information for the specific patient assigned, and review the scenario definition.
  \end{itemize}
\end{enumerate}

\subsubsection{Natural Language Question Formulation}

\begin{enumerate}
  \item \textbf{Craft the Clinical Question}
  \begin{itemize}
    \item Write a natural language question that a \textbf{physician} would realistically ask, given the information of the provided patient.
    \item Ensure the question requires database querying and cannot be answered through simple observation.
    \item Use appropriate medical terminology while maintaining clarity.
  \end{itemize}
\end{enumerate}

\subsection{Step 2: SQL Annotation}

Your task is to analyze the clinical question, identify database requirements, construct the gold-standard SQL query, and document the implementation details.

\subsubsection{Database Analysis and Schema Mapping}

\begin{enumerate}
  \item \textbf{Clinical Question Analysis}
  \begin{itemize}
    \item Identify the specific clinical data elements required to answer the question.
  \end{itemize}
  \item \textbf{MIMIC-IV Database Mapping}
  \begin{itemize}
    \item Identify all MIMIC-IV tables required to answer the question.
    \item Map clinical concepts in the question to specific database tables and columns.
    \item Determine relevant clinical thresholds, normal ranges, and medical domain knowledge.
    \item Map clinical conditions, procedures, and interventions mentioned in the question to their corresponding ICD codes.
  \end{itemize}
\end{enumerate}

\subsubsection{Gold-Standard SQL Annotation}

\begin{enumerate}
  \item \textbf{SQL Query Development}
  \begin{itemize}
    \item Write a complete, executable SQL query that accurately answers the clinical question.
    \item Ensure the query handles edge cases and data quality issues common in clinical databases.
  \end{itemize}
  \item \textbf{Query Execution and Result Generation}
  \begin{itemize}
    \item Execute the SQL query against the MIMIC-IV database.
    \item Generate the complete results table in CSV format.
    \item Verify that results are clinically meaningful and interpretable.
  \end{itemize}
\end{enumerate}

\subsection{Evaluation Guideline Annotation}

As the \textbf{Evaluation Guideline Annotator}, your responsibility is to create comprehensive criteria for evaluating SQL queries and executed results generated by other models attempting to answer the clinical question.

\subsubsection{Understanding Evaluation Rubric Structure}

Before building evaluation rubrics, you must understand the foundational concepts that govern how evaluation scores are calculated in our benchmark system.

\paragraph{Critical vs. Non-Critical Nodes}

Our evaluation system employs two types of assessment nodes:

\begin{itemize}
  \item \textbf{Critical Nodes} [Critical]: Essential criteria whose failure immediately causes the parent node to fail, regardless of other sibling node performance. Critical nodes represent fundamental requirements that must be satisfied for meaningful evaluation.
  \item \textbf{Non-Critical Nodes}: Allow partial scoring at the parent level. When mixed with critical nodes, non-critical nodes contribute to averaging only after all critical nodes pass.
\end{itemize}

\begin{itemize}
  \item \textbf{Score = 1:} The requirement is fully satisfied. The SQL query or result demonstrates correct and clinically appropriate implementation of this component.
  \item \textbf{Score = 0:} The requirement is not satisfied. The component is missing, incorrectly implemented, or produces clinically invalid output.
\end{itemize}

For example:

\begin{itemize}
  \item If Gender Selection [1] [Critical] and Age Range Selection [1] [Critical] $\rightarrow$ Patient Cohort Construction [1]
  \item If Gender Selection [1] [Critical] and Age Range Selection [0] [Critical] $\rightarrow$ Patient Cohort Construction [0]
  \item If Gender Selection [1] [Critical], Age Range Selection [1] [Critical], and Time Filter [0] (non-critical) $\rightarrow$ Patient Cohort Construction = (0)/1 = 0 (average of non-critical nodes)
\end{itemize}

\paragraph{Sequential Dependencies [sequential]}

Some evaluation nodes are marked as \textbf{sequential}, indicating logical dependencies among child nodes where failure at an earlier step renders subsequent evaluations meaningless. For example, if a SQL query fails to correctly filter the patient cohort, evaluating the aggregation logic becomes pointless.

For example:

\begin{itemize}
  \item If Table Join Logic [1] [sequential] and Key Matching [1] [sequential] $\rightarrow$ Data Integration = (1+1)/2 = 1 (all sequential steps succeed)
  \item If Table Join Logic [0] [sequential] and Key Matching [not evaluated] [sequential] $\rightarrow$ Data Integration = (0)/1 = 0 (sequential failure stops evaluation)
  \item If Table Join Logic [1] [sequential], Key Matching [1] [sequential], and Final Validation [0] [sequential] $\rightarrow$ Data Integration = (1+1+0)/3 = 0.67 (sequential failure after partial evaluation)
\end{itemize}

\paragraph{Weight Assignment [Weight X]}

Each major evaluation category is assigned a weight reflecting its relative importance in the overall assessment. Weights enable proportional scoring where more critical aspects (e.g., patient cohort construction) receive higher influence than secondary considerations.

\subsubsection{Quantified Weight Scale}

Our evaluation framework employs a 3-point weight scale based on clinical importance:

\paragraph{Weight 1: Basic Supportive Criteria}

\begin{itemize}
  \item Represents supplementary evaluation components that provide additional context.
  \item Examples: Output formatting, minor data type handling, non-essential temporal constraints.
\end{itemize}

\begin{verbatim}
-- Output formatting and rounding
SELECT ROUND(AVG(procedure_count), 2) as avg_imaging_procedures

-- Column aliasing for readability
COUNT(DISTINCT pr.icd_code) as procedure_count
\end{verbatim}

Typically assigned to elements that enhance quality but are not fundamental to clinical correctness.

\paragraph{Weight 2: Standard Clinical Requirements}

\begin{itemize}
  \item Represents standard clinical database operations and moderate complexity reasoning.
  \item Examples: Medical concept implementation, aggregation functions, procedure identification.
\end{itemize}

\begin{verbatim}
-- Medical concept implementation - ICD code pattern matching
(pr.icd_version = 10 AND (
  pr.icd_code LIKE 'B%' OR     -- Imaging procedures
  pr.icd_code LIKE '3E0%' OR   -- CT procedures
  pr.icd_code LIKE 'BW%' OR    -- X-ray procedures
  pr.icd_code LIKE 'B3%'       -- Ultrasound procedures
))

-- Aggregation functions for clinical analytics
COUNT(DISTINCT pr.icd_code) as procedure_count
AVG(procedure_count)

-- ICD version handling
(pr.icd_version = 9 AND (
  pr.icd_code LIKE '87%' OR    -- Diagnostic radiology
  pr.icd_code LIKE '88%'       -- Other diagnostic procedures
))
\end{verbatim}

Assigned to components that demonstrate competent clinical data analysis capabilities.

\paragraph{Weight 3: Critical Clinical Elements}

\begin{itemize}
  \item Represents essential requirements whose failure undermines clinical validity and elements requiring substantial clinical domain knowledge and SQL proficiency.
  \item Examples: Core patient demographic filtering, critical medical code selection, fundamental table relationships, patient cohort construction, database integration with complex joins, clinical analytics.
\end{itemize}

\begin{verbatim}
-- Critical patient cohort construction
WHERE p.gender = 'M'
  AND p.anchor_age BETWEEN 60 AND 70

-- Fundamental table relationships
FROM `physionet-data.mimiciv_3_1_hosp.patients` p
JOIN `physionet-data.mimiciv_3_1_hosp.procedures_icd` pr
  ON p.subject_id = pr.subject_id

-- Essential grouping for per-patient analysis
GROUP BY p.subject_id

-- Critical medical filtering logic
WHERE p.gender = 'M'
  AND p.anchor_age BETWEEN 60 AND 70
  AND (
    -- Comprehensive ICD version and code handling
    (pr.icd_version = 10 AND (...)) OR
    (pr.icd_version = 9 AND (...))
  )
\end{verbatim}

Reserved for components that are absolutely essential for producing clinically meaningful results and require deep understanding of both clinical domain and advanced SQL capabilities.

Scoring aggregation follows the critical-first protocol described in Algorithm~\ref{alg:score-aggregation}.

\subsubsection{Build SQL Query Evaluation Rubric}

Create a hierarchical evaluation tree tailored to your specific clinical question and SQL implementation. The structure should reflect the logical flow of SQL query construction while identifying critical checkpoints. See \autoref{fig:sql_rubric_tree} for an example sql rubric tree.

\subsubsection{Build Results Validation Rubric}

Create validation criteria based on the actual generated CSV file from your gold-standard SQL execution, combined with your clinical knowledge. See \autoref{fig:results_rubric_tree} for an example results rubric tree.

\subsection{Part II: Validation Guidelines}

As a \textbf{Validator}, your role is to ensure every clinical example meets our benchmark standards. To do this, you will perform a comprehensive review of all its components: the natural language question, the SQL query, the executed results, and the evaluation guideline.

\subsubsection{Clinical Question Assessment}

\begin{itemize}
  \item Real-world Relevance: Question represents authentic clinical decision-making scenarios
  \item Medical Language: Accurate clinical terminology and healthcare concepts
  \item Scenario Match: Aligns with designated clinical category
  \item Linguistic Quality: Clear, grammatically sound, and unambiguous phrasing
  \item Query Requirement: Necessitates database analysis, not simple observation
\end{itemize}

\subsubsection{SQL Implementation Review}

\begin{itemize}
  \item Database Standards: Uses correct MIMIC-IV paths (\texttt{physionet-data.mimiciv\_3\_1\_hosp})
  \item Schema Validation: Accurate table references, columns, and join relationships
  \item Medical Logic: Valid age computation, ICD handling, and temporal analysis
  \item Technical Function: Error-free execution with proper NULL management
  \item Query Coverage: Comprehensively addresses clinical question requirements
\end{itemize}

\subsubsection{Output Verification}

\begin{itemize}
  \item Structure: Well-formed CSV with meaningful column labels
  \item Medical Plausibility: Values fall within clinically acceptable boundaries
  \item Data Integrity: Complete dataset without missing essential information
  \item Logic Alignment: Output corresponds to SQL query operations
\end{itemize}

\subsubsection{Evaluation Framework Review}

\begin{itemize}
  \item Component Coverage: SQL evaluation addresses all query elements
  \item Priority Identification: [Critical] labels properly applied to essential parts
  \item Order Dependencies: [Sequential] tags used where sequence matters
  \item Output Standards: Adequate value ranges and format specifications
  \item Assessment Clarity: Unambiguous binary scoring system
\end{itemize}

\subsubsection{Complexity Level Classification}

\begin{itemize}
  \item Difficulty Assessment: Evaluate and categorize the annotated query--SQL pair according to the appropriate complexity level based on SQL complexity and clinical reasoning requirements.
  \item Classification Accuracy: Ensure each example is correctly assigned to Easy, Medium, or Hard difficulty levels to maintain uniform standards throughout the benchmark.
\end{itemize}

\subsubsection{Action Required}

\begin{itemize}
  \item If the example fails \textbf{any} of the above checks, revise it if corrections are minor (e.g., grammar fixes, small SQL adjustments, or evaluation refinements).
  \item If issues are significant (e.g., clinically inappropriate question, fundamentally incorrect SQL, or incomplete evaluation framework), you may \textbf{reject} the example or heavily revise.
  \item Provide brief justification when making revisions or rejections.
\end{itemize}

\subsubsection{Mark as Validated}

Once all checks have passed, mark the example as \textbf{Validated}. This confirms it is ready for inclusion in the final dataset.

\clearpage
\section{Judge Prompt}
\label{app:judge-prompt}
We use an LLM-as-a-judge to score rubric leaf nodes with binary decisions and short explanations. The SQL- and results-level prompt templates are shown in \autoref{fig:judge-sql-prompt} and \autoref{fig:judge-results-prompt}.

\begin{figure}[!h]
\begin{tcolorbox}[colback=black!3!white, colframe=black!70!white, title=Judge Prompt: SQL Evaluation, fontupper=\footnotesize, fonttitle=\footnotesize]

You are evaluating SQL queries for clinical data analysis based on specific requirements.\\

Evaluation Criteria:\\
\textcolor{blue}{\{node.requirements\}} \\

Clinical Question:\\
\textcolor{blue}{\{query\}} \\

SQL to Evaluate (fenced):\\
\textcolor{blue}{\{test\_sql\}} \\

Gold Standard SQL:\\
\textcolor{blue}{\{gold\_sql\}} \\

Instructions:\\
1. Evaluate if the SQL meets the specific requirement: "\textcolor{blue}{\{node.requirements\}}".\\
2. Focus on whether the implementation satisfies the requirement, not on syntactic perfection.\\
3. Use the gold standard SQL as reference for best practices and expected approach.\\
4. Score: 1 if requirement is fully met, 0 if not met.\\
5. Provide a brief explanation of your assessment.\\

Response Format:\\
Score: [0 or 1] \\
Explanation: [Brief explanation of why the score was given]

\end{tcolorbox}
\caption{LLM judge prompt template for SQL-level rubric evaluation.}
\label{fig:judge-sql-prompt}
\end{figure}

\begin{figure}[!h]
\begin{tcolorbox}[colback=black!3!white, colframe=black!70!white, title=Judge Prompt: Results Evaluation, fontupper=\footnotesize, fonttitle=\footnotesize]

You are evaluating clinical query results based on specific requirements.\\

Evaluation Criteria:\\
\textcolor{blue}{\{node.requirements\}} \\

Clinical Question:\\
\textcolor{blue}{\{query\}} \\

Results to Evaluate:\\
\textcolor{blue}{\{test\_results\}} \\

Gold Standard Results:\\
\textcolor{blue}{\{gold\_results\}} \\

Instructions:\\
1. Evaluate if the results meet the specific requirement: "\textcolor{blue}{\{node.requirements\}}".\\
2. For "CSV File Exists" requirements: if results data is shown above and not empty, it means a CSV file exists.\\
3. Use the gold standard results as reference for expected format and values.\\
4. Consider clinical plausibility, data format, and completeness.\\
5. Score: 1 if requirement is fully met, 0 if not met.\\
6. Provide a brief explanation of your assessment.\\

Response Format:\\
Score: [0 or 1] \\
Explanation: [Brief explanation of why the score was given]

\end{tcolorbox}
\caption{LLM judge prompt template for results-level rubric evaluation.}
\label{fig:judge-results-prompt}
\end{figure}

\clearpage
\section{MIMIC-IV Schema}
\label{app:mimic4-schema}
\begingroup
\setlength{\parindent}{0pt}
\raggedright
\small

\textbf{MIMIC-IV --- HOSP module}\par
\medskip

\textbf{admissions}\par
Columns: \texttt{subject\_id, hadm\_id, admittime, dischtime, deathtime, admission\_type, admit\_provider\_id, admission\_location, discharge\_location, insurance, language, marital\_status, race, edregtime, edouttime, hospital\_expire\_flag}\par
\medskip

\textbf{patients}\par
Columns: \texttt{subject\_id, gender, anchor\_age, anchor\_year, anchor\_year\_group, dod}\par
\medskip

\textbf{transfers}\par
Columns: \texttt{subject\_id, hadm\_id, transfer\_id, eventtype, careunit, intime, outtime}\par
\medskip

\textbf{labevents}\par
Columns: \texttt{labevent\_id, subject\_id, hadm\_id, specimen\_id, itemid, charttime, storetime, value, valuenum, valueuom, ref\_range\_lower, ref\_range\_upper, flag, priority, comments}\par
\medskip

\textbf{d\_labitems}\par
Columns: \texttt{itemid, label, fluid, category, loinc\_code}\par
\medskip

\textbf{microbiologyevents}\par
Columns: \texttt{microevent\_id, subject\_id, hadm\_id, micro\_specimen\_id, order\_provider\_id, chartdate, charttime, spec\_itemid, spec\_type\_desc, test\_seq, storedate, storetime, test\_itemid, test\_name, org\_itemid, org\_name, isolate\_num, quantity, ab\_itemid, ab\_name, dilution\_text, dilution\_comparison, dilution\_value, interpretation, comments}\par
\medskip

\textbf{diagnoses\_icd}\par
Columns: \texttt{subject\_id, hadm\_id, seq\_num, icd\_code, icd\_version}\par
\medskip

\textbf{d\_icd\_diagnoses}\par
Columns: \texttt{icd\_code, icd\_version, long\_title}\par
\medskip

\textbf{procedures\_icd}\par
Columns: \texttt{subject\_id, hadm\_id, seq\_num, chartdate, icd\_code, icd\_version}\par
\medskip

\textbf{d\_icd\_procedures}\par
Columns: \texttt{icd\_code, icd\_version, long\_title}\par
\medskip

\textbf{emar}\par
Columns: \texttt{subject\_id, hadm\_id, emar\_id, emar\_seq, poe\_id, pharmacy\_id, enter\_provider\_id, charttime, medication, event\_txt, scheduletime, storetime}\par
\medskip

\textbf{emar\_detail}\par
Columns: \texttt{subject\_id, emar\_id, emar\_seq, parent\_field\_ordinal, administration\_type, pharmacy\_id, barcode\_type, reason\_for\_no\_barcode, complete\_dose\_not\_given, dose\_due, dose\_due\_unit, dose\_given, dose\_given\_unit, will\_remainder\_of\_dose\_be\_given, product\_amount\_given, product\_unit, product\_code, product\_description, prior\_infusion\_rate, infusion\_rate, infusion\_rate\_adjustment, infusion\_rate\_adjustment\_amount, infusion\_rate\_unit, route, infusion\_complete, completion\_interval, new\_iv\_bag\_hung, continued\_infusion\_in\_other\_location, restart\_interval, side, site, non\_formulary\_visual\_verification}\par
\medskip

\textbf{prescriptions}\par
Columns: \texttt{subject\_id, hadm\_id, pharmacy\_id, poe\_id, poe\_seq, order\_provider\_id, starttime, stoptime, drug\_type, drug, formulary\_drug\_cd, gsn, ndc, prod\_strength, form\_rx, dose\_val\_rx, dose\_unit\_rx, form\_val\_disp, form\_unit\_disp, doses\_per\_24\_hrs, route}\par
\medskip

\textbf{pharmacy}\par
Columns: \texttt{subject\_id, hadm\_id, pharmacy\_id, poe\_id, starttime, stoptime, medication, proc\_type, status, entertime, verifiedtime, route, frequency, disp\_sched, infusion\_type, sliding\_scale, lockout\_interval, basal\_rate, one\_hr\_max, doses\_per\_24\_hrs, duration, duration\_interval, expiration\_value, expiration\_unit, expirationdate, dispensation, fill\_quantity}\par
\medskip

\textbf{poe}\par
Columns: \texttt{poe\_id, poe\_seq, subject\_id, hadm\_id, ordertime, order\_type, order\_subtype, transaction\_type, discontinue\_of\_poe\_id, discontinued\_by\_poe\_id, order\_provider\_id, order\_status}\par
\medskip

\textbf{poe\_detail}\par
Columns: \texttt{poe\_id, poe\_seq, subject\_id, field\_name, field\_value}\par
\medskip

\textbf{hcpcsevents}\par
Columns: \texttt{subject\_id, hadm\_id, chartdate, hcpcs\_cd, seq\_num, short\_description}\par
\medskip

\textbf{d\_hcpcs}\par
Columns: \texttt{code, category, long\_description, short\_description}\par
\medskip

\textbf{drgcodes}\par
Columns: \texttt{subject\_id, hadm\_id, drg\_type, drg\_code, description, drg\_severity, drg\_mortality}\par
\medskip

\textbf{services}\par
Columns: \texttt{subject\_id, hadm\_id, transfertime, prev\_service, curr\_service}\par
\medskip

\textbf{provider}\par
Columns: \texttt{provider\_id}\par
\medskip

\textbf{omr}\par
Columns: \texttt{subject\_id, chartdate, seq\_num, result\_name, result\_value}\par
\medskip

\textbf{MIMIC-IV --- ICU module}\par
\medskip

\textbf{icustays}\par
Columns: \texttt{subject\_id, hadm\_id, stay\_id, first\_careunit, last\_careunit, intime, outtime, los}\par
\medskip

\textbf{chartevents}\par
Columns: \texttt{subject\_id, hadm\_id, stay\_id, caregiver\_id, charttime, storetime, itemid, value, valuenum, valueuom, warning}\par
\medskip

\textbf{datetimesevents}\par
Columns: \texttt{subject\_id, hadm\_id, stay\_id, caregiver\_id, charttime, storetime, itemid, value, valueuom, warning}\par
\medskip

\textbf{inputevents}\par
Columns: \texttt{subject\_id, hadm\_id, stay\_id, caregiver\_id, starttime, endtime, storetime, itemid, amount, amountuom, rate, rateuom, orderid, linkorderid, ordercategoryname, secondaryordercategoryname, ordercomponenttypedescription, ordercategorydescription, patientweight, totalamount, totalamountuom, isopenbag, statusdescription, originalamount, originalrate}\par
\medskip

\textbf{ingredientevents}\par
Columns: \texttt{subject\_id, hadm\_id, stay\_id, caregiver\_id, starttime, endtime, storetime, itemid, amount, amountuom, rate, rateuom, orderid, linkorderid, statusdescription, originalamount, originalrate}\par
\medskip

\textbf{outputevents}\par
Columns: \texttt{subject\_id, hadm\_id, stay\_id, caregiver\_id, charttime, storetime, itemid, value, valueuom}\par
\medskip

\textbf{procedureevents}\par
Columns: \texttt{subject\_id, hadm\_id, stay\_id, caregiver\_id, starttime, endtime, storetime, itemid, value, valueuom, location, locationcategory, orderid, linkorderid, ordercategoryname, ordercategorydescription, patientweight, isopenbag, continueinnextdept, statusdescription, originalamount, originalrate}\par
\medskip

\textbf{d\_items}\par
Columns: \texttt{itemid, label, abbreviation, linksto, category, unitname, param\_type, lownormalvalue, highnormalvalue}\par
\medskip

\textbf{caregiver}\par
Columns: \texttt{caregiver\_id}\par
\medskip

\textbf{Concise notes (commonly confusing columns)}\par
\begin{itemize}
    \item \texttt{hadm\_id} vs. \texttt{stay\_id}: \texttt{hadm\_id} is the hospital admission identifier; \texttt{stay\_id} tracks an ICU stay within an admission.
    \item \texttt{charttime} vs. \texttt{storetime}: \texttt{charttime} captures when the event occurred; \texttt{storetime} records when it was entered or verified.
    \item \texttt{itemid}: numeric key for labs, measurements, or medications (lookup in \texttt{d\_labitems} or \texttt{d\_items}).
    \item \texttt{value} / \texttt{valuenum} / \texttt{valueuom}: textual value, numeric value, and unit respectively; use \texttt{valuenum} for calculations.
    \item \texttt{seq\_num}: ordering field for diagnoses/procedures, where lower values often imply higher priority.
    \item \texttt{poe\_id} / \texttt{poe\_seq}: provider order identifier plus sequence; detailed attributes live in \texttt{poe\_detail}.
    \item \texttt{orderid} / \texttt{linkorderid}: link infusion segments and associated orders over time in ICU inputs.
    \item \texttt{interpretation}: microbiology susceptibility call (e.g., S/I/R).
\end{itemize}

\endgroup

\end{document}